\pdfoutput=1

\documentclass[11pt]{article}

\usepackage[preprint]{acl}

\usepackage{times,pifont}

\usepackage{latexsym}

\usepackage[T1]{fontenc}

\usepackage[utf8]{inputenc}

\usepackage{microtype}

\usepackage{inconsolata}

\usepackage{graphicx}

\usepackage{soul}
\usepackage{amsmath}
\usepackage{amsthm}
\usepackage{booktabs}
\usepackage{algorithm}
\usepackage{algorithmic}
\usepackage{amssymb}

\usepackage{multirow}
\usepackage{array}
\usepackage{tcolorbox}
\usepackage{listings}
\title{Conditional and Modal Reasoning in Large Language Models}

\author{
Wesley H. Holliday$^1$ \quad
Matthew Mandelkern$^2$ \quad
Cedegao E. Zhang$^3$ \\
$^1$University of California, Berkeley\\
$^2$New York University\\
$^3$Massachusetts Institute of Technology\\
\small{\texttt{wesholliday@berkeley.edu},
\texttt{mandelkern@nyu.edu},
\texttt{cedzhang@mit.edu}}
}

\begin{document}

\maketitle

\begin{abstract}
The reasoning abilities of large language models (LLMs) are the topic of a growing body of research in AI and cognitive science. In this paper, we probe the extent to which twenty-nine LLMs are able to distinguish logically correct inferences from logically fallacious ones. We focus on inference patterns involving conditionals (e.g., `\textit{If} Ann has a queen, \textit{then} Bob has a jack') and epistemic modals (e.g., `Ann \textit{might} have an ace', `Bob \textit{must} have a king'). These inferences have been of special interest to logicians, philosophers, and linguists, since they play a central role in the fundamental human ability to reason about distal possibilities. Assessing LLMs on these inferences is thus highly relevant to the question of how much the reasoning abilities of LLMs match those of humans. All the LLMs we tested make some basic mistakes with conditionals or modals, though zero-shot chain-of-thought prompting helps them make fewer mistakes. Even the best performing LLMs make basic errors in modal reasoning, display logically inconsistent judgments across inference patterns involving epistemic modals and conditionals, and give answers about complex conditional inferences that do not match reported human judgments. These results highlight gaps in basic logical reasoning in today's~LLMs. 
\end{abstract}

\section{Introduction}\label{Intro}

One of the most distinctive human cognitive abilities is the ability to think about what follows \emph{if} something is the case---conditional thinking---and about what \emph{might} or \emph{must} be the case---modal thinking \cite{evans2004if, Portner2009}. Such reasoning about distal possibilities is crucial to the human capacity for \emph{planning} (we try to choose the action that \emph{would} bring about the best effects \emph{if} we were to take it \cite{GibbardHarper}), \emph{causal reasoning} (C causes E if E \emph{wouldn't} have happened \emph{if} C hadn't \cite{Lewis:1973a,Beller:2023}), \emph{retroactive evaluation}, and more.  

Conditional and modal language has thus been a central focus of philosophers~\cite[e.g.,][]{Stalnaker1968,Lewis1973,KhooBook}, linguists ~\cite[e.g.,][]{KratzerModals,Portner2009}, and logicians~\cite[e.g.,][]{Kripke1963,Stalnaker:1970,vanBenthem2023}, as well as an interest of computer scientists~\cite[e.g.,][]{Friedman1994,Fagin1995}, leading to a variety of sophisticated models of conditional and modal reasoning \cite{Egre2021,Garson:2024}. 

\begin{figure}[t]

\includegraphics[scale=.3]{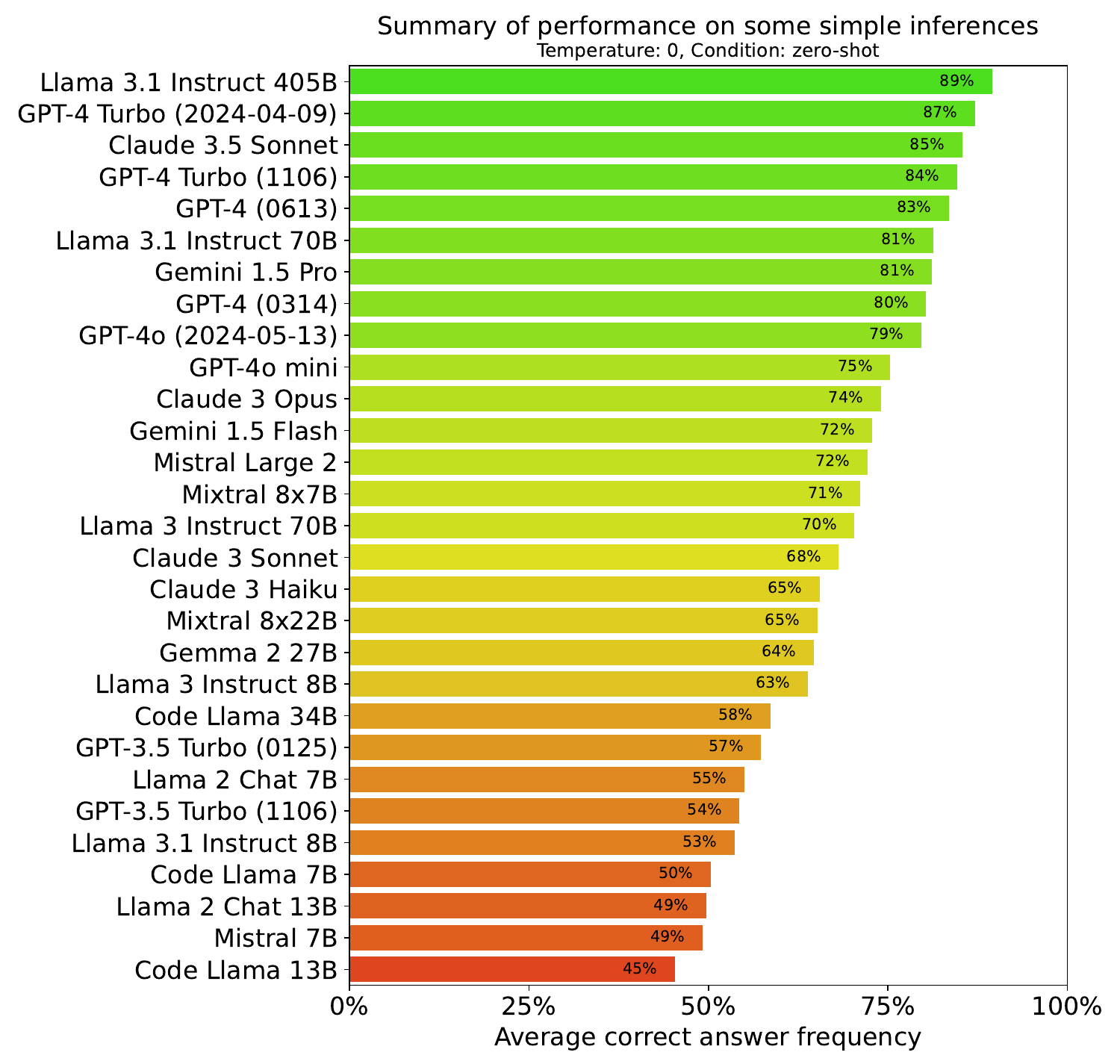}

\caption{Summary of performance on the uncontroversial logical inference patterns discussed in \S~\ref{Infs}. Guessing accuracy is 50\%. Larger models generally perform better, and most models show clear weakness at this task.}\label{Summary}
\vspace{-10pt}
\end{figure}

With the rapid recent development of large language models (LLMs) that at least superficially resemble human speakers and reasoners in many respects \cite{huang2022towards,wei2022chainofthought, bubeck2023sparks,zhao2023survey}, a natural question to ask is to what extent LLMs have mastered conditional and modal reasoning. In this paper, we begin to tackle this problem from the perspective of philosophers and logicians, probing the degree to which different LLMs have mastered the \emph{logical inference patterns} characteristic of conditional and modal reasoning. For example, consider the pattern known as \textit{Modus Tollens} (MT): `If $p$, then $q$. Not $q$. Therefore, not $p$'. We tested whether LLMs draw inferences in accord with this pattern by prompting them with many instances of the pattern, as in:
\begin{quote}

\textbf{User prompt}: From `If Alex finished the race, then Chris finished the race' together with `Chris did not finish the race', can we infer `Alex did not finish the race'? (\textbf{System prompt}: Answer only with `yes' or `no' and nothing else.)

\textbf{GPT-4}: yes. \textbf{Mistral 7B}: no. Etc.

\end{quote}

\noindent We then gave other instances of the pattern to each LLM, asssessing their performance in terms of accuracy on the pattern of inference. \autoref{Summary} summarizes performance across several inference patterns to be discussed. We also compared performance on the zero-shot condition shown above with few-shot and chain-of-thought conditions (\autoref{table:performance}).

After providing some background in \S~\ref{Background} and detailing our experimental setup in \S~\ref{Exp}, we discuss results for a number of inference patterns in \S~\ref{Infs}. We find that among the LLMs tested, all models commit basic fallacies in reasoning with modals and conditionals. Even the best performing models display logically inconsistent judgments across certain inference patterns involving modals and conditionals. And almost all models give answers to certain complex conditional inferences that do not match reported human judgments. We also show that models' performance on our reasoning tasks is highly correlated with that of Chatbot Arena Elo ratings \cite{chiang2024chatbot}, MMLU \cite{hendrycks2020measuring}, and GSM8K \cite{cobbe2021training}, supporting the hypothesis that logical reasoning abilities are predictive of general model capabilities and performance on downstream tasks. 
In sum, our main contributions in this paper are:
\begin{itemize}
    \item Emphasizing the importance and nuances of reasoning about conditionals and modals, grounded in up-to-date evidence and theories from the relevant literature.
    \vspace{-5pt}
    \item Proposing a focused, novel benchmark that tests LLMs' ability to engage in logical reasoning with conditionals and modals. 
    \vspace{-5pt}
    \item Reporting the performance of a large set of LLMs in different prompting settings and identifying some of their gaps and undesirable behaviors in basic logical reasoning.
\end{itemize}

\section{Background and related work}
\label{Background}

Our goal is to apply methodologies from the philosophical, logical, and linguistic literature on conditionals and modals to the study of LLMs. 

\subsection{Logical inference}

First and most generally, we draw on a philosophical understanding of what a \emph{logical inference} is.
Logical inferences are those inferences that are valid just in virtue of the meaning of \emph{logical} words like `and', `or', `not', `if', `must', `might', and so on. That is, a logically valid inference is one whose conclusion is always true when its premises are, \emph{no matter how the non-logical words in the premises and conclusion are understood} \cite{Tarski:1936}. 

This contrasts with more colloquial uses of `logical reasoning' that are current in the literature on LLMs, where `logical reasoning' is often used for reasoning in general, involving inferential leaps of various kinds that go beyond deductive inference proper (this holds for many of the tasks studied in  \citet{xu2023large,chen2023learning,huang2023reasoning,liu2020logiqa} and nearly all the \href{https://github.com/google/BIG-bench/blob/main/bigbench/benchmark_tasks/keywords_to_tasks.md#logical-reasoning}{BigBench} tasks \cite{srivastava2023beyond} categorized under the keyword `logical reasoning'). For instance, in the logicians' sense, the inference `A is to the left of B, hence B is to the right of A' is not \textit{logically} valid, since its correctness depends on the meaning of the non-logical words `left' and `right'. By contrast, `A is to the left of B, hence something is to the left of B' is logically valid, since its correctness relies only on the meaning of the logical word `something'. While studying content-based reasoning in LLMs is obviously of great interest, we believe it is also of fundamental interest to study purely logical reasoning in LLMs, since such reasoning is plausibly part of the backbone of human inference and knowledge of meanings. 

Regarding purely logical reasoning, a series of benchmarks have been created in recent years \cite{tafjord2020proofwriter,tian-etal-2021-diagnosing,han2022folio,saparov2023language,saparov2023testing}, and various strategies have been proposed to solve some of them \cite{creswell2023selection,kazemi2023lambada,olausson-etal-2023-linc, pan-etal-2023-logic, poesia2023certified, ye2023satlm}. Those studies (with the exceptions of \citealt{tafjord2020proofwriter,creswell2023selection}) primarily focus on \textit{multi-step} reasoning, where a proof is required from premises to the hypothesis. Here we target single-step inference patterns, which we treat as more fundamental. Indeed, the inability to recognize the basic inference patterns we study here could provide further explanations of failures on  multi-step reasoning problems. Most importantly, none of the work above studies modal operators, conditional operators, and their interactions, our novel focus in this paper.

\subsection{Modals and conditionals}

Here we draw specifically on the logical and philosophical literature on \textit{modals} and \textit{conditionals}. Enormous progress has been made in the last half century on both topics. First, modal operators (like `must' and `might') have been successfully modeled as \emph{quantifiers over possible worlds} \cite{Kripke1963,Kratzer:1981}. That is, just as `Every boy is sitting' quantifies universally over all boys (in a given domain), `It must be raining' quantifies over all possible worlds (in a given domain---in this case, an \emph{epistemic} domain) and says that it is raining in all of them; and just as `Some boy is sitting' quantifies existentially over boys,  `It might be raining' says that it is raining in some epistemically possible world. Similarly, a deontic modal like `may' can be interpreted as quantifying over a domain of deontically possible worlds, so that `You may eat a cookie' is true just in case you eat a cookie in some deontically possible world, i.e., one where all the actual deontic requirements are satisfied. Likewise, `must', on its deontic interpretation (as in `You must eat this cookie') quantifies \emph{universally} over deontically possible worlds, and says that you eat cookies in all of them.

This interpretation of modals yields corresponding \emph{logics} of modal language, with the details depending on how the domain of possible worlds is obtained (and the interpretation of the other connectives and operators with which modals interact).

Conditional operators have likewise been analyzed with possible worlds semantics. In classical logic, `if $p$, then $q$' is treated as the \emph{material conditional}, which is true whenever $p$ is false or $q$ is true. However, it is almost universally accepted by philosophers, linguists, and logicians that this treatment is a very poor approximation to the actual meaning of `if' in natural language. For instance, on the material analysis of `if', `No student will fail if she studies hard' would entail `Every student will study hard', which obviously does not follow. 
Likewise, if the material analysis were correct, then the probability of `if $p$, then $q$' would go up as the probability of $p$ goes down, but this is wrong. Consider a fair coin. The probability that the coin will land heads if it is flipped is intuitively .5, and it is intuitively probabilistically \emph{independent} of whether the coin is flipped. That is, finding out that the coin {probably} will not be flipped does not make it any more likely that if it is flipped, it will land heads \cite{DouvenVerbrugge}. \citet{Edgington:1995} provides a battery of widely accepted further arguments against the material analysis.

These points are worth emphasizing, since although the material analysis is almost universally rejected by theorists of the conditional, it is still assumed in much existing work testing the logical capacities of humans and LLMs, in both cognitive science and artificial intelligence (e.g., in the recent \citealt{wan2024}, which treats the material analysis as one of the benchmarks of \emph{correct} reasoning with conditionals). This is a serious blindspot, since failing to reason in accord with the material conditional may be logically \emph{correct}; and, conversely, reasoning in accord with the material conditional may be a serious logical mistake. 

The most popular alternative treats `if $p$, then $q$' as a restricted modal operator, which says that $q$ is true in all $p$-worlds (in a given domain). Just as for modals, this yields corresponding logics, with the details again depending on  assumptions about which $p$-worlds are in the domain, together with the interpretation of other connectives \cite{Stalnaker1968,Lewis1973,Egre2021}. 

Although the material analysis is almost universally rejected, there is ongoing controversy about the correct logic of conditionals and modals. 
We have chosen a wide range of inference patterns to test: in many of these cases there is (near) universal agreement about whether the inference pattern is valid. In other cases, there is less agreement about whether the pattern is truly valid, but even in those cases, there is for the most part agreement about whether typical human reasoners are inclined to draw the inference, and the remaining controversy is about how to model those patterns (as genuine (in)validities or the result of systematic shifts in interpretation).
We do not aim to take a position in these complex debates here but rather to compare the behavior of LLMs to widely reported human inferential dispositions. In future work, we plan to compare the behavior of LLMs with human subjects \cite[compare the methodology of][]{10.1162/tacl_a_00293,dasgupta2022language,webson2023language}; here we compare LLMs against expert claims about inference from the literature in philosophy, logic, and semantics. 

\subsection{Natural language inference}
Our task format and evaluation method is similar to the one used in the natural language inference (NLI) paradigm \cite{bowman-etal-2015-large,williams-etal-2018-broad,nie2019adversarial}, which has a rich and long tradition \cite{katz1972semantic, condoravdi-etal-2003-entailment,vanBenthem2008brief,maccartney2009extended,dagan2010recognizing}. There, a problem comes with a premise $P$ and a hypothesis $H$, and the goal is to decide whether the premise entails, contradicts, or is neutral with respect to $H$. The notion of entailment is typically based on common sense, whereas in this work we exclusively study logical entailment in the sense specified above.

In sum, our approach differs from previous work on LLMs in two central ways: (i) we focus on one-step \emph{logical} inference, in the austere philosophical sense, rather than common-sense reasoning in general, differing from most extant benchmarks; (ii) we bring sophisticated approaches to the logic of conditionals and modals from  philosophy, linguistics, and logic, yielding new ways to assess how closely LLMs match human reasoning in this key domain. In particular, in contrast to the work on logical reasoning cited above, we go beyond propositional/predicate logic to incorporate more realistic approaches to the logic of conditionals and modals, which to our knowledge has not been explored.

\begin{table*}[t]
\begin{center}

{\tiny\begin{tabular}{@{}l@{\hspace{0.25mm}}l@{\hspace{0.25mm}}|l@{\hspace{0.25mm}}l@{}}
\toprule
\textbf{Valid Inferences} & \textbf{Examples} & \textbf{Controv. Conditional} & \textbf{Examples} \\ \cmidrule[.8pt]{1-4}

\emph{DS:} & Either Fido is inside or Fido is in the garden. Fido is not in the garden. & \emph{CT}:   & If it's raining, then it's not raining hard.  \\
$p\vee q,\neg q\vdash p$ & $\vdash$ Fido is inside. & $p\to q\vdash \neg q\to\neg p$  & $\vdash$ If it's raining hard, then it's not raining.  \\
\cmidrule{1-4}
\emph{MP:} & If Mary was at the wedding, then Sue was at the wedding. &  \emph{AS}:   & If the match is struck, then it will light.  \\ 
$p\to q,p\vdash q$ & Mary was at the wedding. & $p\to q\vdash (p\wedge r)\to q$ & $\vdash $ If the match is struck and has been soaked in water,  \\
& $\vdash$ Sue was at the wedding. & & then it will light. \\
\cmidrule{1-4}
\emph{MT:} & If Mary was at the wedding, then Sue was at the wedding. & \emph{CMP}:  & If the Warriors don't win, then \\
$p\to q,\neg q\vdash \neg p$ & Sue was not at the wedding. $\vdash$ Mary was not at the wedding. & $p\to (q\to r), p$ & if the Lakers don't win, the Celtics will. \\
\cmidrule{1-2}
\emph{MiN:} & Mary might not have been at the wedding. & $\vdash q\to r$ & The Warriors won't win. \\
$\lozenge\neg p\vdash\neg\Box p$ & $\vdash$ It's not the case that Mary must have been at the wedding. & & $\vdash$ If the Lakers don't win, the Celtics will. \\
\cmidrule{1-2}
\emph{NMu:} & It's not the case that Mary must have been at the wedding. & & \\
$\neg \Box p\vdash \lozenge\neg p$ & $\vdash$ Mary might not have been at the wedding. & & \\
\cmidrule[.8pt]{1-4}
\textbf{Invalid Inferences} &  & 
\textbf{Controv. Modal} &   \\ \cmidrule[.8pt]{1-4}
\emph{AC:} & If Mary was at the wedding, then Sue was at the wedding. & \emph{DSmu:} & Either Fido is inside or Fido must be in the garden.\\
$p\to q, q\vdash p$ & Sue was at the wedding. & 
 $p\vee \Box q, \neg \Box q\vdash  p$ & It's not the case that Fido must be in the garden. \\ 
 & $\vdash$ Mary was at the wedding & & $\vdash$ Fido is inside. \\ \cmidrule{1-4}
\emph{CONV}: & If Mary was at the wedding, then Sue was at the wedding. & 
\emph{DSmi}: & Either Fido is inside or Fido must be in the garden. \\ 
$p\to q\vdash q\to p$ & $\vdash$ If Sue was at the wedding, then Mary was at the wedding. & $p\vee \Box q, \lozenge\neg q\vdash  p$ & Fido might not be in the garden.\\ & & &
 $\vdash$ Fido is inside. \\ \cmidrule{1-4}
  \emph{DA}: & If Mary was at the wedding, then Sue was at the wedding. 
& \emph{MTmi}: & If Mary was at the wedding, then Sue must have been there.\\
$p\to q,\neg p\vdash \neg q$ & Mary was not at the wedding.
& $p\to\Box q,\lozenge\neg p\vdash \neg p$ & Sue might not have been there.\\ & 
$\vdash$ Sue was not at the wedding.
 & & $\vdash $ Mary was not at the wedding.
\\\cmidrule{1-4}
\emph{INV}: & If Mary was at the wedding, then Sue was at the wedding. & \emph{MTmu}: &If Mary was at the wedding, then Sue must have been there.\\
$p\to q\!\vdash\!\neg p\!\to\!\neg q$ & $\vdash$ If Mary was not at the wedding, then Sue was not at the wedding. & $p\to \Box q, \neg \Box q\vdash\neg p$ & It's not the case that Sue must have been at the wedding. \\
 & & & $\vdash$ Mary was not at the wedding.
 \\ 
 \cmidrule{1-4}
\emph{MuDistOr}: & The envelope must have been upstairs or under a bed.& \emph{WSFC}: & John might clean his room or he might go to the lecture. \\
$\Box(p\!\vee\! q)\!\vdash\!\Box p\!\vee\!\Box q$ &  $\vdash$ The envelope  must have been upstairs or it must have been under a bed. & $\lozenge p\vee\lozenge q\vdash\lozenge p\wedge\lozenge q$ & $\vdash$ John might clean his room and he might go to the lecture. \\ \cmidrule{1-4}
\emph{MiAg}: & The envelope might be upstairs and the envelope might be under a bed.  & \emph{NSFC}: &John might clean his room or go to the lecture.\\
$\lozenge  p\! \wedge\!\lozenge q \vdash\! \lozenge (p\!\wedge \!q) $ & $\vdash$ The envelope might be upstairs under a bed. & $\lozenge (p\vee q) \vdash \lozenge p\wedge\lozenge q $ & $\vdash$ John might clean his room and John might go to the lecture.
 \\ 
 \bottomrule
\end{tabular}}
\end{center}
\vspace{-5pt}
\caption{\small Key inferences tested; $p,q$ stand for modal/conditional-free propositions, $\neg$ for `not', $\vee$ for `or', $\to$ for `if$\dots$then', $\lozenge$ for `might', and $\Box$ for `must'. $\varphi_1,\dots,\varphi_n\vdash \psi$ is the inference from the list of premises $\varphi_1,\dots,\varphi_n$ to the conclusion $\psi$. ``Controv.'' stands for controversial inferences. \autoref{Summary} summarizes success on all and only the uncontroversial inferences.}\label{InfList}
\vspace{-5pt}
\end{table*}

\section{Experiments}\label{Exp}

\subsection{Models}\label{Models}

We tested the logical inference judgments of the 29 LLMs listed in \autoref{Summary}, including both open and closed ones, cited in \autoref{sec:llms}. The Anthropic, Google, Mistral, and OpenAI models were run through their respective APIs. Open-weight models were run through cloud providers (together.ai and fireworks.ai). All experiments cost $\sim$\$3,000 for API calls. All code and data for the experiments are available at \href{https://github.com/wesholliday/llm-logic}{github.com/wesholliday/llm-logic}, which also includes data for OpenAI's o1 models (released after this paper was submitted).

\subsection{Data}\label{Data}

For our experiments, we created a bank of questions for dozens of inference patterns, allowing us to individually probe LLMs' inferential capacities with respect to each inference pattern. For each of these, we began by handcrafting a paradigm instance. E.g., for  Modus Tollens (MT) (see \autoref{InfList}), our paradigm instance was: ``From `If Mary was at the wedding, then Sue was at the wedding' together with `Sue was not at the wedding', can we infer `Mary was not at the wedding'?'' With the help other other LLMs (mostly Claude 2, supplemented by GitHub Copilot and Devin), we created 19 additional instances of the pattern, e.g., ``From `If the alien visited Mars, then the robot visited Jupiter' together with `The robot did not visit Jupiter', can we infer `The alien did not visit Mars'?'' To guard against an LLM judging an inference based on world knowledge rather than the inference's logical form, we also created 20 instances with \textit{nonsense predicates}, e.g., ``From `If the flugel was blimmed, then the flugel was zargled' together with `The flugel was not zargled', can we infer `The flugel was not blimmed'?'' We denote the version of an inference with nonsense predicates with an `x' at the end of its name, e.g., MTx. We also tested order effects for all inferences with two premises by switching the order of premises in an `o' version. Using nonsense predicates and switched premise order yields an `ox' version that we also tested. We reviewed all LLM-generated stimuli and made necessary adjustments by hand.

\subsection{Evaluation}

For each of the 20 instances of an inference pattern question and each LLM, we posed the instance to the LLM with temperature 0 and then 1, along with a prompt for either zero-shot, few-shot, or zero-shot chain-of-thought \cite{kojima2022large} conditions (see \autoref{sec:prompts}). For temperature $=1$, we asked the question repeatedly to get an empirical frequency for `yes' and for `no'.\footnote{We used the following stopping rule: if for a particular question the LLM answered `yes' 10 times in a row or  `no' 10 times in a row, then we proceeded to the next question; otherwise we posed the same question a total of 20 times to the model. For the models with the higher costs per token, we reduced 10 to 5 to reduce cost. For OpenAI models, the empirical frequencies obtained closely agree with the log probabilities, but log probs were not available for all models.}

To assess the sufficiency of using 20 instances of each inference pattern, we looked at the Pearson correlation coefficients between the yes-frequency of model responses to the 20 instances of an inference with nonsensical predicates (like MTx) and the yes-frequency of model responses to the 20 instances with sensical predicates (like MT). The overall correlation for all inferences in Table \ref{InfList} is .85. While we could  run more instances for all inference patterns, limited only by time and cost, the high correlations between responses to sensical and nonsensical instances  suggest that we already have enough instances to see how an LLM responds to the logical form in question, not just to contingent features of  particular instances of that form.

We also tested to what extent our results are sensitive to the choice of the word `infer' in our prompts, as opposed to other phrases we consider equivalent in this context: `deduce', `conclude', `logically infer', `logically deduce', and `logically conclude'. Could it be, e.g., that models might commit fewer conditional fallacies if we prompt them with one of these other phrases? We ran the instances of the AC inference (see \autoref{InfList}) using each of the mentioned substitutes for `infer'. The results are qualitatively the same as for `infer', and the correlation coefficients between the yes-frequencies for `infer' and for each of the substitutes are over $.9$. Thus, which of the phrases above we use apparently makes little difference.\footnote{By contrast, there was a bigger difference when we explicitly asked LLMs, ``Is this form of argument logically valid?'' (we call this the `v' variant of each question). This decreased the rate at which some LLMs accepted fallacious inferences, perhaps due to the presence of texts on formal logic in the training data. However, it also decreased the rate at which some LLMs accepted valid inferences (see the GitHub repository). In any case, the qualitative observations we make still apply with this alternative formulation of the questions.}

\section{Results}\label{Infs}

Key inferences we tested are summarized in \autoref{InfList}. To establish a baseline for performance, we tested a variety of inferences whose (in)validity is uncontroversial (as labeled in the left-hand side of the table): DS, MP, MT, AC, DA, and x, o, and ox variants thereof, as well as INV, CONV, MiN, NMu, MuDistOr, MiAg, and x variants thereof. The aggregate performance of our models on these inferences is reported in \autoref{table:performance} for the temperature 0 settings, in addition to \autoref{Summary} at the beginning of the paper. We see that the results significantly vary across models, and performance roughly correlates with model size.  No model achieves above 90\% accuracy on this set of tasks. Moreover, Llama 3.1 405B and 70B are the only open-weights models that achieve 80\%+ accuracy, with the former being the best performer overall. The GPT-4 model family show generally strong performance.

\begin{table}[h!]
\centering
{\scriptsize
\begin{tabular}{lccc}
\toprule
\textbf{Model} & \textbf{0-shot} & \textbf{Few-shot} & \textbf{0-shot Cot} \\ 
$(T = 0)$ & \textbf{Accuracy \%} & \textbf{Delta} & \textbf{Delta} \\ 
\cmidrule{1-4}
Llama 3.1 Instruct 405B & 89.50 & 0.17 & 1.00 \\ 
GPT-4 Turbo (2024-04-09) & 87.17 & 2.00 & 4.33 \\ 
Claude 3.5 Sonnet & 85.33 & 2.83 & 2.33 \\ 
GPT-4 Turbo (1106) & 84.67 & 1.33 & 1.67 \\ 
GPT-4 (0613) & 83.50 & 2.50 & 5.17 \\ 
Llama 3.1 Instruct 70B & 81.33 & 3.33 & 5.50 \\ 
Gemini 1.5 Pro & 81.17 & -4.83 & 7.67 \\ 
GPT-4 (0314) & 80.33 & 5.33 & 7.50 \\ 
GPT-4o (2024-05-13) & 79.67 & 0.50 & 9.00 \\ 
GPT-4o mini & 75.33 & 8.17 & 13.50 \\ 
Claude 3 Opus & 74.00 & 6.00 & 15.83 \\ 
Gemini 1.5 Flash & 72.83 & -2.33 & 13.50 \\ 
Mistral Large 2 & 72.17 & 10.67 & 12.50 \\ 
Mixtral 8x7B & 71.17 & -7.83 & 0.17 \\ 
Llama 3 Instruct 70B & 70.33 & 5.33 & 11.00 \\ 
Claude 3 Sonnet & 68.17 & 2.17 & 12.00 \\ 
Claude 3 Haiku & 65.50 & -3.67 & 11.00 \\ 
Mixtral 8x22B & 65.17 & 12.50 & 21.50 \\ 
Gemma 2 27B & 64.67 & 1.33 & 16.50 \\ 
Llama 3 Instruct 8B & 63.83 & -4.00 & 8.17 \\ 
Code Llama 34B & 58.67 & -7.83 & 2.67 \\ 
GPT-3.5 Turbo (0125) & 57.33 & 2.33 & 17.00 \\ 
Llama 2 Chat 7B & 55.00 & -1.00 & 3.50 \\ 
GPT-3.5 Turbo (1106) & 54.33 & 1.17 & 14.67 \\ 
Llama 3.1 Instruct 8B & 53.67 & 8.00 & 19.50 \\ 
Code Llama 7B & 50.33 & 4.50 & 12.67 \\ 
Llama 2 Chat 13B & 49.67 & 7.83 & 9.00 \\ 
Mistral 7B & 49.17 & 0.17 & 21.17 \\ 
Code Llama 13B & 45.33 & -8.00 & 8.17 \\ 
\bottomrule
\end{tabular}}
\caption{Model performance on uncontroversial inferences with different prompting setups.}
\label{table:performance}
\vspace{-12pt}
\end{table}

We observe that the models perform similarly in the few-shot setting compared to the zero-shot setting (paired t-test not statistically significant). This is presumably because the few-shot examples we use (see Appendix \ref{sec:prompts}) are about conjunctions and disjunctions, not directly about conditionals and modals. The motivation of the few-shot setting is to make the logical reasoning task clear in context. The results suggest that models understand the task  in the zero-shot setting, yet they are not capable of consistently recognizing valid and invalid inference patterns. On the other hand, zero-shot chain-of-thought (CoT) prompting does dramatically improve performance (paired t-test statistically significant):
the best models all achieve accuracy near 90\%, while still systematically making some mistakes with modal or conditionals. These results contribute to the cumulative evidence that CoT elicits and improves reasoning in LLMs \cite{wei2022chainofthought,kojima2022large,suzgun2022challenging}, while pointing to lacunae that persist even with CoT. The performance trends of the temperature 1 settings are similar, and we include figures for those, and for CoT prompting, in \autoref{sec:additional}. Next, we summarize some noteworthy findings based on specific inferences from Table \ref{InfList}.

\subsection{Divergences from the material analysis}

Under the material analysis of conditionals discussed in Section \ref{Background}, CT would be valid.
However, CT is not valid according to the modal analysis of conditionals mentioned in Section \ref{Background}, and indeed, there are well-known intuitive counterexamples to CT \cite{Stalnaker1968}: for instance, the inference from `If it's raining, it's not raining hard' to `If it's raining hard, it's not raining', is obviously not valid, but this inference would be valid if CT were (given very  weak background assumptions). 

Another inference pattern that is valid according to the material analysis is AS. But there are again well-known counterexamples to AS: from `If the match is struck, it will light', we cannot  infer `If the match is struck and it has been soaked in water, it will light'  \cite{Stalnaker1968}.

The LLMs we tested  largely agree with human judgment in rejecting CT and AS (see Appendices \ref{ASappendix}-\ref{CTappendix}). This underscores the importance of not assuming the material analysis when evaluating LLMs, since if we did, we would wrongly ascribe mistakes to them in this case. And it shows that LLMs, like ordinary human speakers, do not interpret the natural language conditional `if$\dots$then' as a material conditional.

\subsection{Inconsistency and overgeneralization}

We tested a number of inference patterns that involve the interaction of modals with conditionals or disjunction.
This is an especially interesting domain, since work in philosophy and logic has shown that substituting a modal sentence for a non-modal one can change a pattern from being apparently valid to apparently invalid.

For instance, $p\vee q, \neg q$ uncontroversially entails $p$ (DS) \emph{provided $p$ and $q$ are Boolean}  (that is, do not themselves contain modals or conditionals). But it is not  clear that $p\vee \Box q$, together with $\neg \Box q$, entails $p$ (DSmu; see \citet{KlinedinstRothschildConnectives}).
For instance, if we know that Fido is either inside or outside, but don't know where he is, then it seems we know (i) Fido is either inside or else must be outside; and (ii) it's not true that Fido must be outside (since he might be inside). But we need not conclude that Fido is inside, contrary to DSmu.

Similarly, we can uncontroversially conclude $\neg p$ from $p\to q$ together with $\neg q$, when $p$ and $q$ are Boolean (MT). But it is not so clear that the inference from $p\to \Box q$ and $\neg \Box q$ to $\neg p$ (MTmu) is valid \cite{YalcinMT}. It seems that we can know (i) if Fido is not inside, he must be outside and (ii) it's not true that Fido must be outside (since he might be inside), without being compelled to conclude that Fido is inside, contrary to MTmu. 

For a final case, \citet{McGee:1985} pointed out that, while the inference from $p\to q, p$ to $q$ is obviously valid when $p,q$ are Boolean (MP), the inference from $p\to(q\to r),p$ to $q\to r$ (CMP) appears \emph{invalid}. Suppose that the Lakers, Warriors, and Celtics are the only finalists in a tournament, so it's certain that (i) if the Warriors don't win, then if the Lakers don't win, the Celtics will. Suppose moreover that (ii) the Warriors are very likely not to win, because the Lakers are way ahead. But now suppose further that the Celtics are heavy underdogs, so that if the Lakers don't win, the Warriors probably will. We then can't conclude from (i) and (ii) that it's likely that, if the Lakers don't win, the Celtics will. But this would follow if CMP were valid, since valid inference preserves probability.

These cases are of special interest to test on LLMs since human subjects can immediately recognize that while DS, MT, and MP are valid for Boolean sentences, they apparently fail for modal/conditional substitution instances of these patterns like DSmu, MTmu, and CMP.
That is, humans \emph{do not overgeneralize from the simple (Boolean) case to the general case}. It is thus interesting to explore whether LLMs are human-like in this respect or rather overgeneralize from simple to complex cases---especially since these complex cases have only been discussed in a relatively small number of philosophy and logic papers and are presumably somewhat rare in naturalistic settings and hence presumably not very frequent in training data.

\begin{figure}[h!]
\begin{center}
\includegraphics[width=0.95\linewidth]{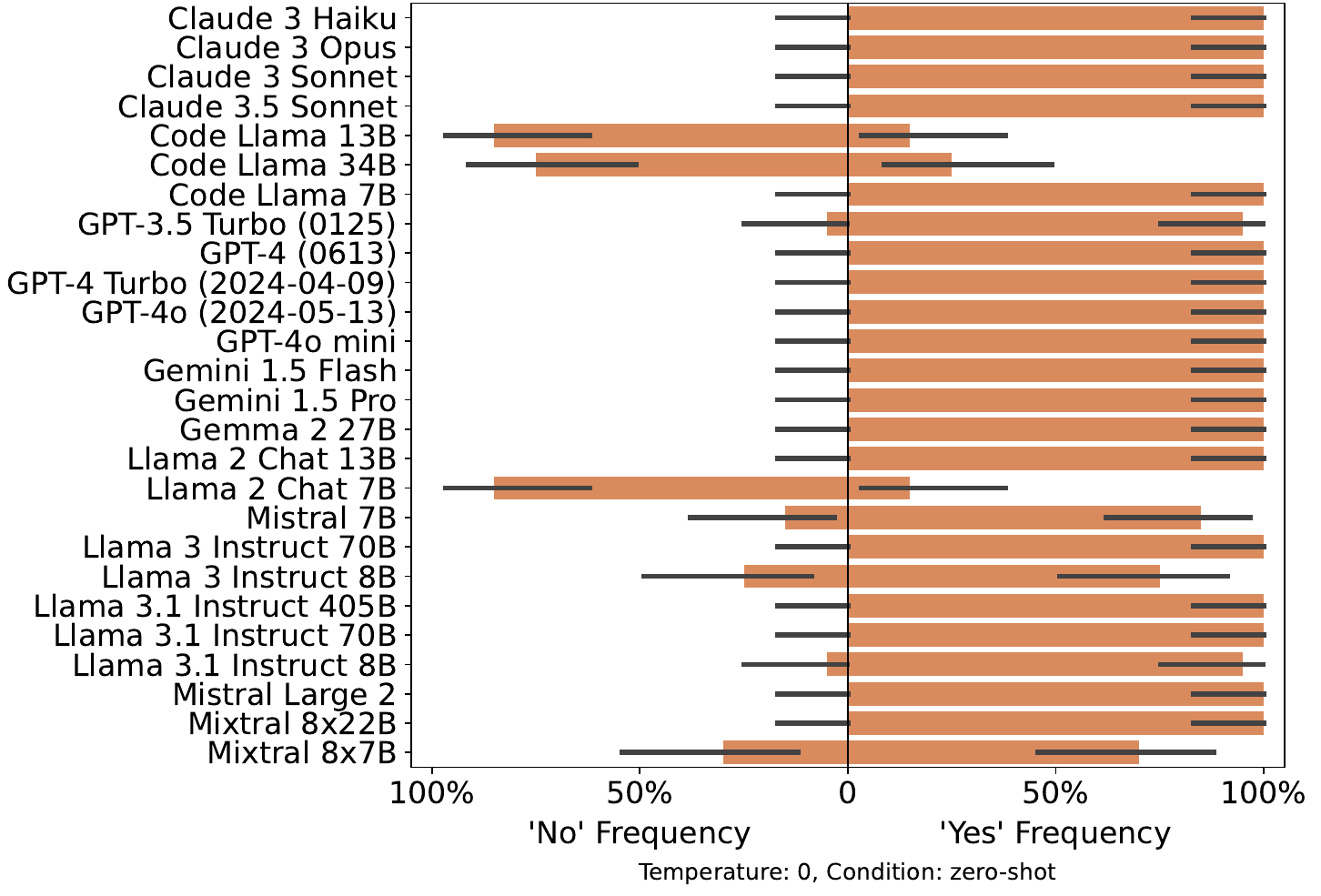}
\includegraphics[width=0.95\linewidth]{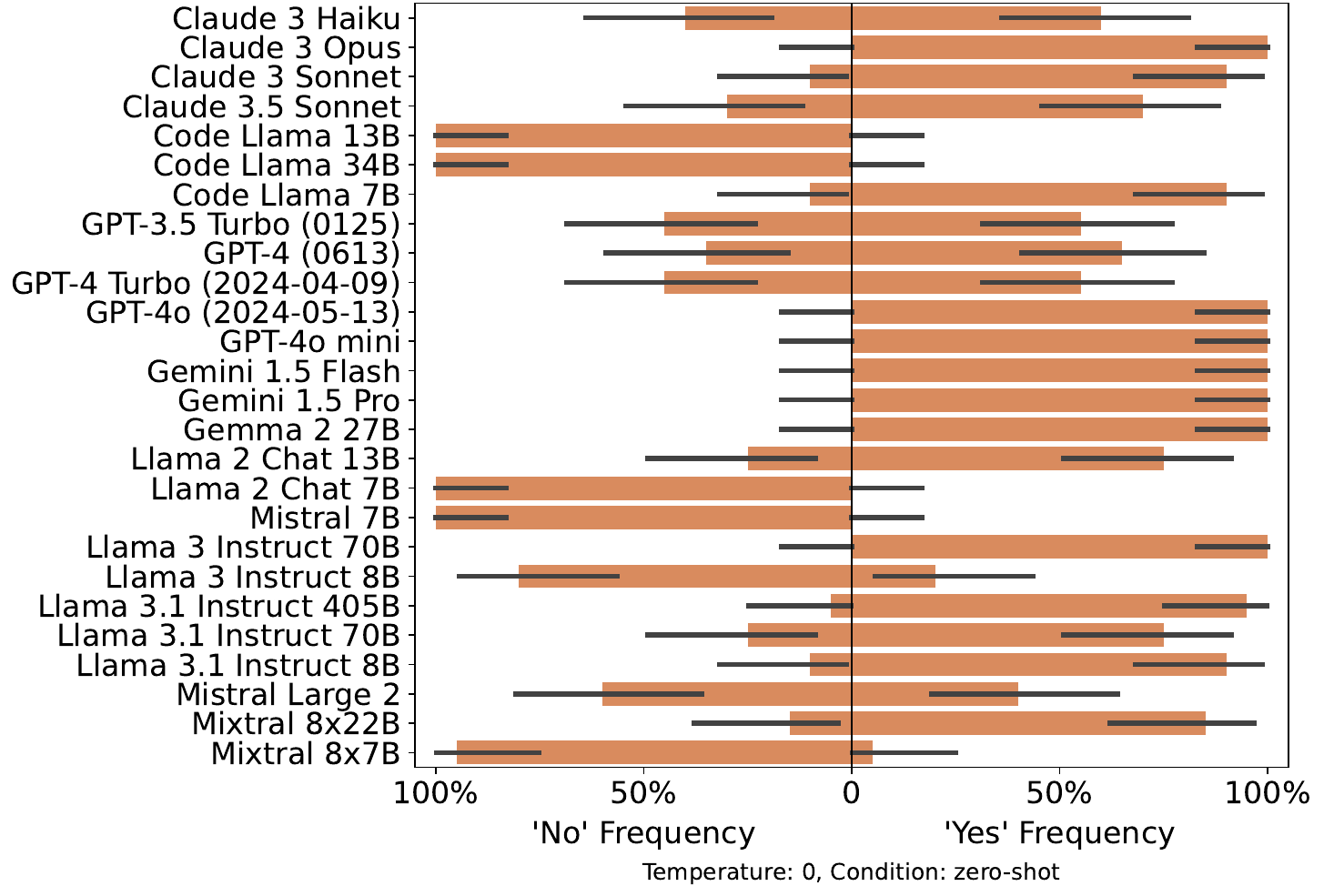}
\end{center}
\vspace{-10pt}
\caption{Zero-shot responses for MTmu (above) and MTmi (below) show inconsistency for many~models. All error bars, including in subsequent figures, represent 95\% confidence intervals.}
\label{MT}
\vspace{-5pt}
\end{figure}

We found that, indeed, many LLMs do overgeneralize: they do not  agree with human judgments about the invalidity of MTmu, DSmu, and CMP. Intriguingly, some models exhibit human-like judgments in rejecting MTmi, but at the same time they accept MTmu (\autoref{MT}).  This is logically inconsistent, since those models also accept that `might not' is logically equivalent to `not must' (MiN and NMu), \footnote{In fact, the inconsistenty arises just using MiN. For given MiN, the premises of MTmi entail the premises of MTmu, so one cannot reject the former and accept the latter.} 
and MTmi and MTmu differ only by substituting these equivalent phrases. This intriguing pattern suggests that LLMs may indeed overgeneralize from the validity of MT to judging MTmu to be valid, while recognizing that MTmi (which is not syntactically an instance of MT) is invalid.\footnote{Our examples were designed to elicit an epistemic reading of the modals in question. However, this is immaterial to our logical points, since even if the target LLM instead accessed a deontic reading, the judgments we report would remain jointly inconsistent, as long as the LLM's interpretation does not change across different instances of the modal in question.}

\begin{figure}[h!]
\begin{center}

\includegraphics[width=0.95\linewidth]{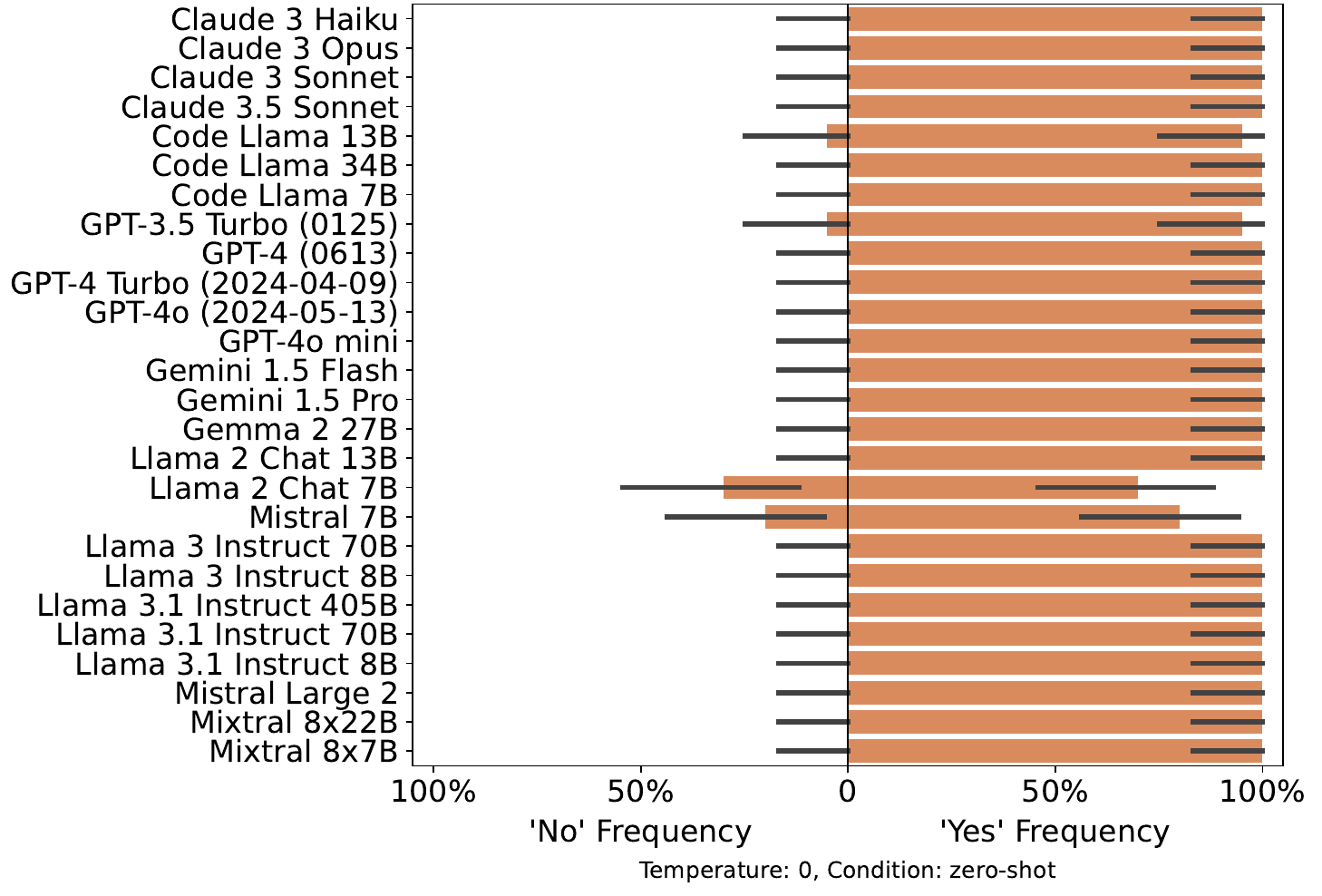}
\includegraphics[width=0.95\linewidth]{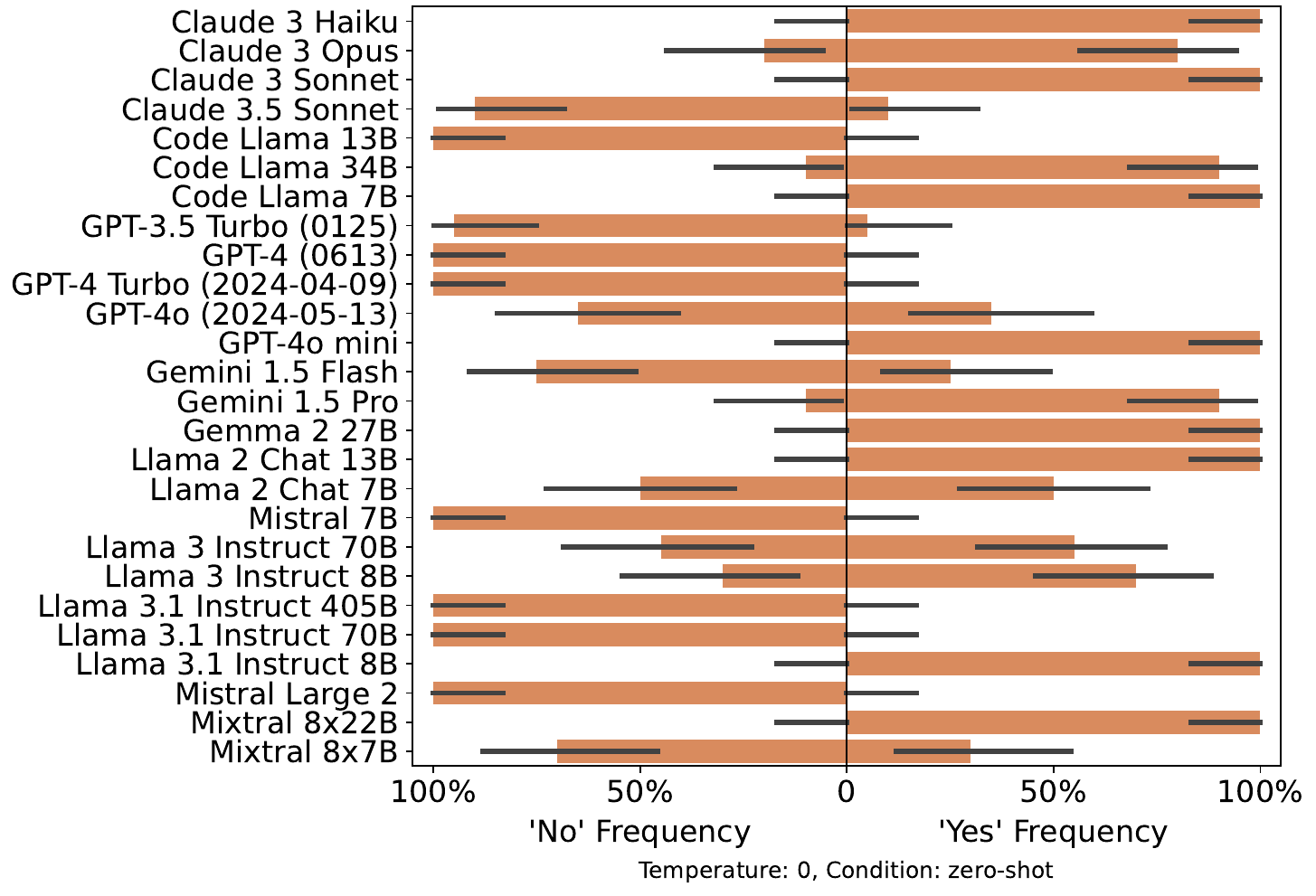}
\end{center}
\vspace{-10pt}
\caption{Zero-shot responses for DSmu (above) and DSmi (below) show inconsistency for many models.}
\label{DS}
\vspace{-5pt}
\end{figure}

\begin{figure}[h!]
\centering
\includegraphics[scale=.38]{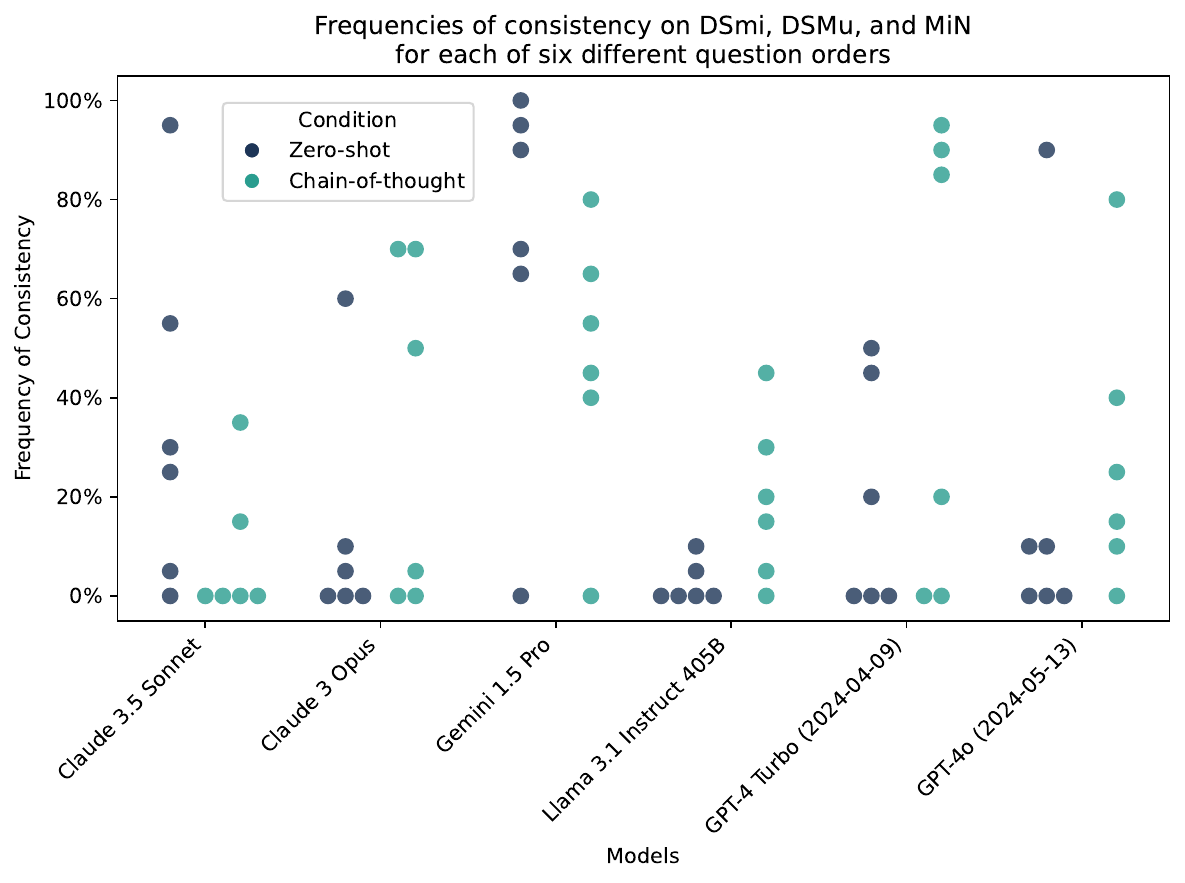}
\vspace{-15pt}
\caption{Percentage of responses that were jointly consistent when we asked leading models about DSmu, MiN, and DSmi in the same context window, in one of the six possible orders. Each dot represents such an order. The results show strong sensitivity to question order, which is highly undesirable.}
\label{MeanConsistency}
\vspace{-10pt}
\end{figure}

\begin{figure}[h!]
\begin{center}
\includegraphics[width=0.95\linewidth]{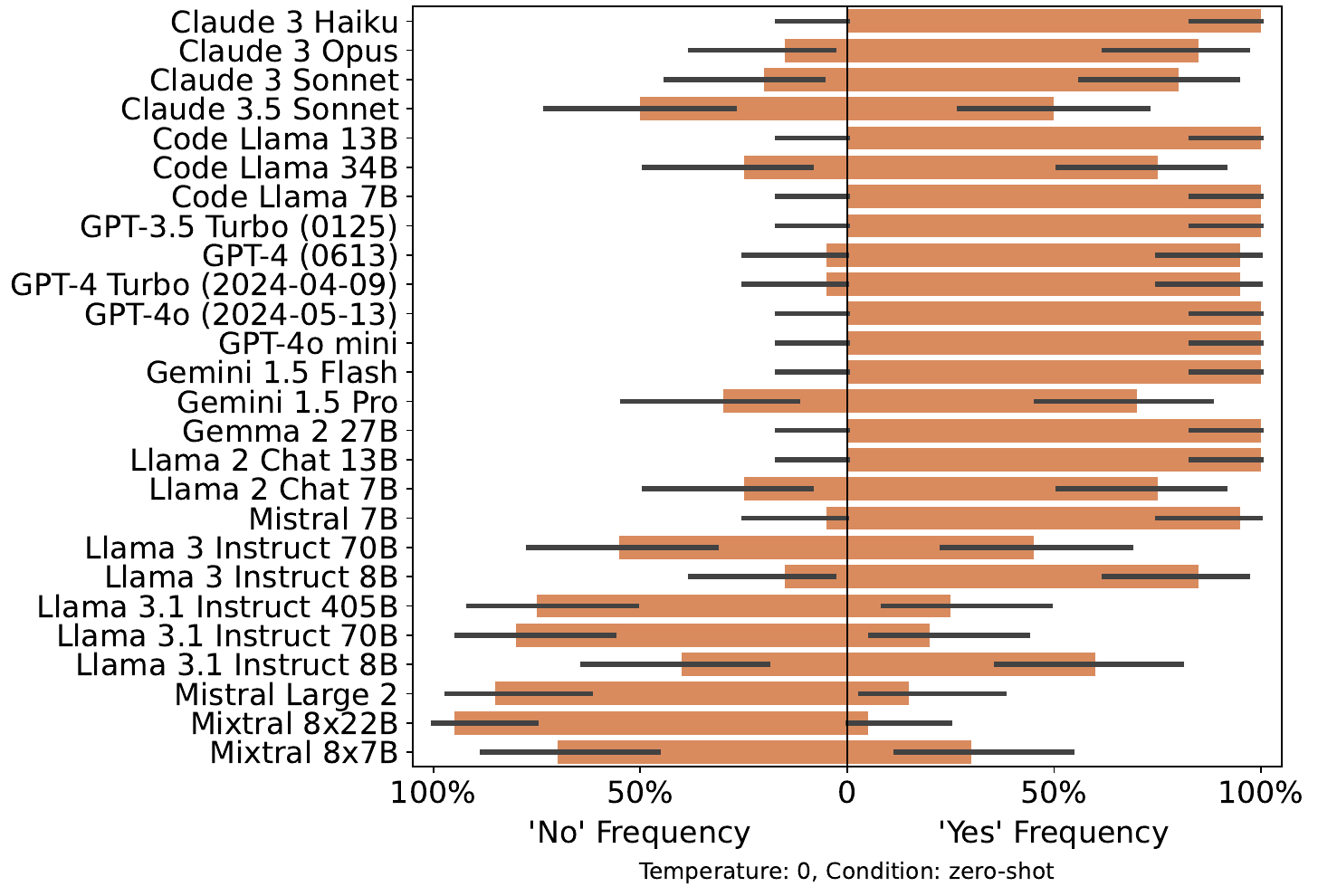}
\includegraphics[width=0.95\linewidth]{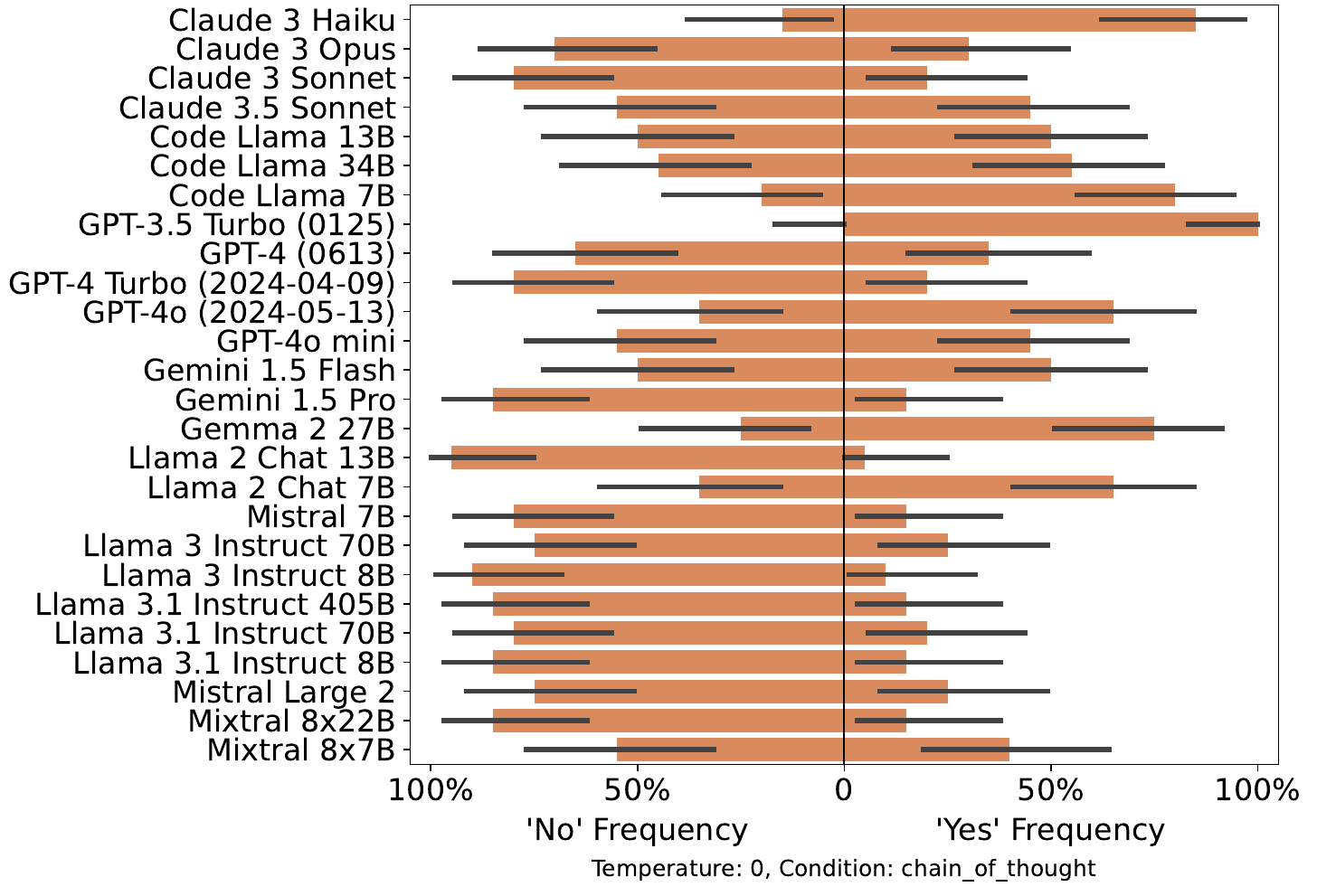}
\end{center}
\vspace{-5pt}
\caption{Responses for CMP, zero-shot (above) and chain-of-thought (below); LLMs were asked whether the inference preserved likelihood, i.e., if $q\to r$ must be likely when $p\to(q\to r)$ is certain and $p$ is likely.}
\label{CMPzero}
\vspace{-15pt}
\end{figure}

\begin{figure*}[h!]
\begin{center}
\includegraphics[width=1\linewidth]{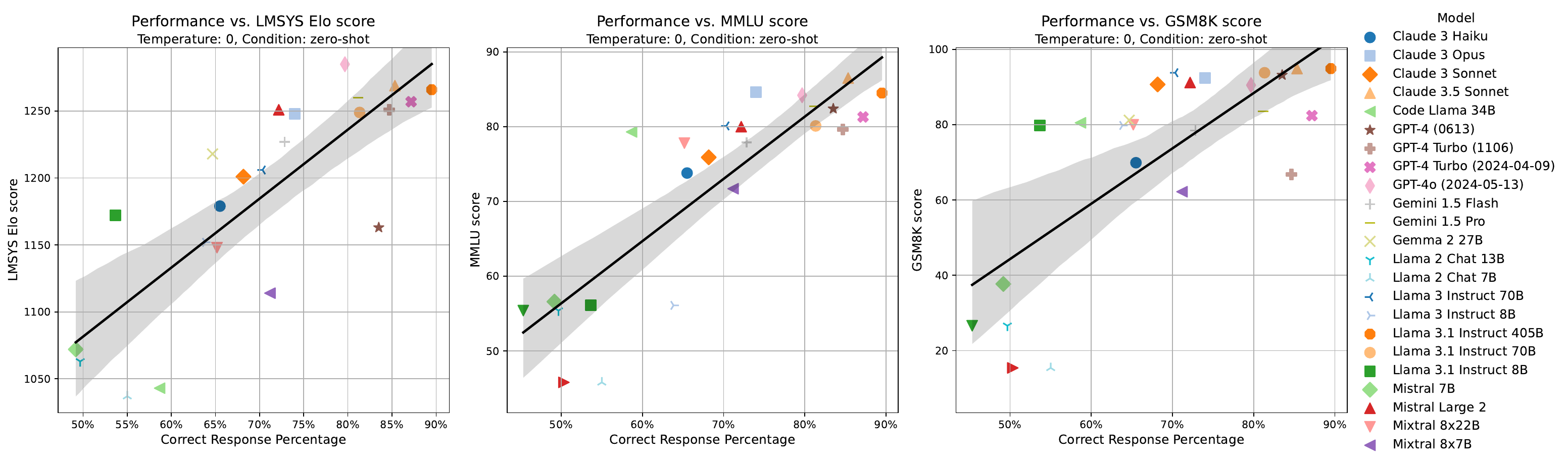}
\end{center}
\vspace{-10pt}
\caption{Correlations of our evaluation results (zero-shot) vs.~LMSYS Elo ratings, MMLU scores, and GSM8k scores. The correlations are 0.81, 0.85, and 0.75, respectively. All p-values are less than 0.01.}
\label{correlations}
\vspace{-10pt}
\end{figure*}

We found similar patterns for DS, as shown in Figure \ref{DS}, where many models accepted DSmu but rejected DSmi. Indeed, even the GPT-4 models we tested exhibit logically inconsistent behavior in this area, accepting DSmu and MiN while rejecting DSmi. This inconsistency was displayed for both zero-shot and chain-of-thought prompts. We wondered whether this incoherence could be due to the fact that the inconsistent triad of responses was given in response to different prompts in different context windows, so we probed more deeply, asking the leading models about each of the three inference patterns within the same context window, but their responses remained logically inconsistent much of the time; see \autoref{MeanConsistency} for a summary.

Also strikingly, the models that performed best overall on other inferences uniformly judge CMP to be valid with zero-shot prompts, contrary to human judgment. Performance improved substantially with CoT prompts, but most of the best-performing models still accept CMP at surprisingly high rates (\autoref{CMPzero}), pointing to a stark contrast between reflective human judgment and state of the art LLM performance. Along with our findings about MT and DS, this supports the hypothesis that LLMs may overgeneralize from the validity of inference patterns for Boolean sentences to the unrestricted validity of those patterns for all substitutions. 

\subsection{Modal fallacies and free choice}

We also found intriguing patterns involving LLMs' behavior with purely modal inferences, shown in Appendix~\ref{MuDistOr}-\ref{WSFC}. While, as noted, the models were generally able to correctly reason about duality (inferring `not must' from `might not', and vice versa), they systematically made basic errors in modal reasoning in other cases. Strikingly, almost all models, including those that performed best overall, accepted both MuDistOr and MiAg as valid, despite these being clearly invalid. (For MuDistOr, asssume the keys must be upstairs or downstairs; it doesn't follow that either they must be upstairs or they must be downstairs. After all, they might not be upstairs, and they might not be downstairs. For MiAg, assume the keys might be upstairs and might be downstairs; it doesn't follow that they might be both upstairs and downstairs.) The situation with WSFC and NSFC is also  interesting. There is controversy in the semantics literature about the status of these patterns, which strike many speakers as valid but are not valid on standard modal semantics \cite{Kamp73}. Intriguingly, many models accepted NSFC but rejected WSFC. In fact, this conforms to one position in the literature which maintains that, indeed, NSFC but not WSFC is valid \cite{Simons:2005,Meyer:2017,Fusco:2019}. However, this threatens inconsistency, since most models also accept the (standardly valid) equivalence of $\lozenge(p\vee q)$ with $\lozenge p\vee \lozenge q$ (see the GitHub repository).

\subsection{Relationship to some popular benchmarks}

To position our logical reasoning tasks with respect to the broad landscape of LLM evaluations, we compare the models' performance on our benchmark (uncontroversial inferences) to that of Chatbot Arena \cite[general assistance,][]{chiang2024chatbot}, MMLU \cite[domain knowledge,][]{hendrycks2020measuring}, and GSM8K \cite[math reasoning,][]{cobbe2021training}, all of which are popular benchmarks to assess LLM capabilities. We show that our results are highly correlated with the results of each of the three in \autoref{correlations}. The Arena Elo ratings come from LMSYS directly.\footnote{\url{https://chat.lmsys.org}} The MMLU and GSM8k scores are obtained from the HELM leaderboard \cite{liang2022holistic}.\footnote{\url{https://crfm.stanford.edu/helm}} The high correlations support the hypothesis that logical reasoning abilities are related to and predictive of not only mathematical reasoning abilities but also domain-general capabilities. It would be interesting to investigate the causal connections: whether improving logical reasoning also improves some general reasoning abilities.

\section{Discussion}

On the one hand, our results mirror what we have learned in the field over the past few years: larger models are likely to perform better at reasoning, and chain-of-thought prompting often but does not always help. On the other hand, we have identified inconsistent and counterintuitive reasoning behaviors from even the best models with and without chain-of-thought. The inference patterns that give rise to those behaviors reflect state-of-the-art research in philosophy and logic, which by their nature means they are less present in the training and fine-tuning data of LLMs (though most likely not absent). This is suggestive of sources from which we can acquire more out-of-distribution evaluation data to test LLMs, and it illustrates that LLMs' judgments may not be reliable when they encounter novel inference patterns, even if those are natural and intuitive for humans. Lastly, we note that a neurosymbolic, semantic-parsing approach to logical reasoning, as in \citealt{olausson-etal-2023-linc}, will not automatically handle the tasks we present here. To our knowledge, theorem provers do not natively implement the nuanced meaning representations we discussed in this paper, and in some cases it is still controversial what the correct semantics for certain modal operators are. Thus, working towards an automated natural language reasoning system that sensibly solves our tasks is a challenge regardless of the models and approaches. 

\section{Conclusion}

We have drawn on philosophical, logical, and linguistic analyses of inference to explore the extent to which current LLMs are accurate logical reasoners about conditionals and modals. We hope this will be the start of a new strand of research on LLMs. There are many natural follow-ups. First, it would be interesting to compare the behavior of LLMs with experimental human subjects on all the inferences we tested. We have reported expert human judgments from the literature, but the judgments of experimental subjects might exhibit mistakes interestingly similar to or different from those we find in LLMs. Second, there are many more inference patterns to explore, involving modals, conditionals, as well as many other logical constructions, like  quantifiers, attitude predicates, and degree constructions. Third, modal and conditional reasoning is also  connected to probabilistic reasoning, causal models, and mental simulations, and hence could provide another perspective for studying these in LLMs and humans. More generally, we hope that there will be more work studying LLMs using  philosophical, linguistic, and logical insights into the building blocks of reasoning and meaning.

\section{Limitations}
In this paper, we have focused on studying to what extent LLMs understand logical inferences with conditionals and modals. The conditional connective and `must' and `might' operators are the target, while we do not systematically study other important logical operators such as `and' and `or', or quantifiers such as `some' and `all'. Even in the modal domain, operators like `probably' and `certainly' also deserve careful analysis. In particular, the interactions of many of these different kinds of operators could lead to interesting inference patterns, which would be worthy of future study. Additionally, we employ simple syntactical constructions to create the evaluation data. While this is natural for testing logical inference, the models' performance on an inference pattern may not generalize when it is made of more complex phrases. Lastly, our dataset is in English, and the models' logical inference abilities may differ by language. We hope future work could study the multilingual aspect of logical inference as well.

\section*{Acknowledgments}

For helpful feedback, we thank Dominic Hughes, James Kirkpatrick, Roger Levy, Hailey Schoelkopf, Shane Steinert-Threlkeld, and the reviewers for EMNLP 2024.

\bibliography{llm}

\appendix

\newpage
\ 
\newpage

\section{Language models used}
\label{sec:llms}

In this section, we cite all the LLMs used to conduct our experiments: the GPT-4 model family \cite{openai2023gpt4}; the GPT-3.5 model family \cite{brown2020language};\footnote{\url{https://openai.com/index/chatgpt}} the Claude 3 model family\footnote{\url{https://www.anthropic.com/news/claude-3-family}} and Claude 3.5 Sonnet;\footnote{\url{https://www.anthropic.com/news/claude-3-5-sonnet}} Gemini 1.5 Pro and Flash \cite{reid2024gemini}; Gemma 2 27B \cite{team2024gemma}; the Llama 3 model family \cite{dubey2024llama}; Mistral Large 2;\footnote{\url{https://mistral.ai/news/mistral-large-2407}} Mixtral 8x7B \cite{jiang2024mixtral} and Mistral 7B \cite{jiang2023mistral}; Llama 2 Chat 7B and 13B \cite{touvron2023llama}; Code Llama 7B, 13B, and 34B \cite{roziere2023code}.

\section{Prompts used in the experiments}
\label{sec:prompts}

We use the following prompts to run the experiments reported in this work. The prompts are the same for all models, except that for some small models (e.g., Llama 2 7B and Code Llama 7B), we had to modify the chain-of-thought prompt with extra reminders to answer only after first thinking step by step. Each prompt is shown with a concrete example of an inference pattern.

\subsection{Zero-shot}
\begin{lstlisting}[breaklines=true]
Answer only with 'yes' or 'no' and nothing else.

From 'If Mary was at the wedding, then Sue was at the wedding' together with 'Sue was not at the wedding', can we infer 'Mary was not at the wedding'?
\end{lstlisting}

\subsection{Few-shot}
\begin{lstlisting}[breaklines=true]
Consider the following examples:

From 'Ann went to the store', can we infer that 'Ann went to the store and Bob went to the beach'? Correct answer: No.

From 'Ann went to the store', can we infer that 'Ann went to the store or Bob went to the beach'? Correct answer: Yes.

From 'Ann went to the store and Bob went to the beach', can we infer that 'Ann went to the store'? Correct answer: Yes.

From 'Ann went to the store or Bob went to the beach', can we infer that 'Ann went to the store'? Correct answer: No.

Now here is a question for you:

From 'If Mary was at the wedding, then Sue was at the wedding' together with 'Sue was not at the wedding', can we infer 'Mary was not at the wedding'?
\end{lstlisting}

\subsection{Zero-shot chain-of-thought}
\begin{lstlisting}[breaklines=true]
In response to the following question, think step by step and explain your reasoning; then when you are ready to answer, simply write 'Answer: ' followed by 'yes' or 'no'.

From 'If Mary was at the wedding, then Sue was at the wedding' together with 'Sue was not at the wedding', can we infer 'Mary was not at the wedding'?
\end{lstlisting}

\section{Additional results}
\label{sec:additional}

\subsection{Individual inferences}\label{IndInfAppendix}

In this Appendix, we display the zero-shot performance of the LLMs on the inferences shown in Table \ref{InfList} when operating at both temperature 0 and temperature 1.

\subsubsection{Disjunctive Syllogism (DS)}

\begin{center}
\includegraphics[scale=.3]{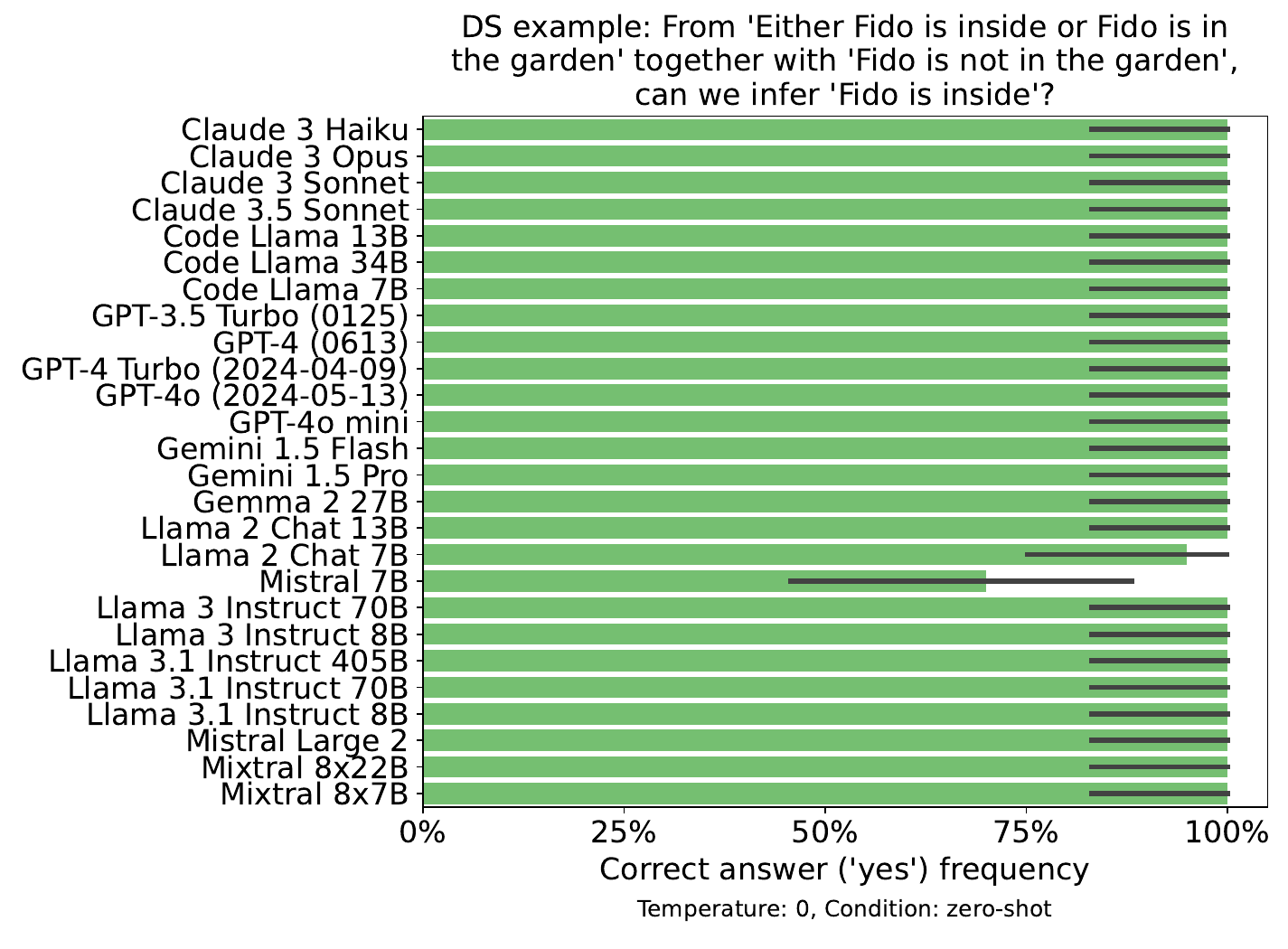}
\includegraphics[scale=.3]{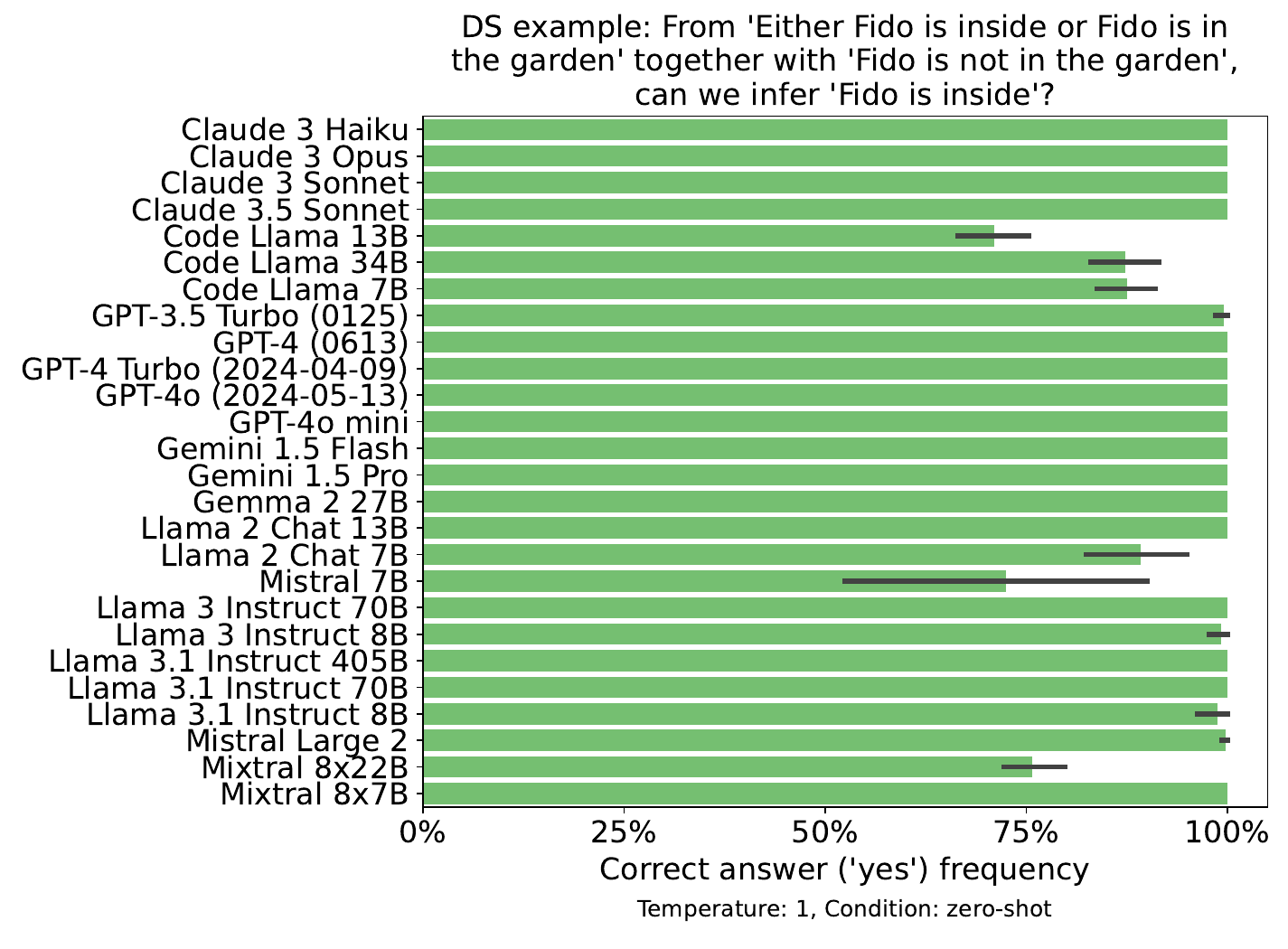}
\end{center}

\subsubsection{Modus Ponens (MP)}

\begin{center}
\includegraphics[scale=.3]{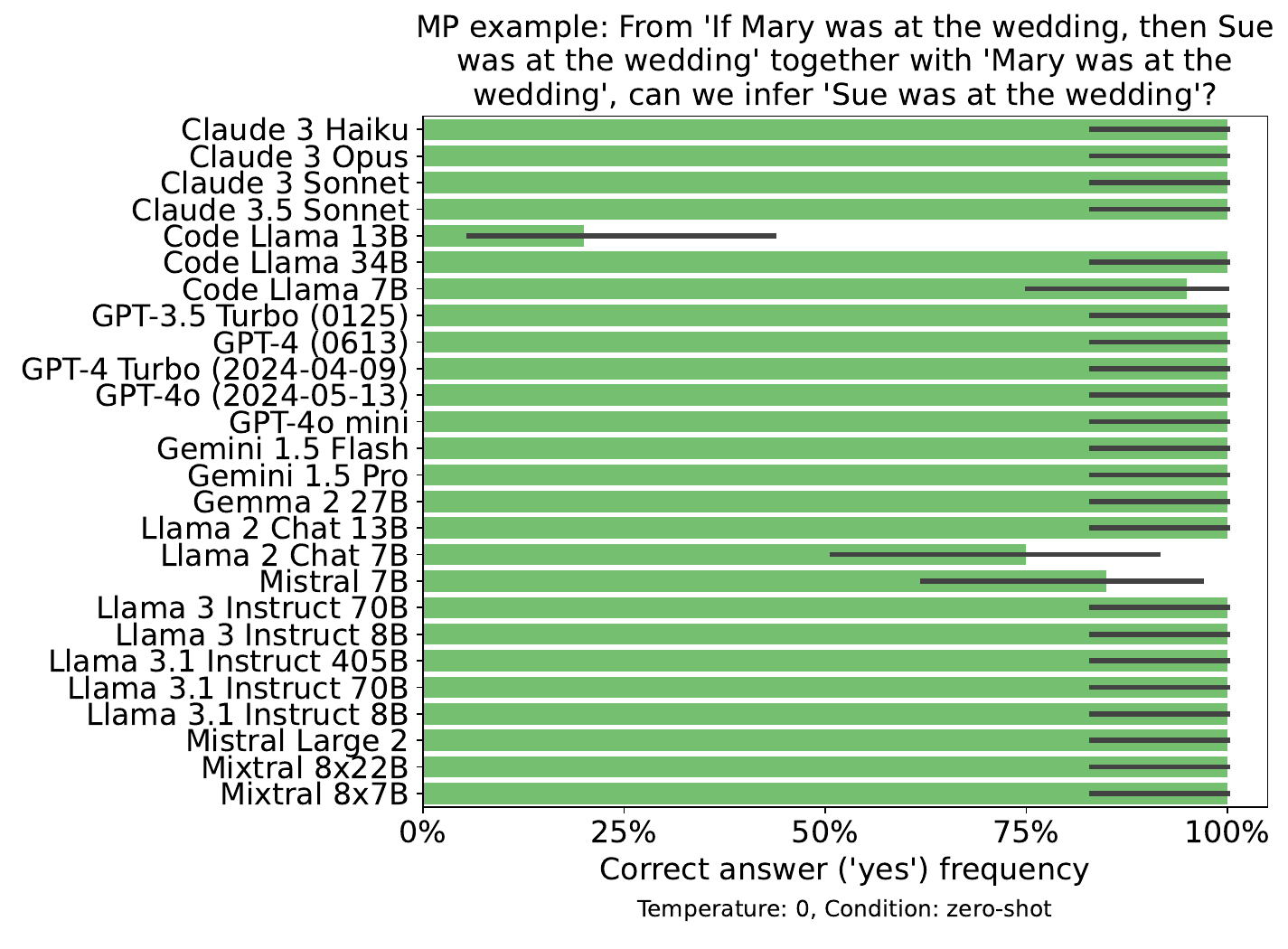}
\includegraphics[scale=.3]{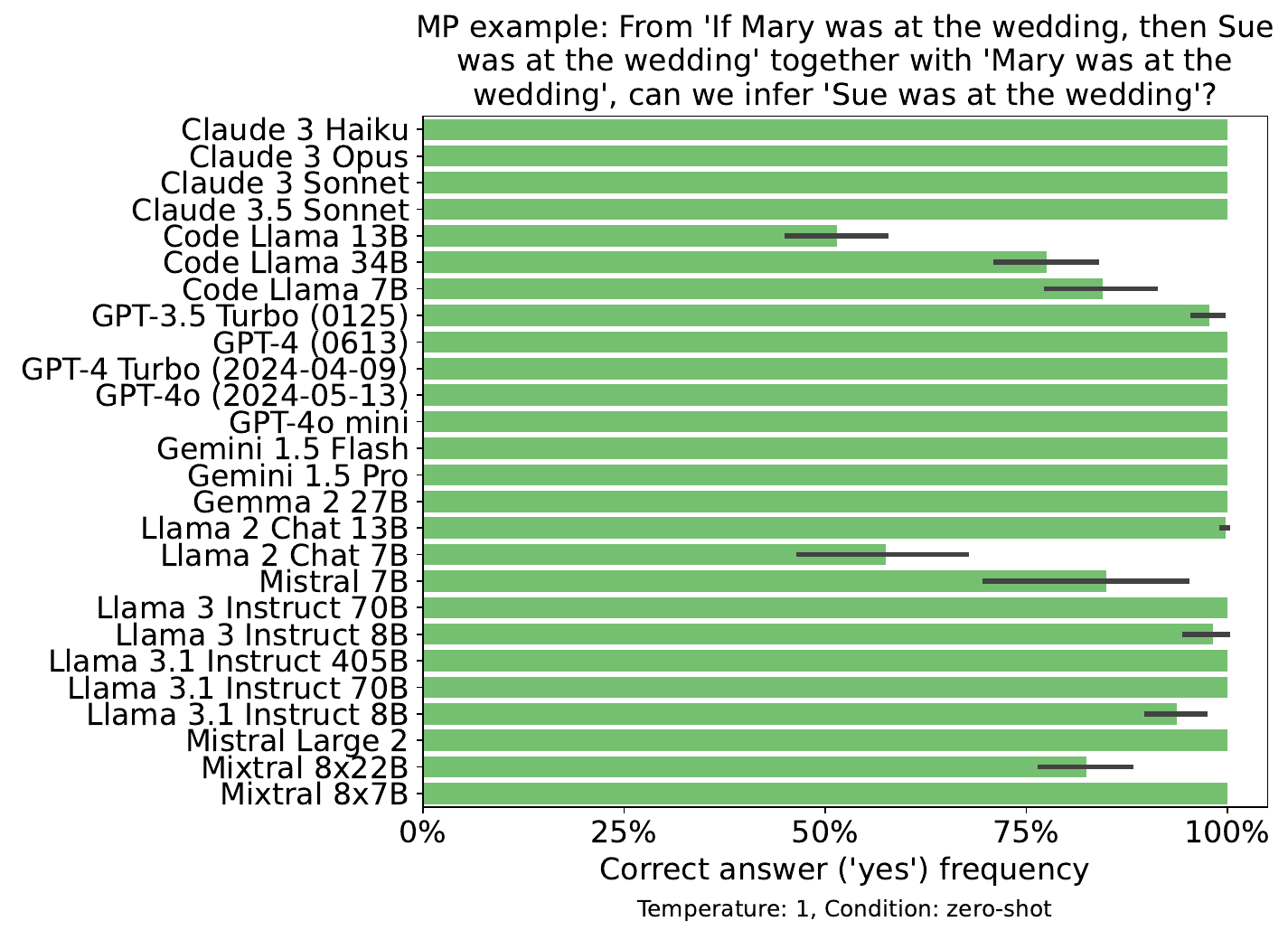}
\end{center}

\subsubsection{Modus Tollens (MT)}

\begin{center}
\includegraphics[scale=.3]{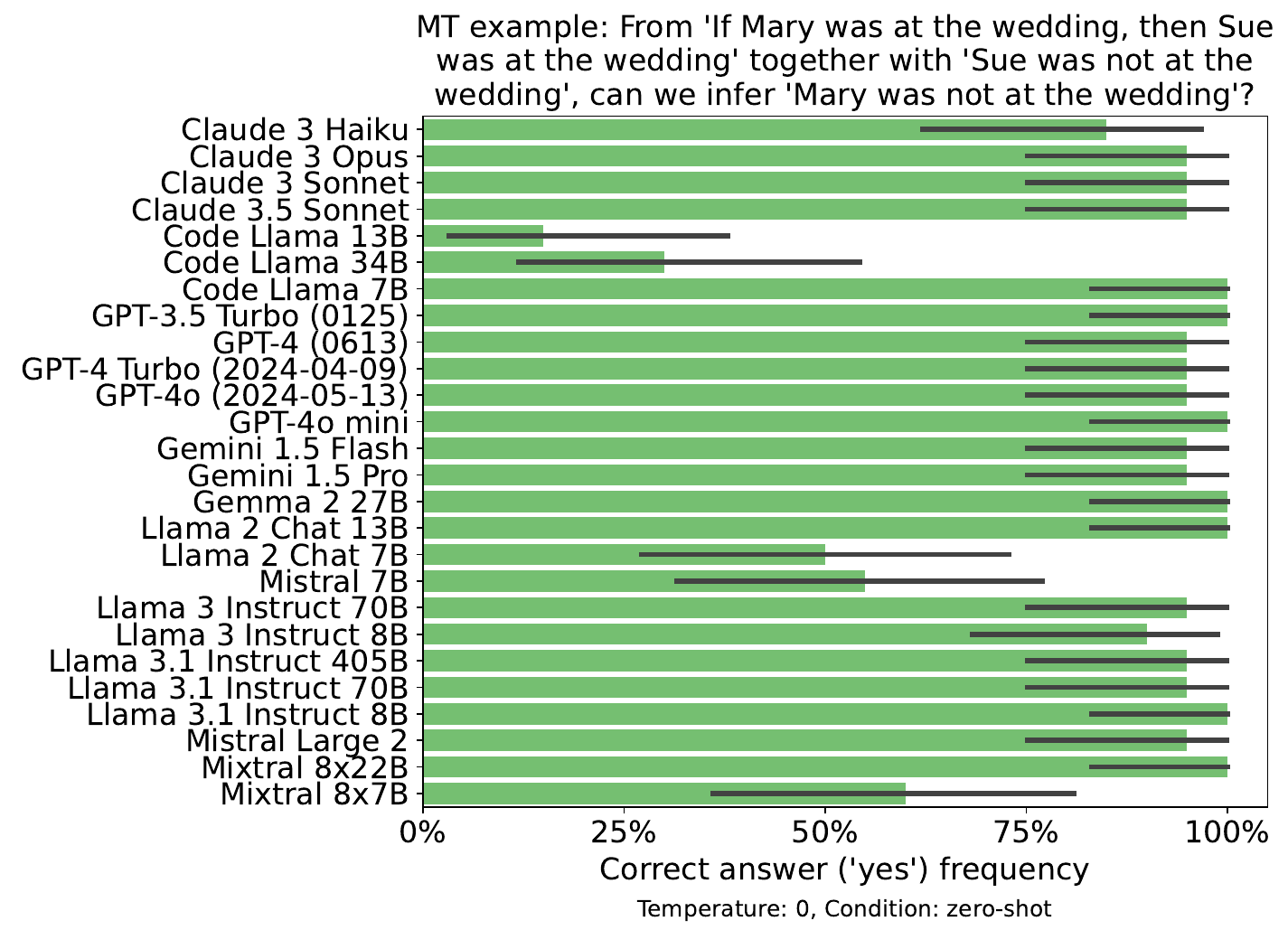}
\includegraphics[scale=.3]{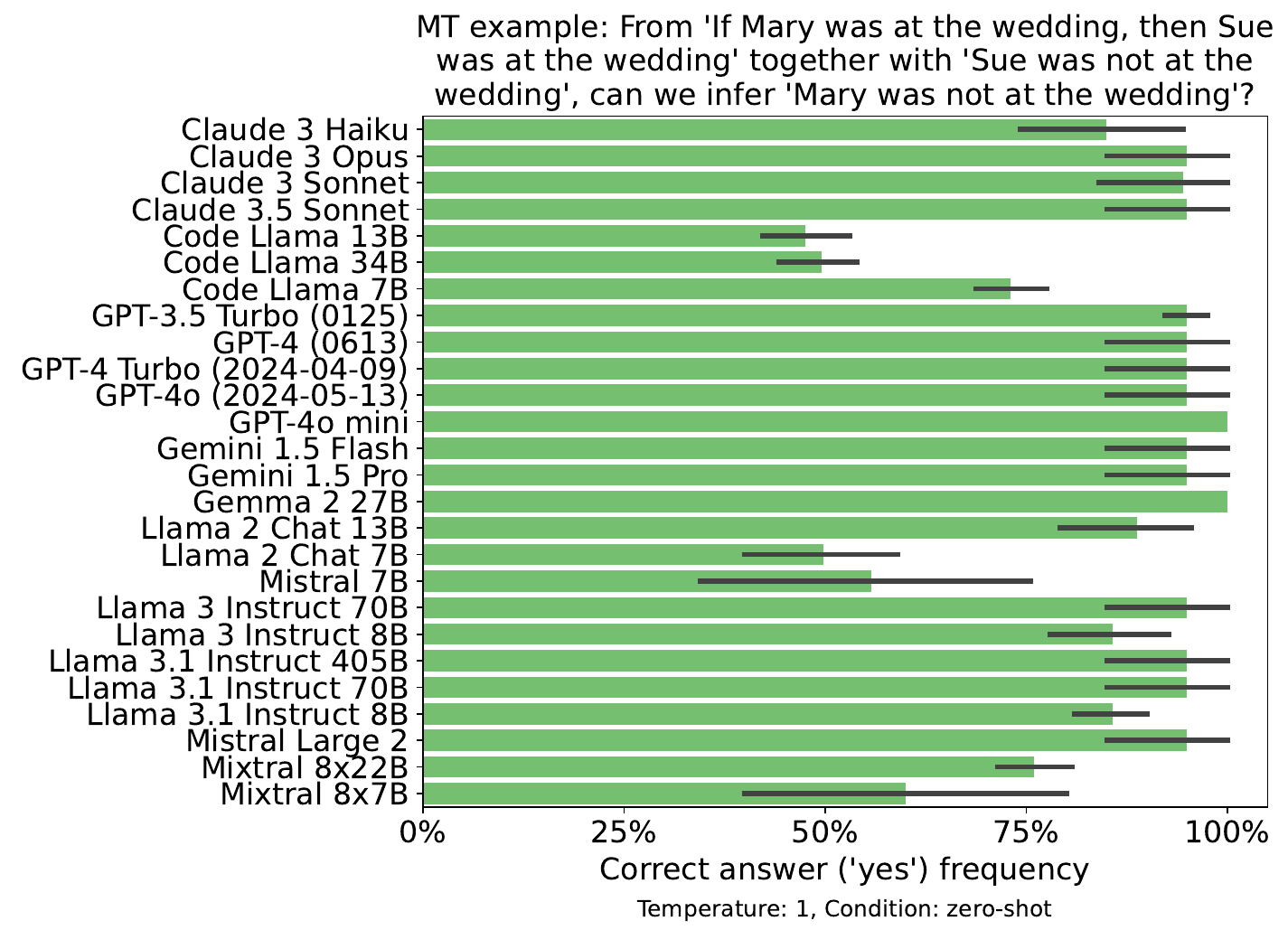}
\end{center}

\subsubsection{Affirming the Consequent (AC)}

\begin{center}
\includegraphics[scale=.3]{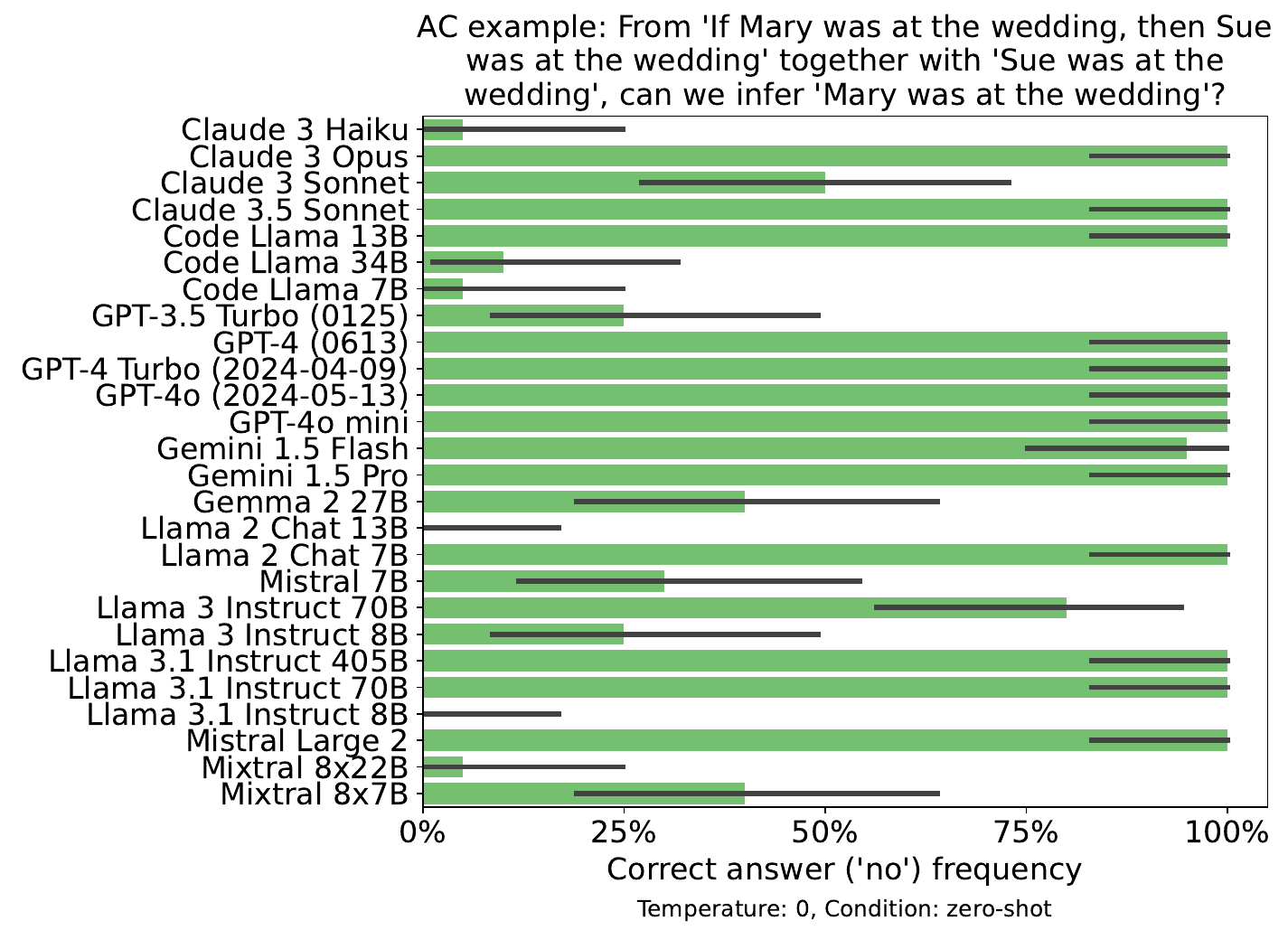}
\includegraphics[scale=.3]{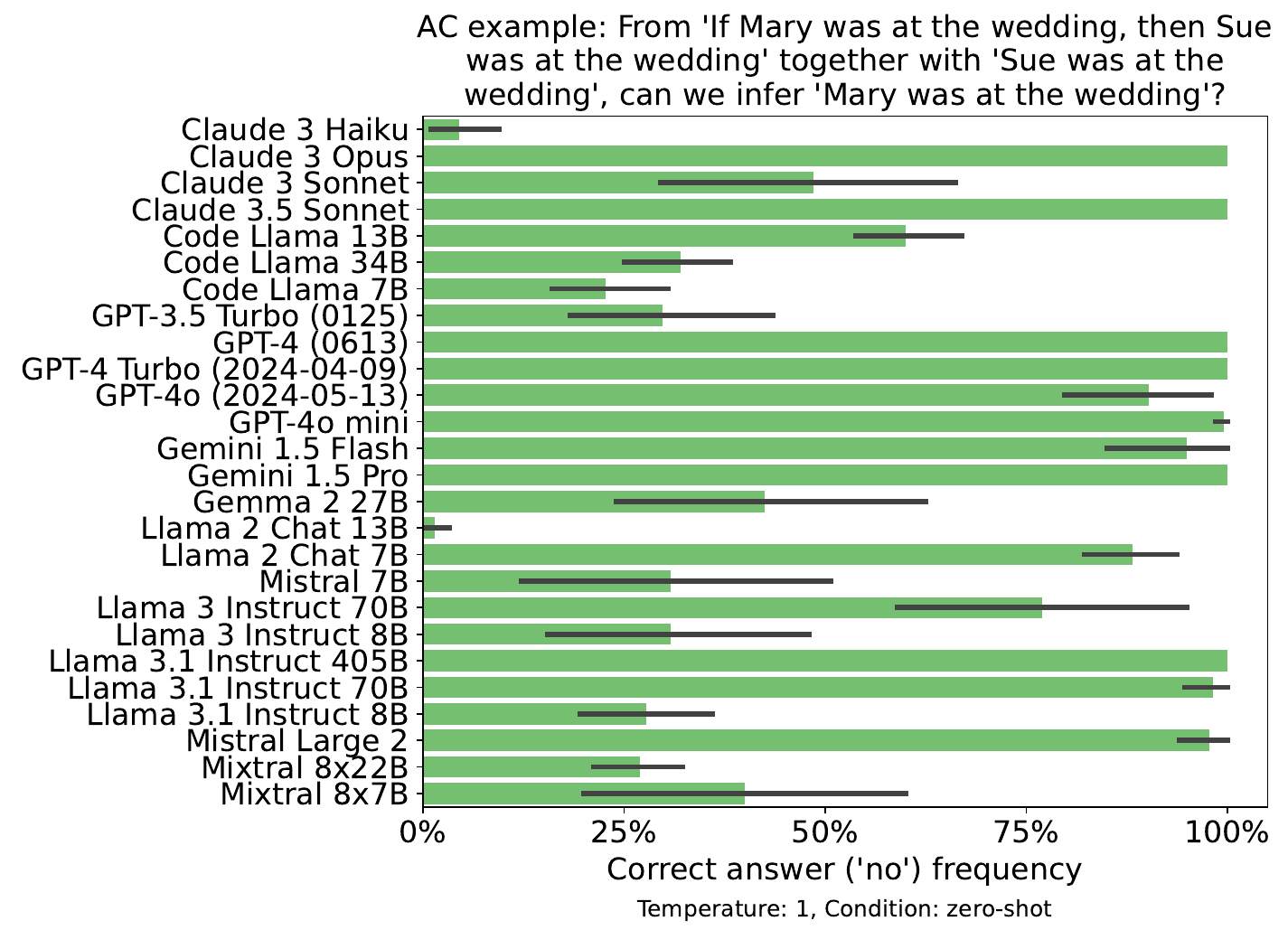}
\end{center}

\subsubsection{Conversion {CONV)}}

\begin{center}
\includegraphics[scale=.3]{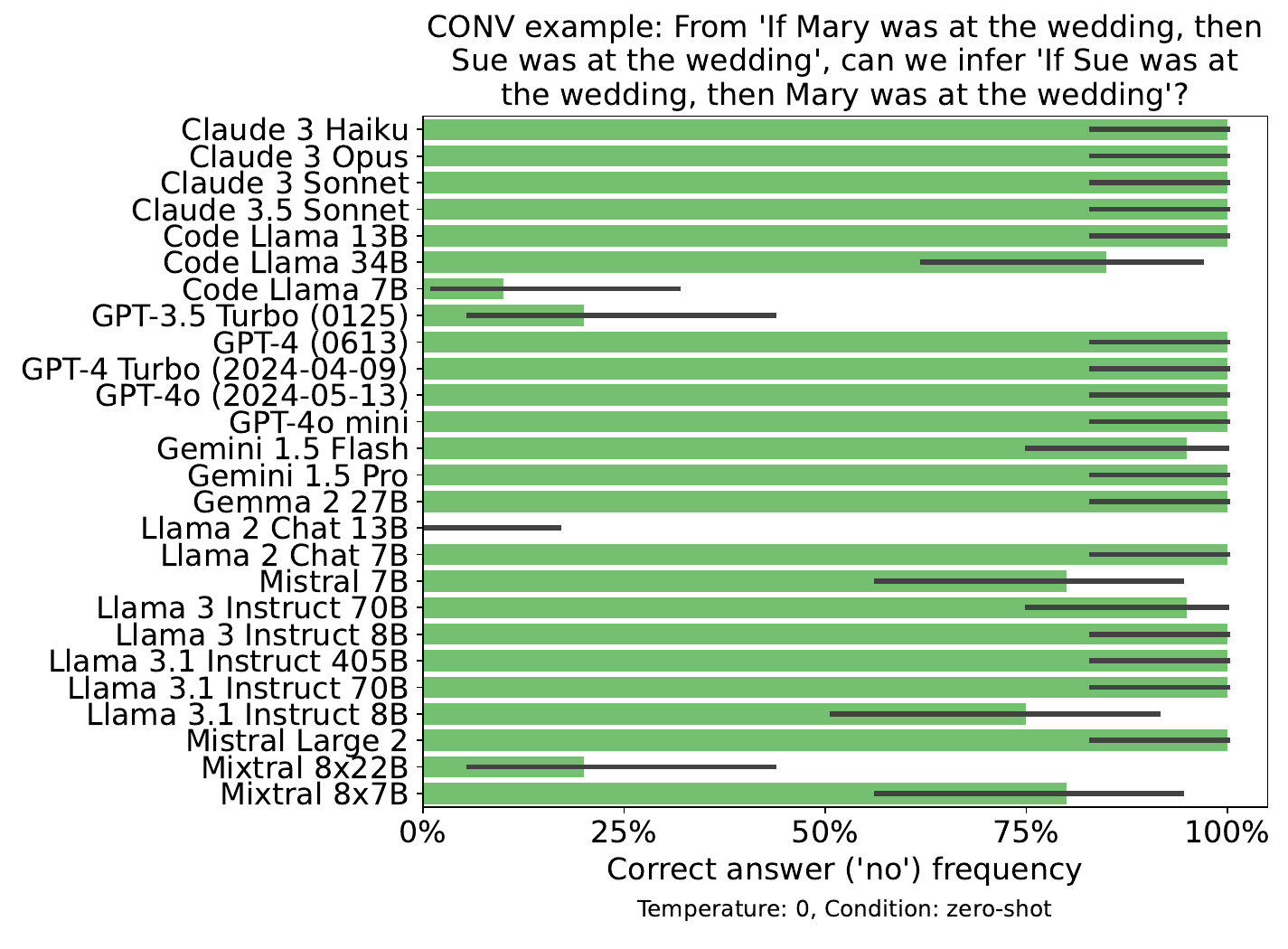}
\includegraphics[scale=.3]{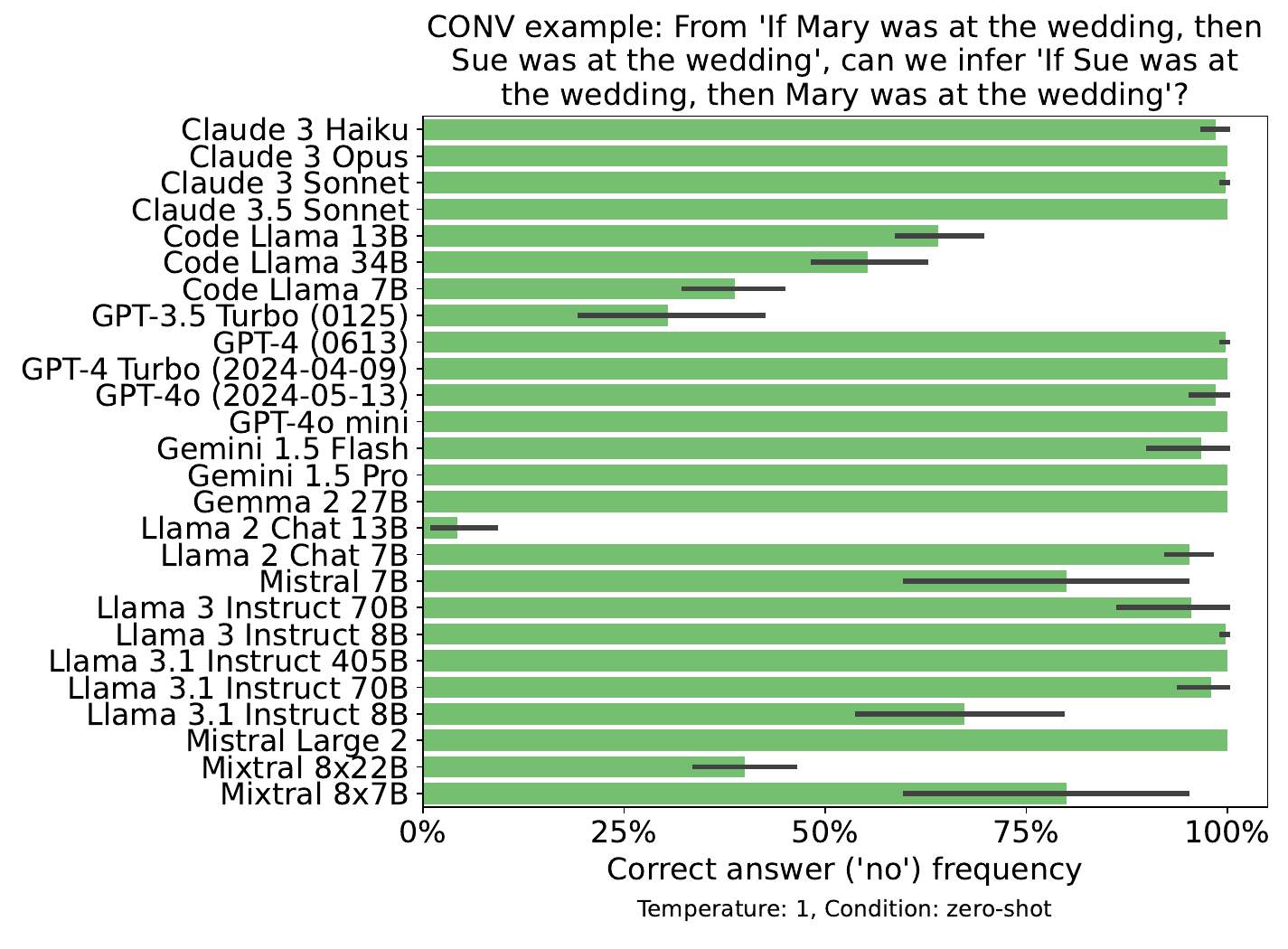}
\end{center}

\subsubsection{Denying the Antecedent (DA)}

\begin{center}
\includegraphics[scale=.3]{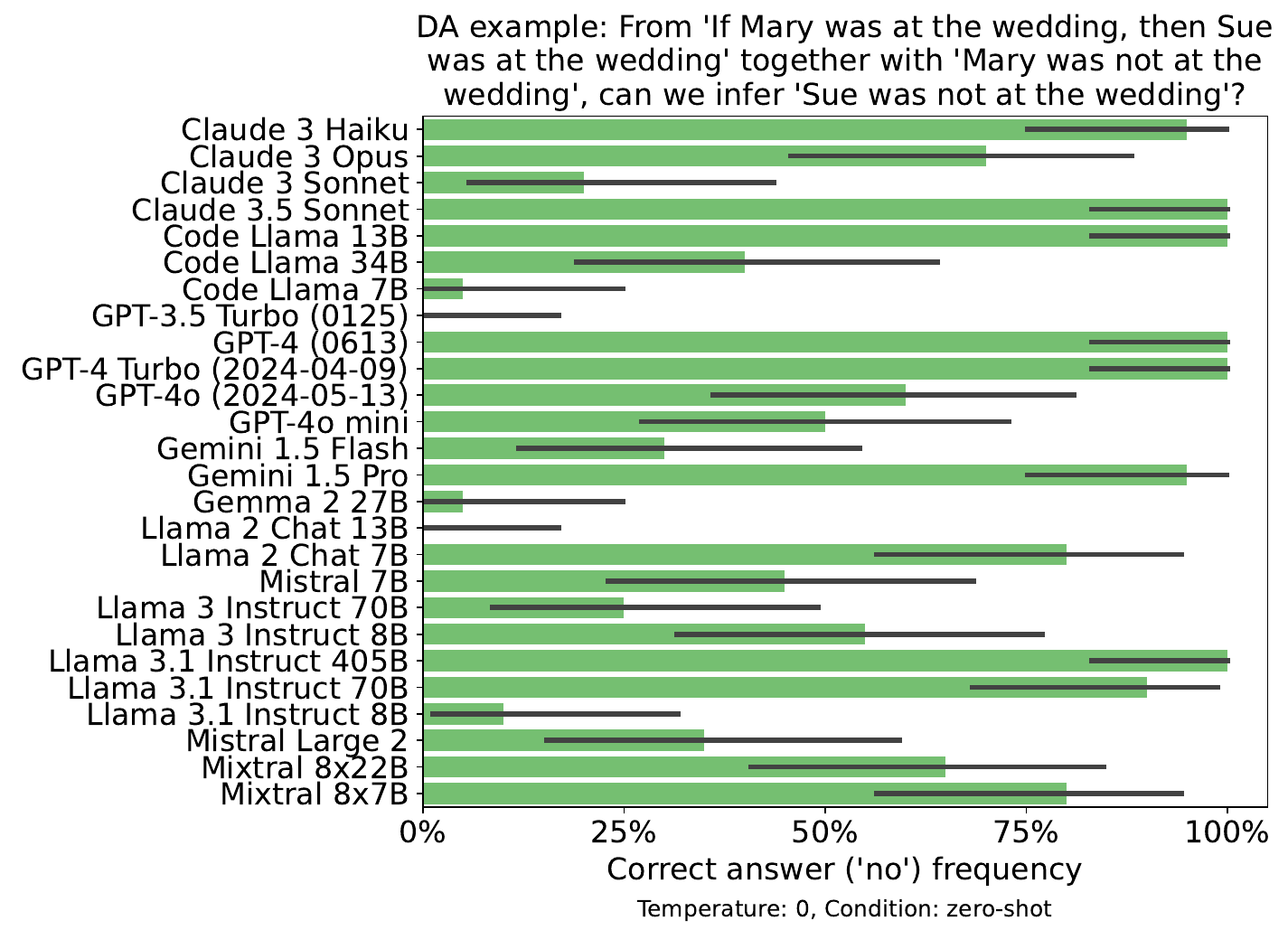}
\includegraphics[scale=.3]{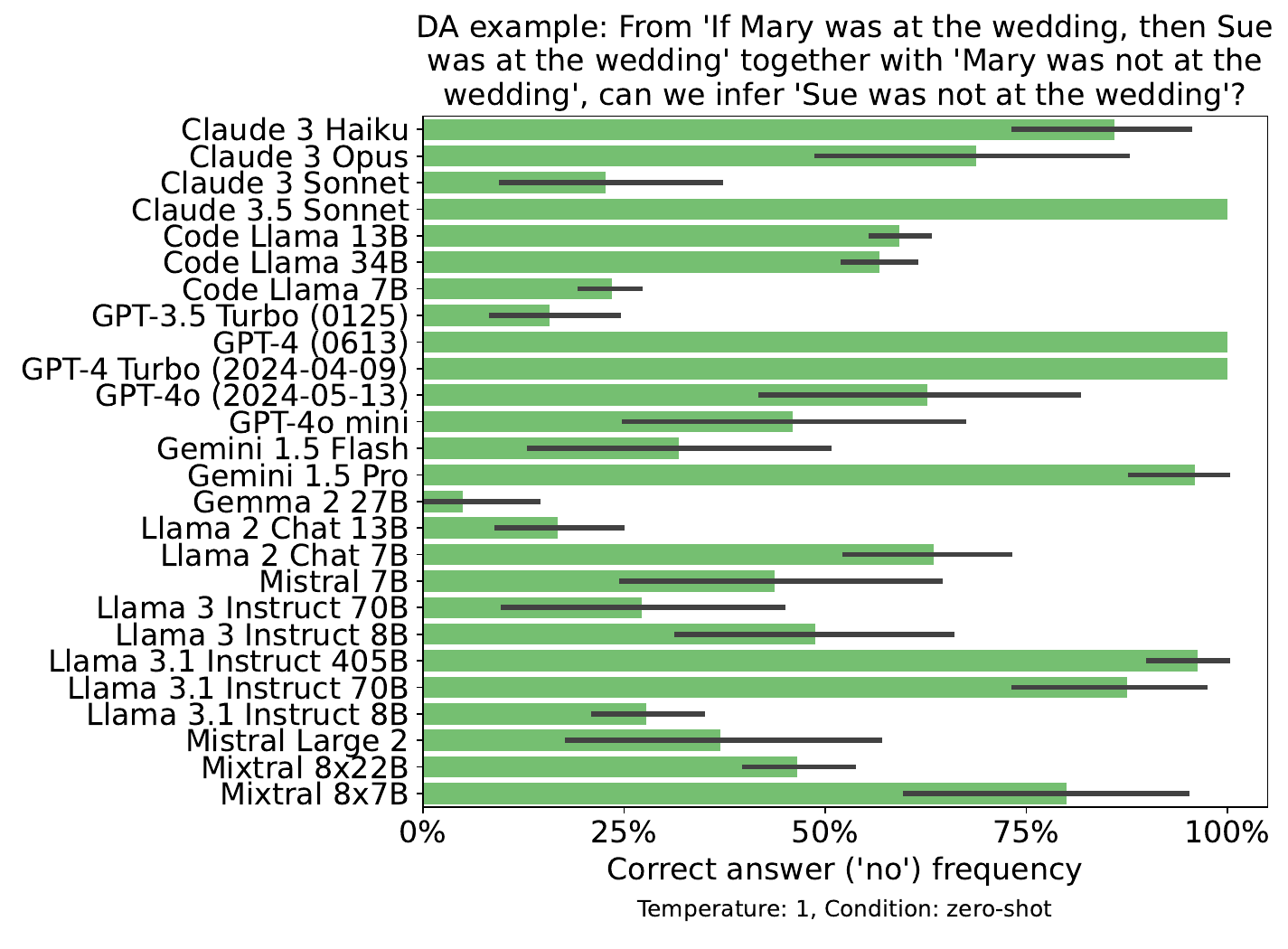}
\end{center}

\subsubsection{Inversion (INV)}

\begin{center}
\includegraphics[scale=.3]{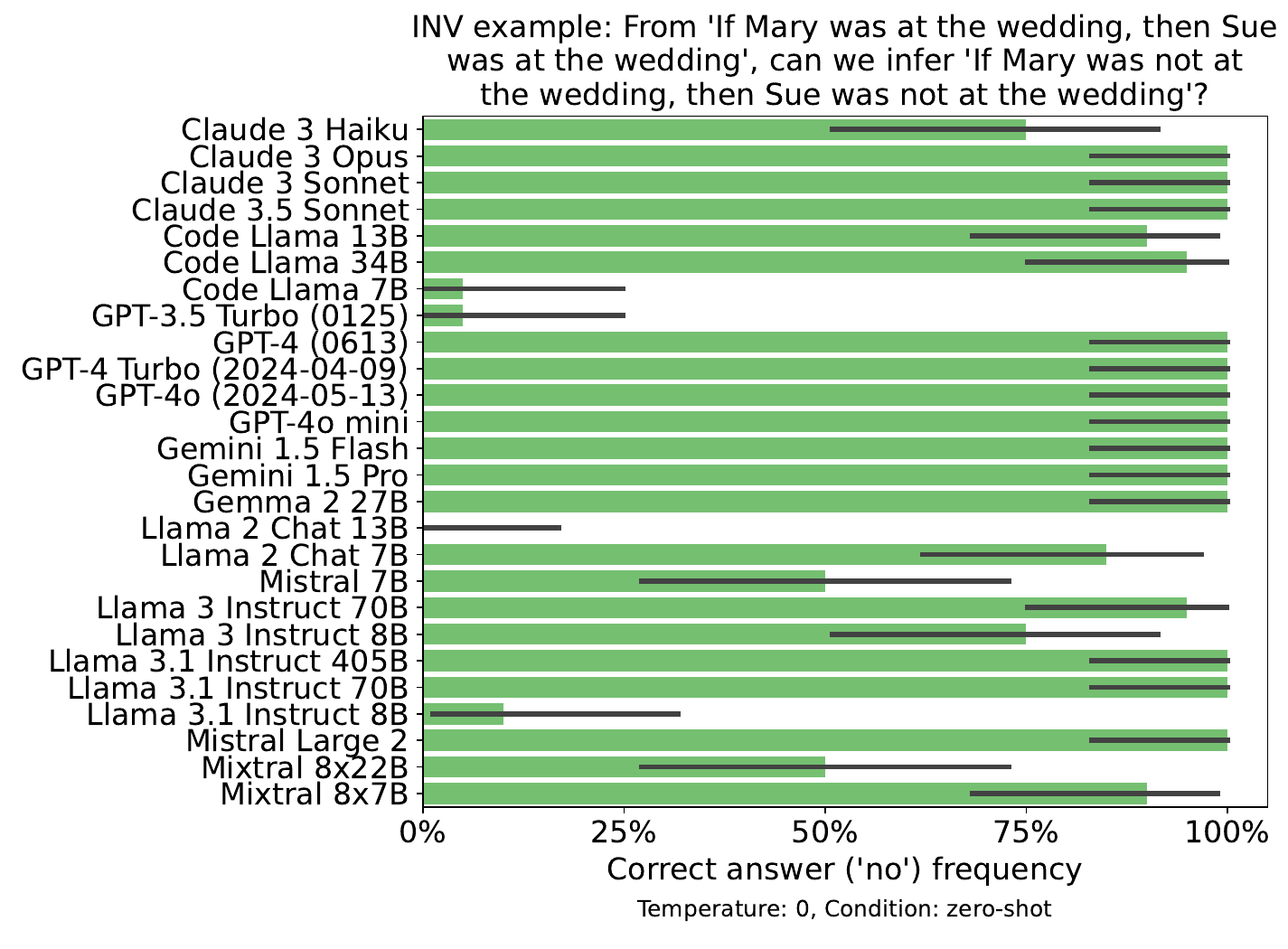}
\includegraphics[scale=.3]{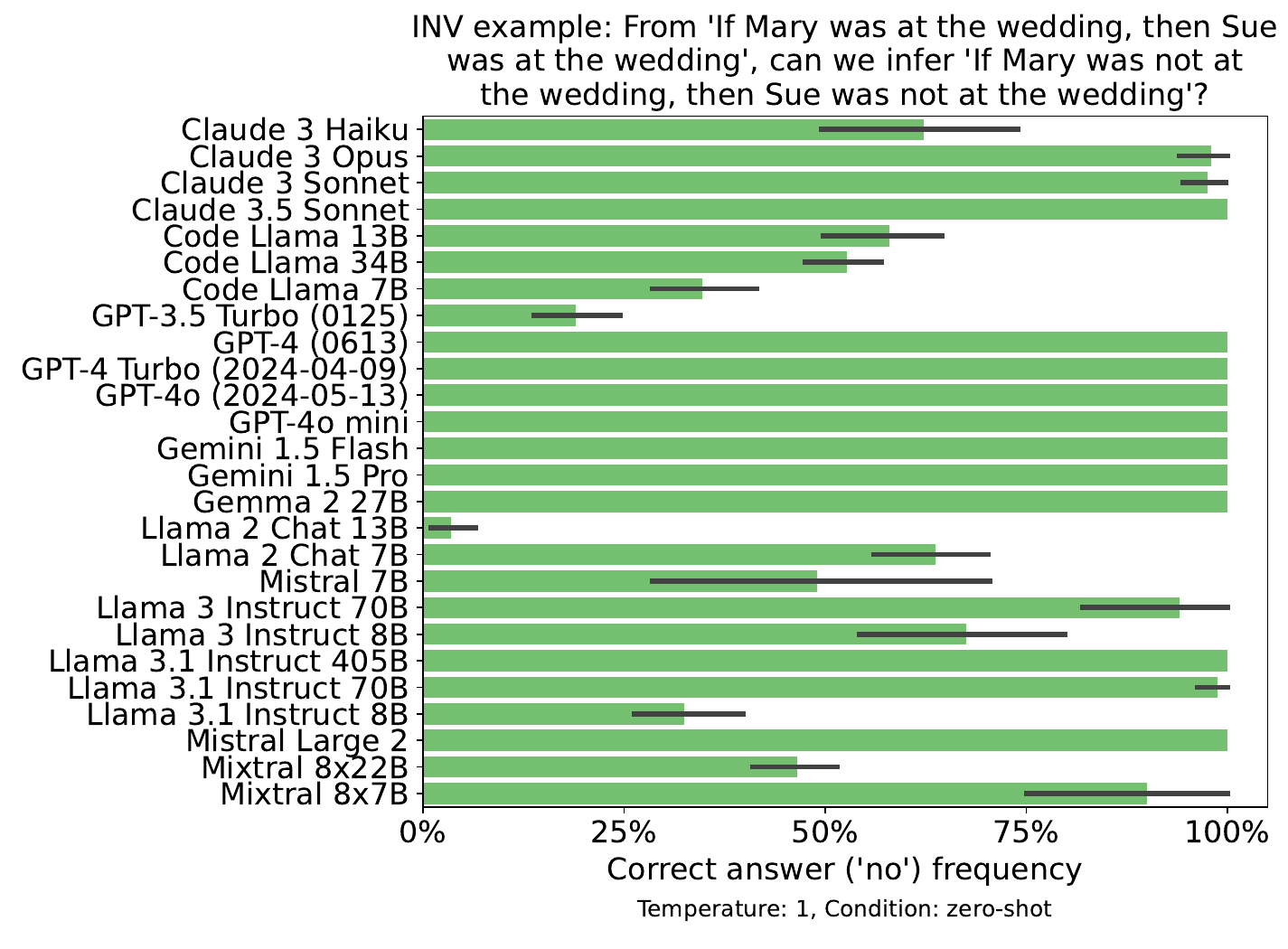}
\end{center}

\subsubsection{Might Not (MiN)}

\begin{center}
\includegraphics[scale=.3]{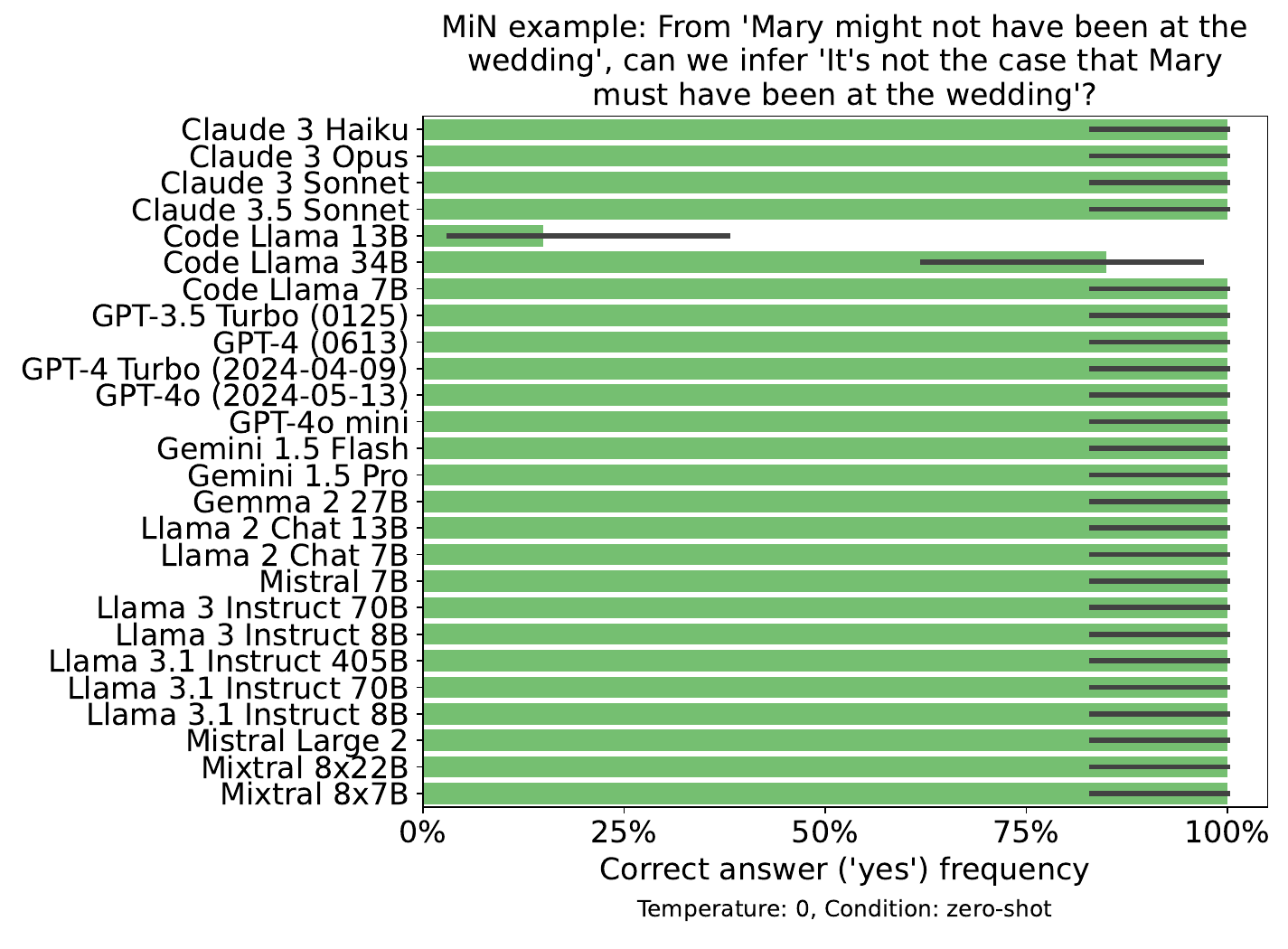}
\includegraphics[scale=.3]{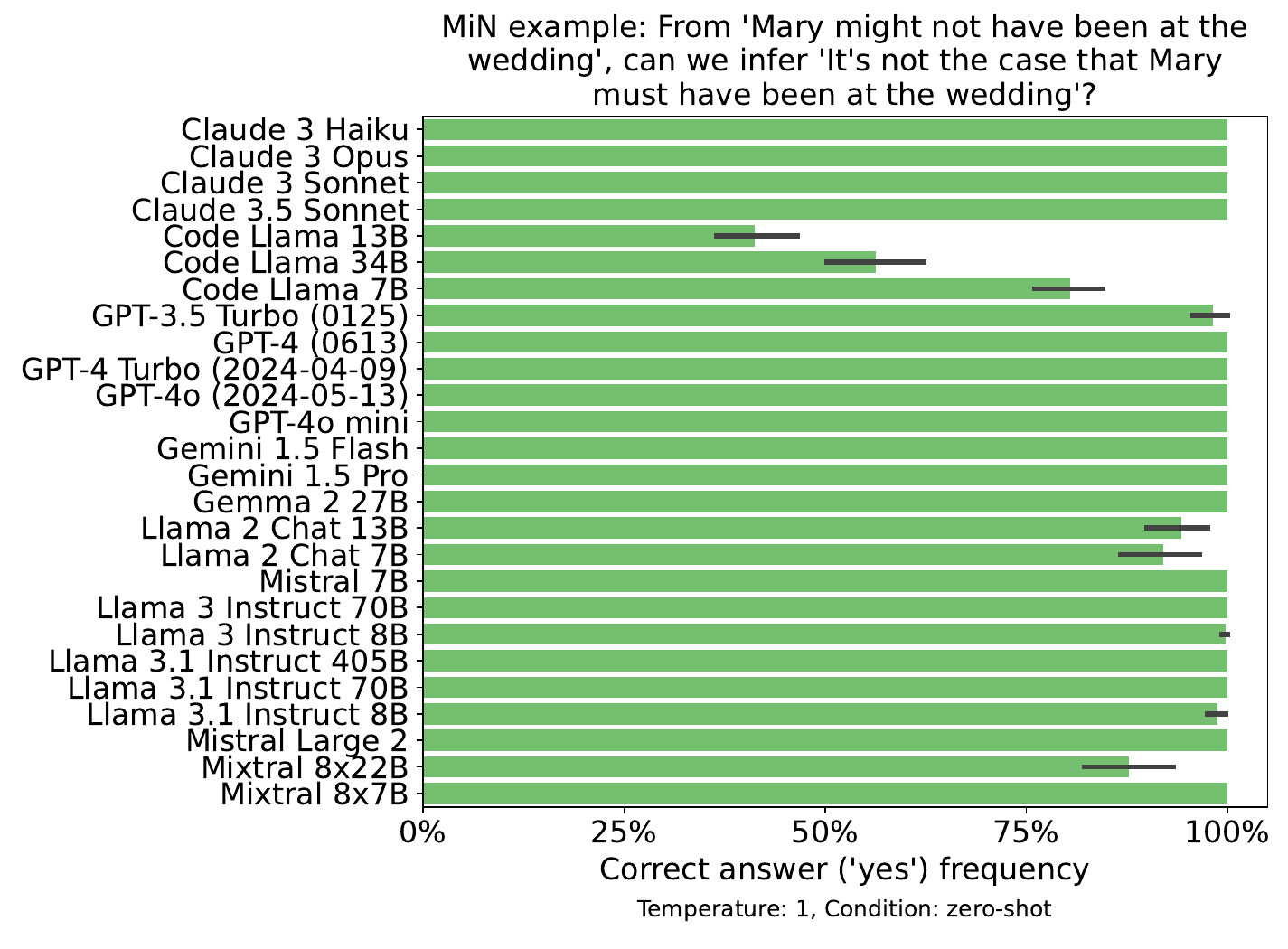}
\end{center}

\subsubsection{Not Must (NMu)}

\begin{center}
\includegraphics[scale=.3]{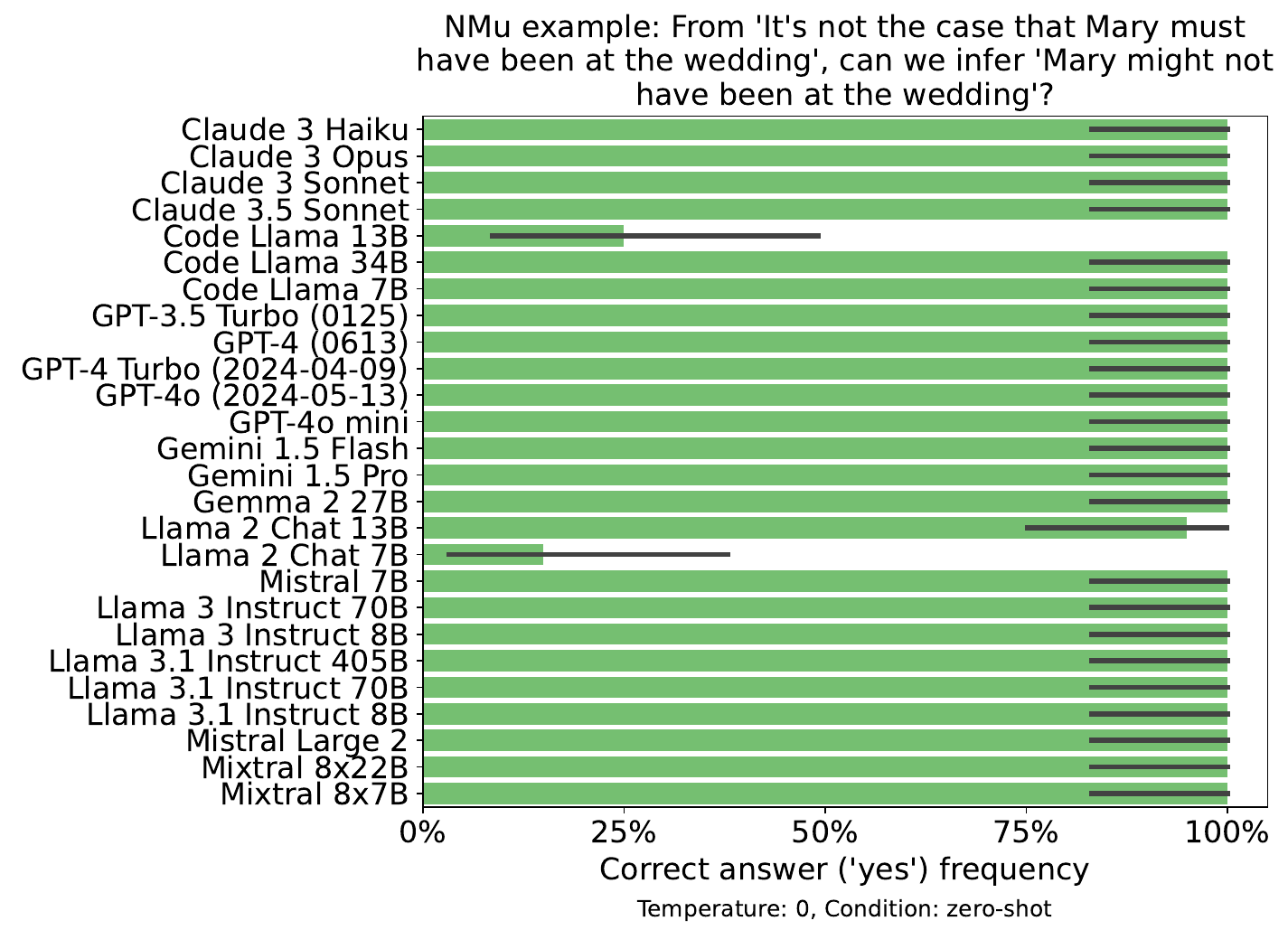}
\includegraphics[scale=.3]{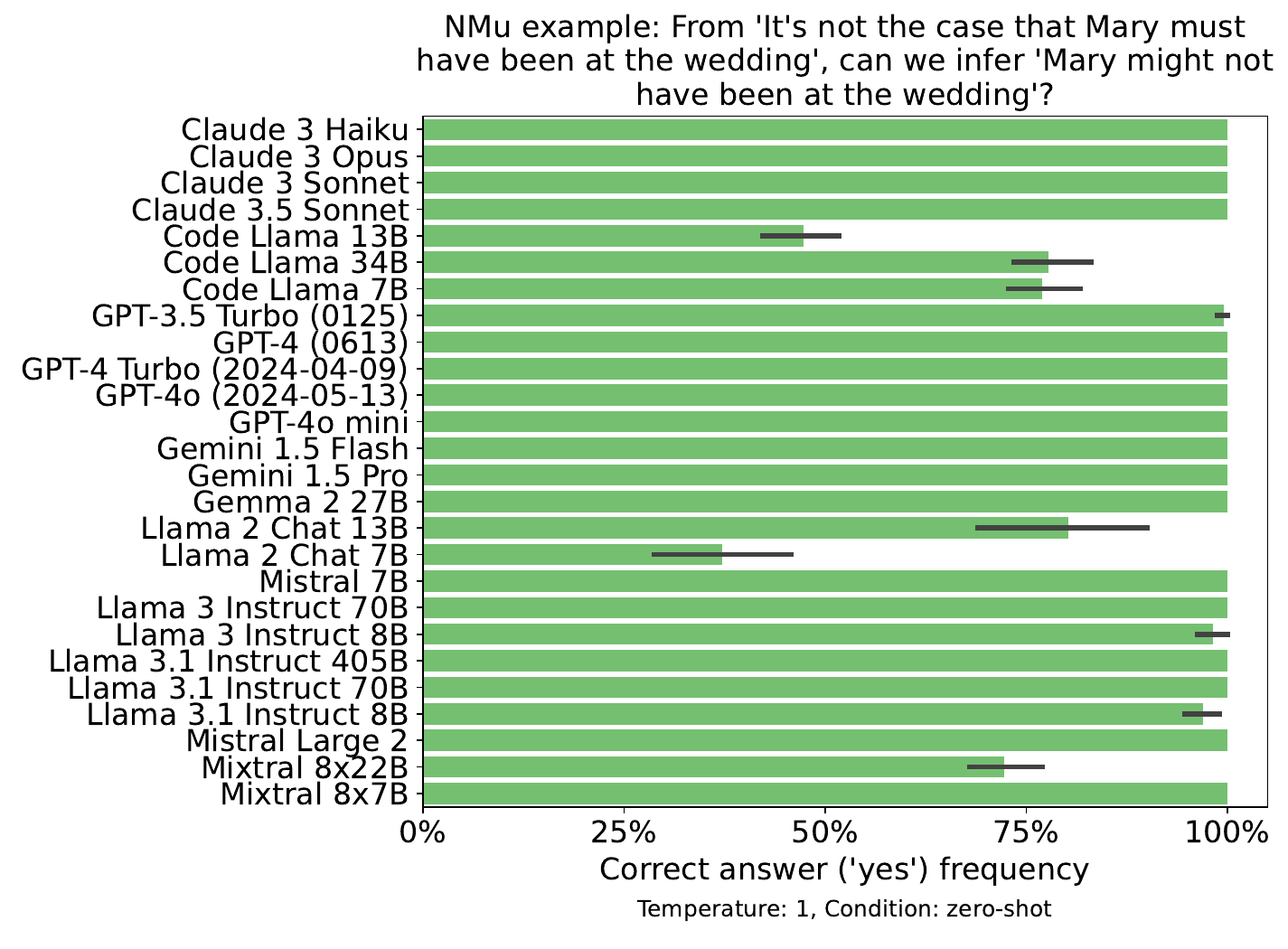}
\end{center}

\newpage

\subsubsection{Antecedent Strengthening (AS)}\label{ASappendix}

In the graphs below, the `d' in `ASd' indicates \textit{deviant} instances, designed to bring out the invalidity of antecedent strengthening as in \citealt{Stalnaker1968}.

\begin{center}
\includegraphics[scale=.3]{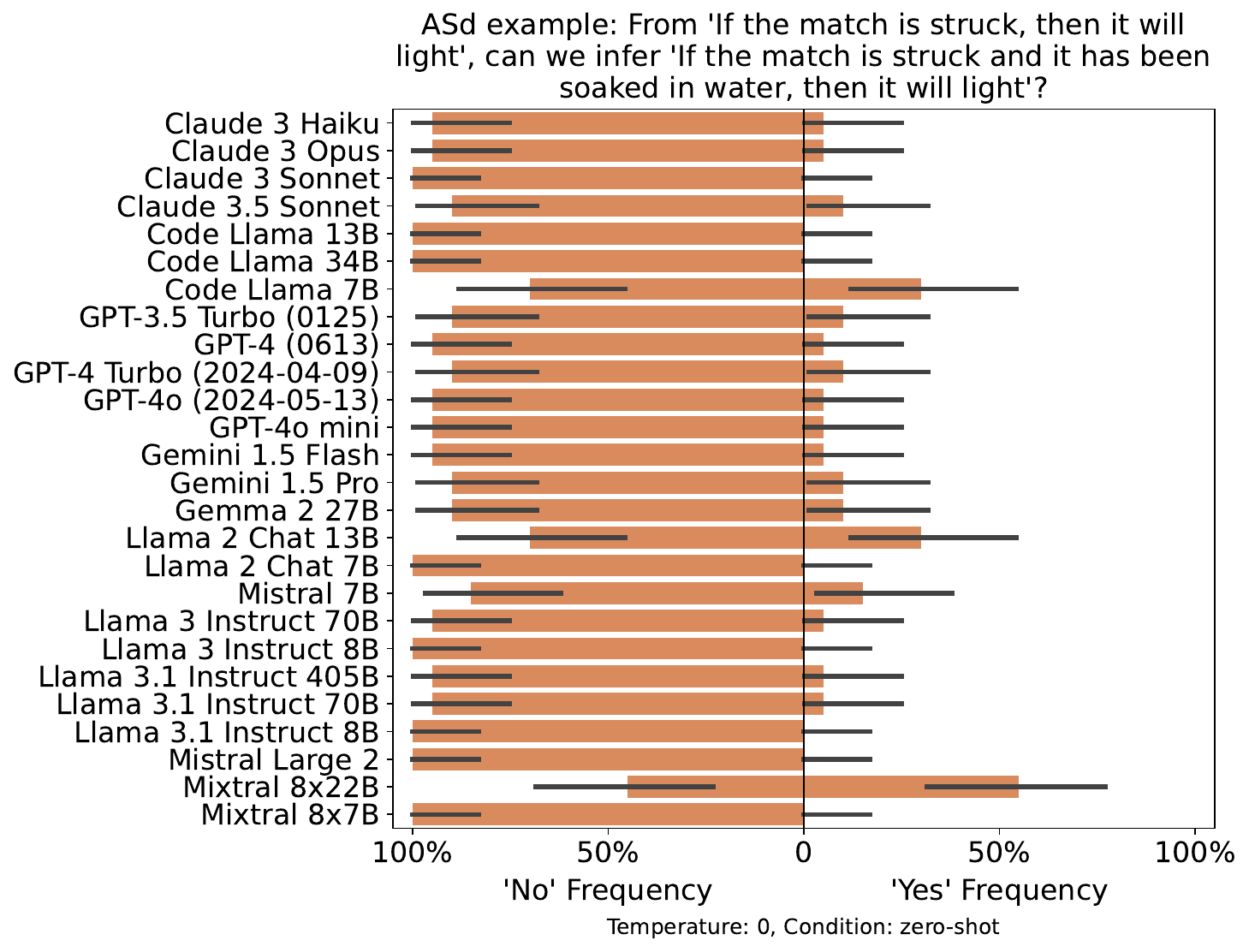}
\includegraphics[scale=.3]{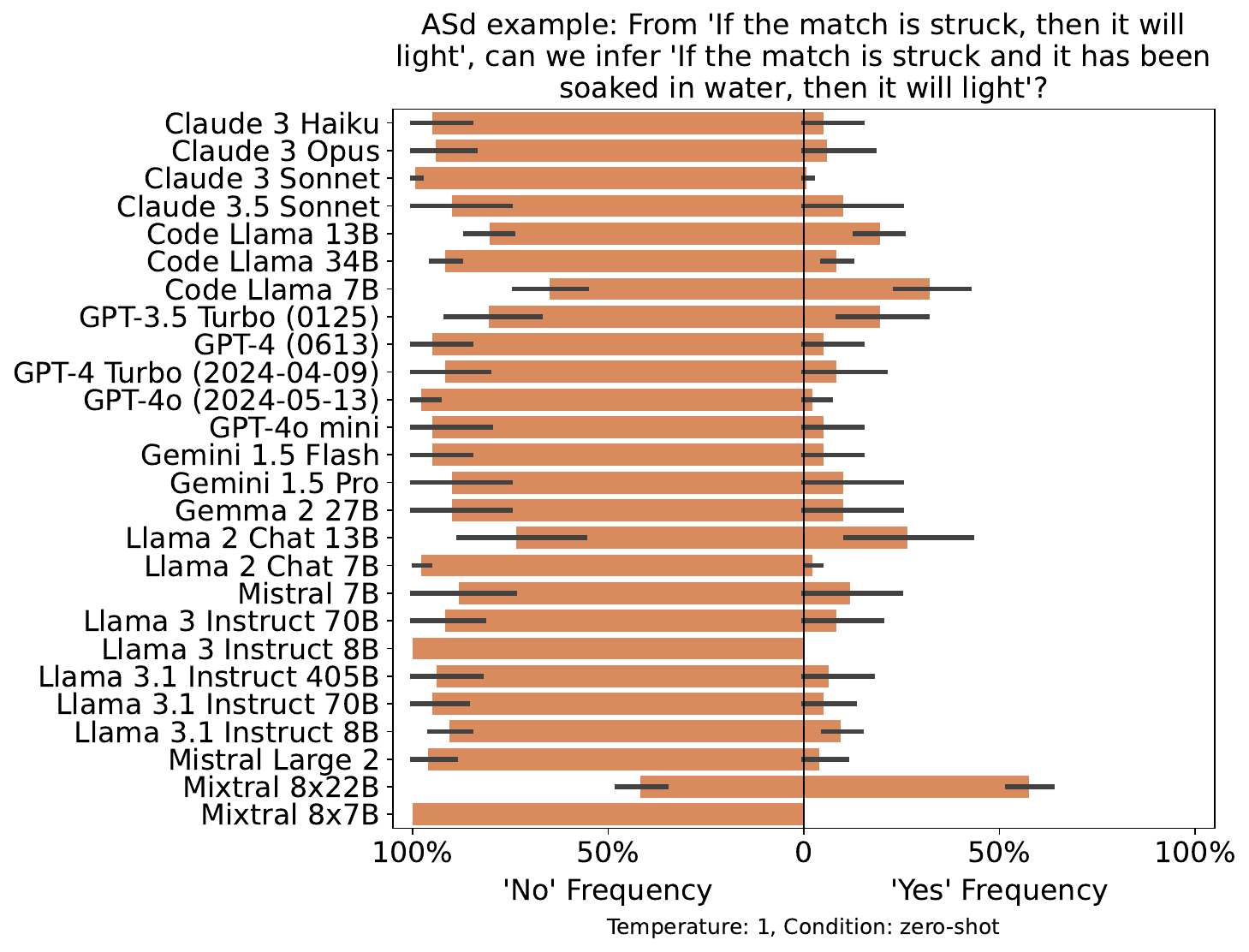}
\end{center}

\newpage

\subsubsection{Contraposition (CT)}\label{CTappendix}

In the graphs below, the `n' in `CTnd' indicates that we use a version of contraposition with negation in the premise ($p\to \neg q\vdash q\to\neg p$), while the `d' again indicates \textit{deviant} instances, designed to bring out the invalidity of CT as in \citealt{Stalnaker1968}.

\begin{center}
\includegraphics[scale=.3]{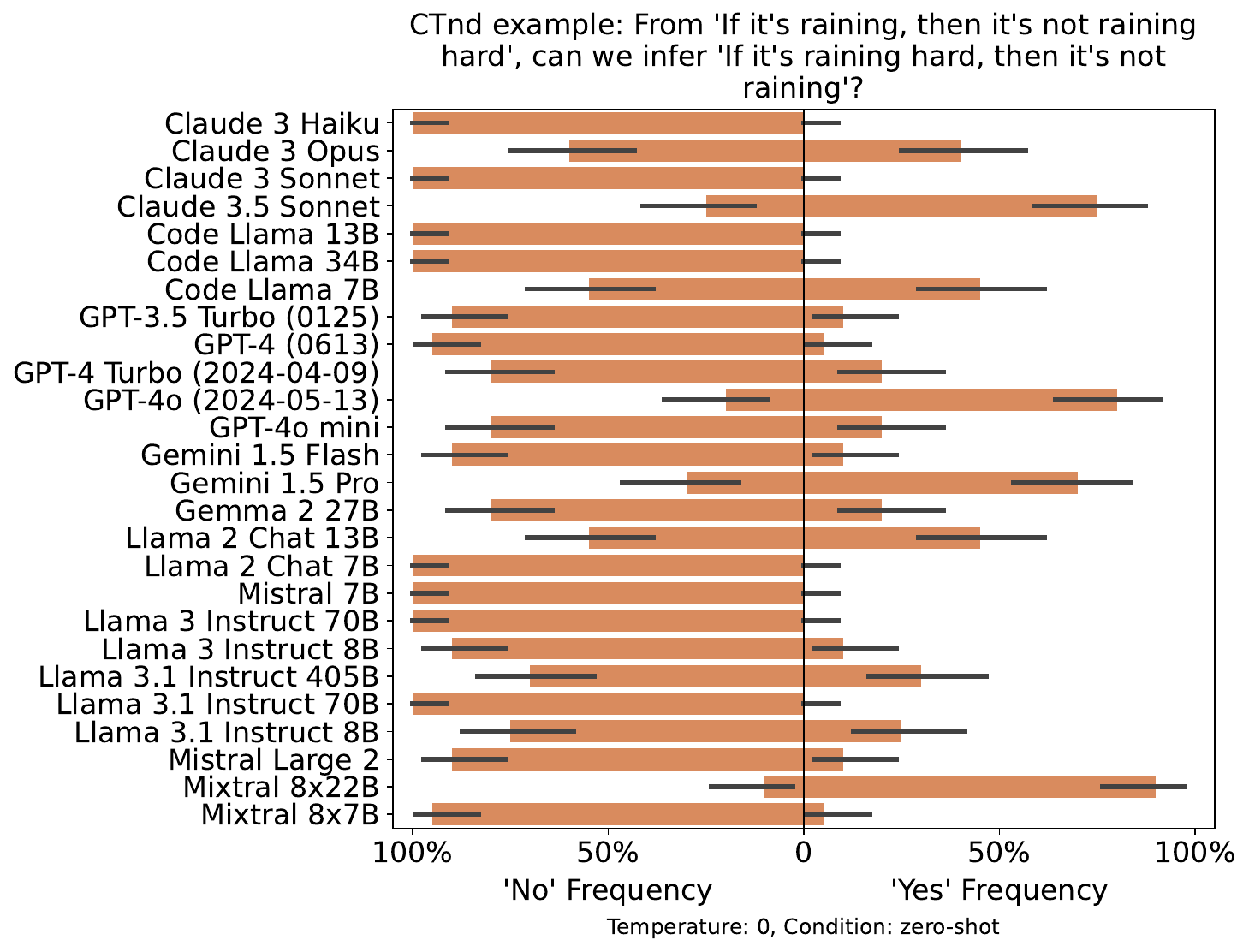}
\includegraphics[scale=.3]{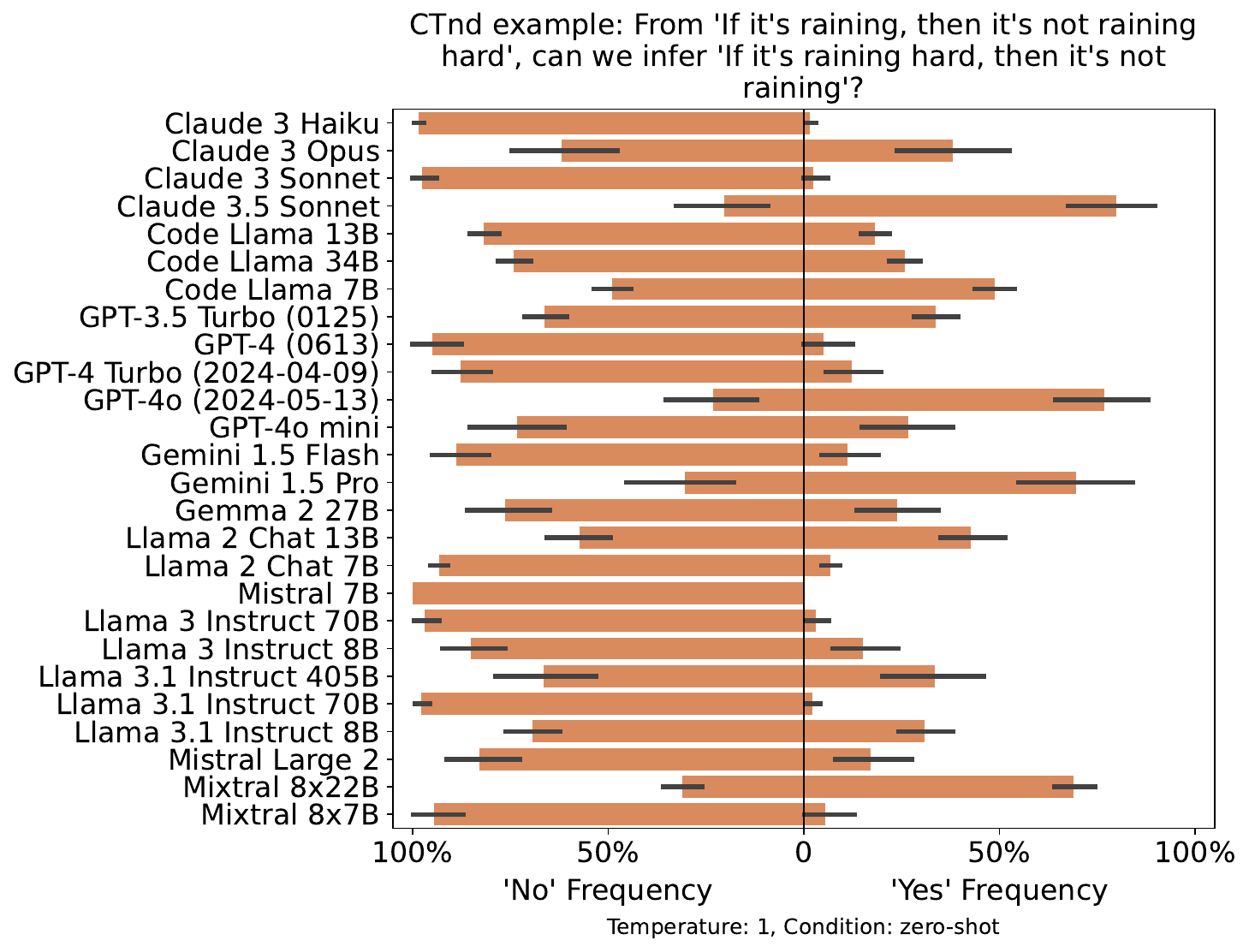}
\end{center}

\newpage

\subsubsection{DS with `must' (DSmu)}

\begin{center}
\includegraphics[scale=.3]{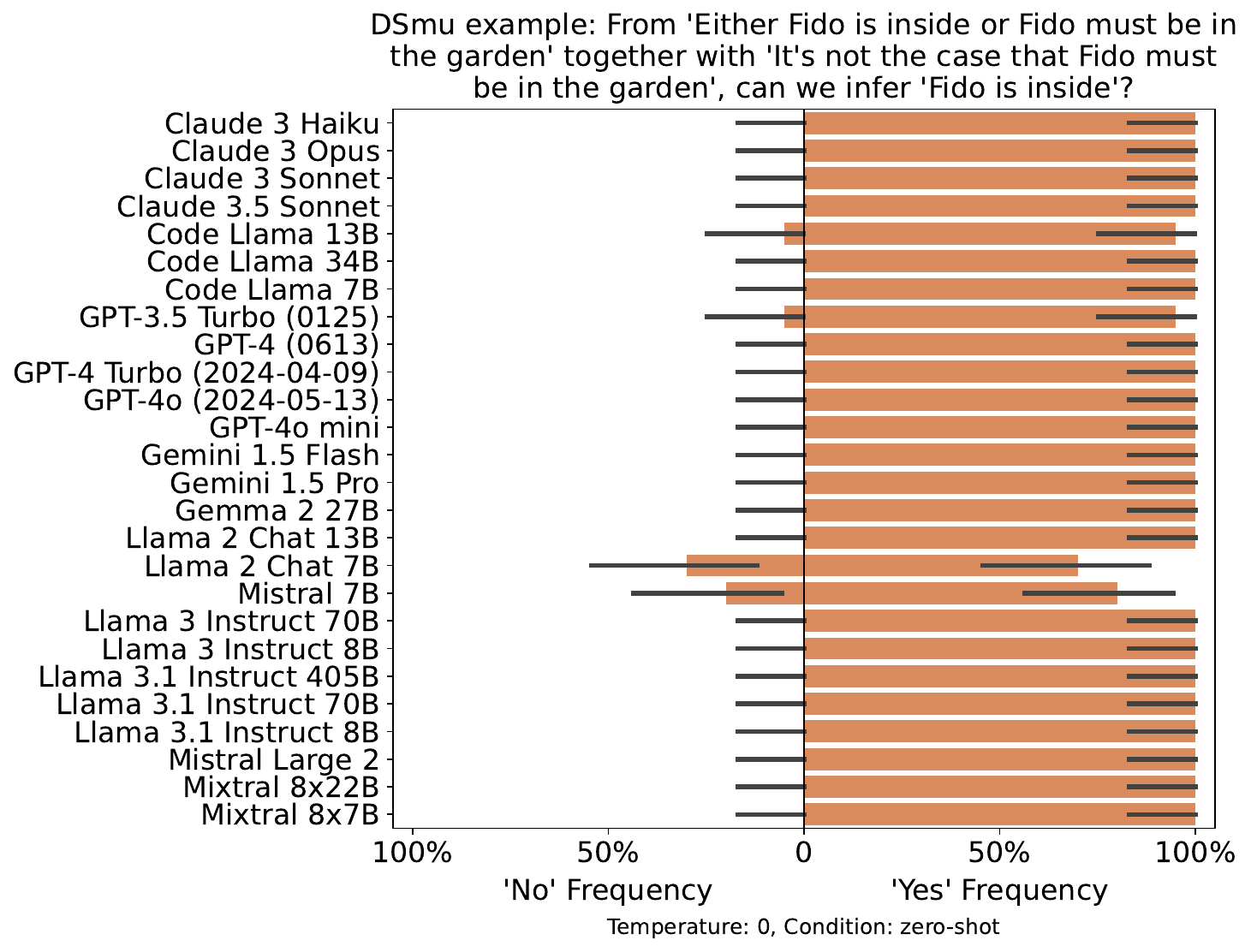}
\includegraphics[scale=.3]{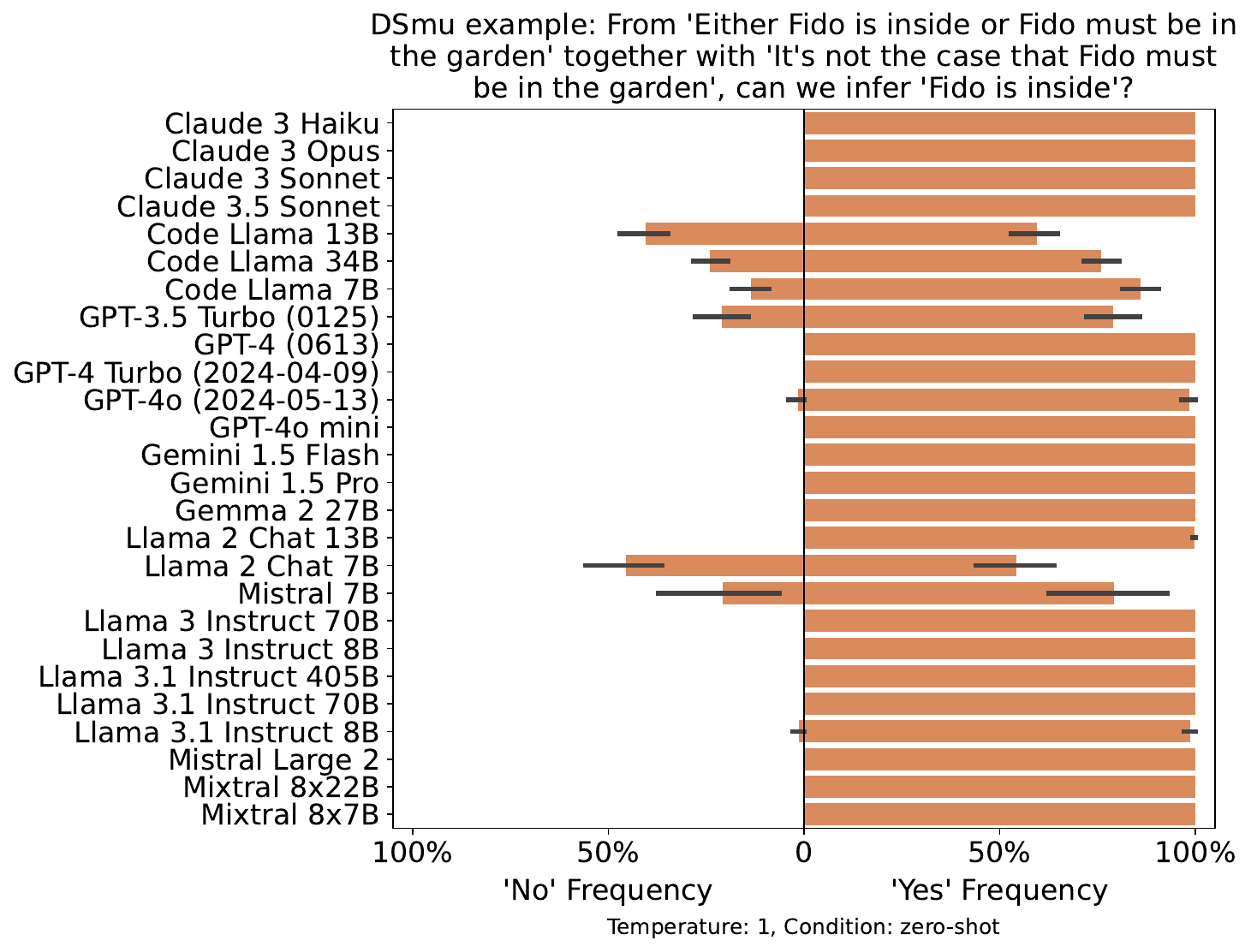}
\end{center}

\newpage

\subsubsection{DS with `might' (DSmi)}

\begin{center}
\includegraphics[scale=.3]{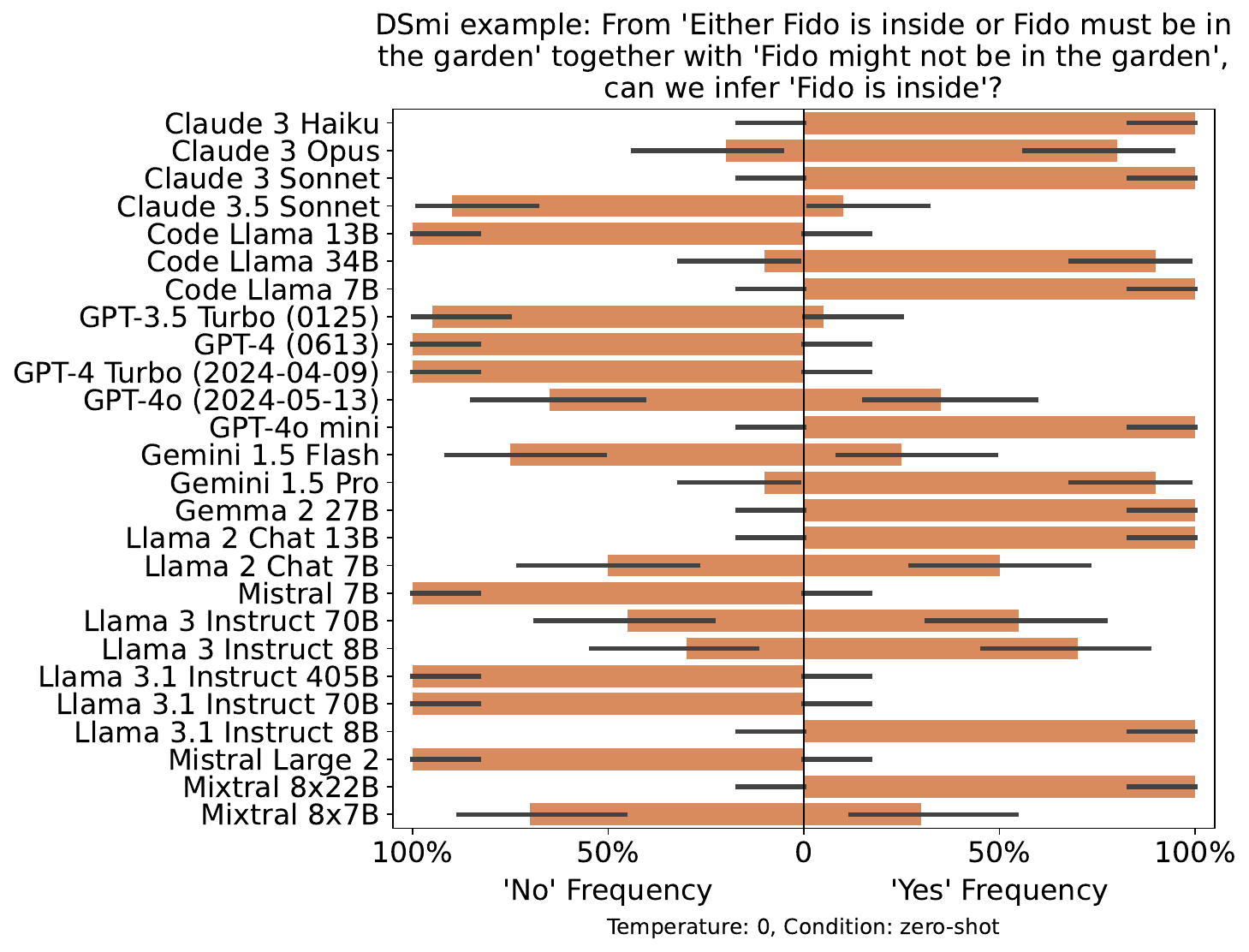}
\includegraphics[scale=.3]{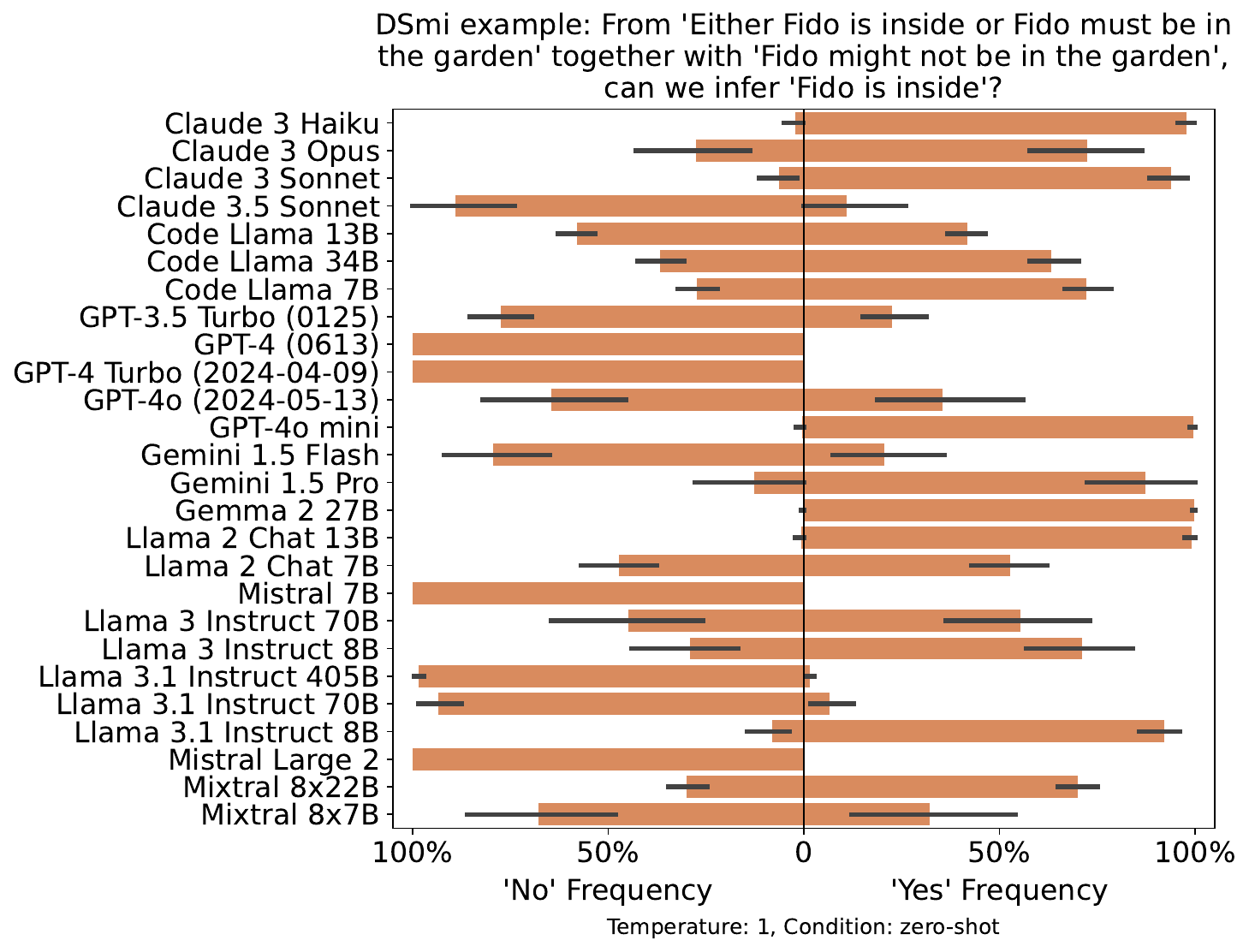}
\end{center}

\newpage

\subsubsection{MT with `must' (MTmu)}

\begin{center}
\includegraphics[scale=.3]{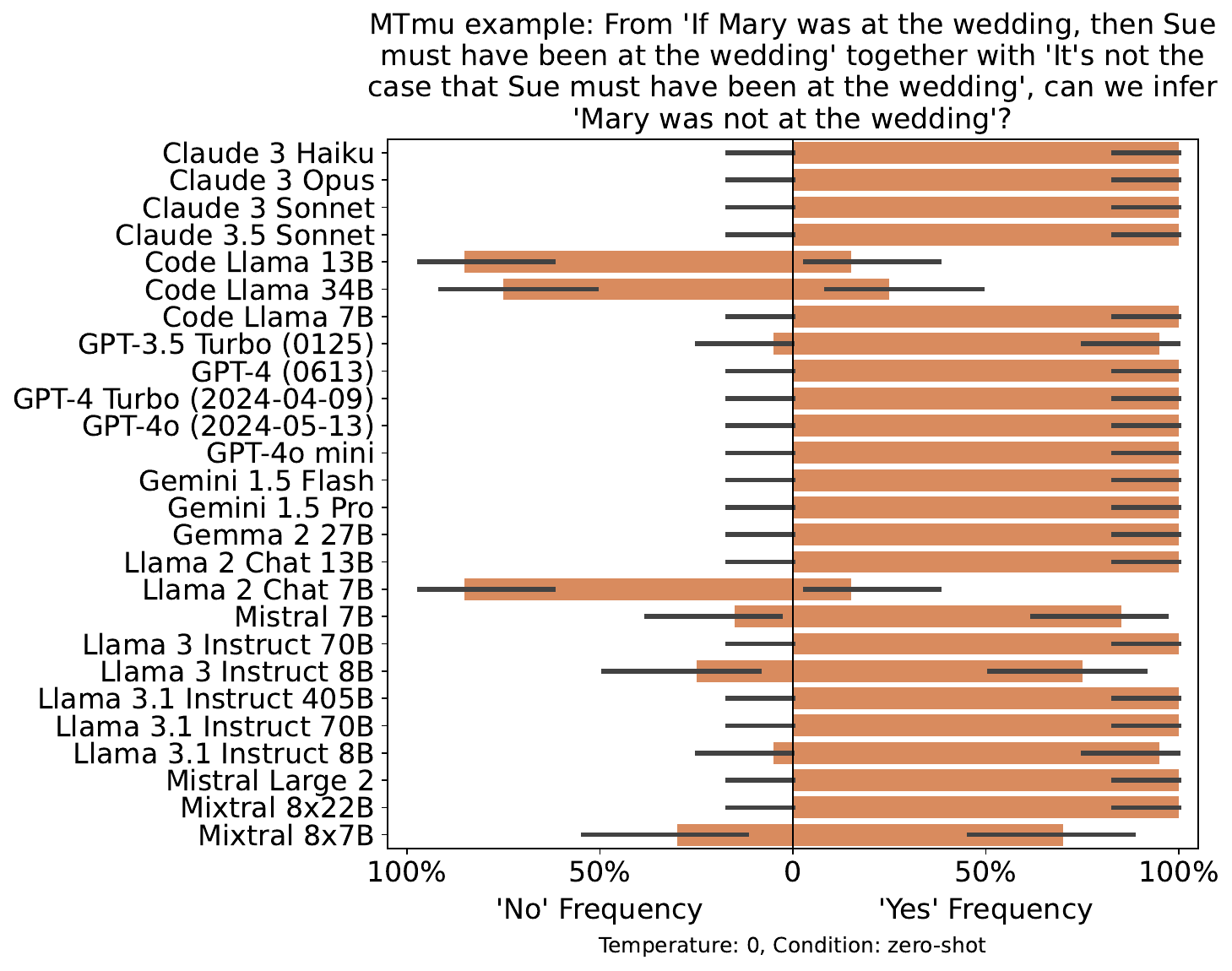}
\includegraphics[scale=.3]{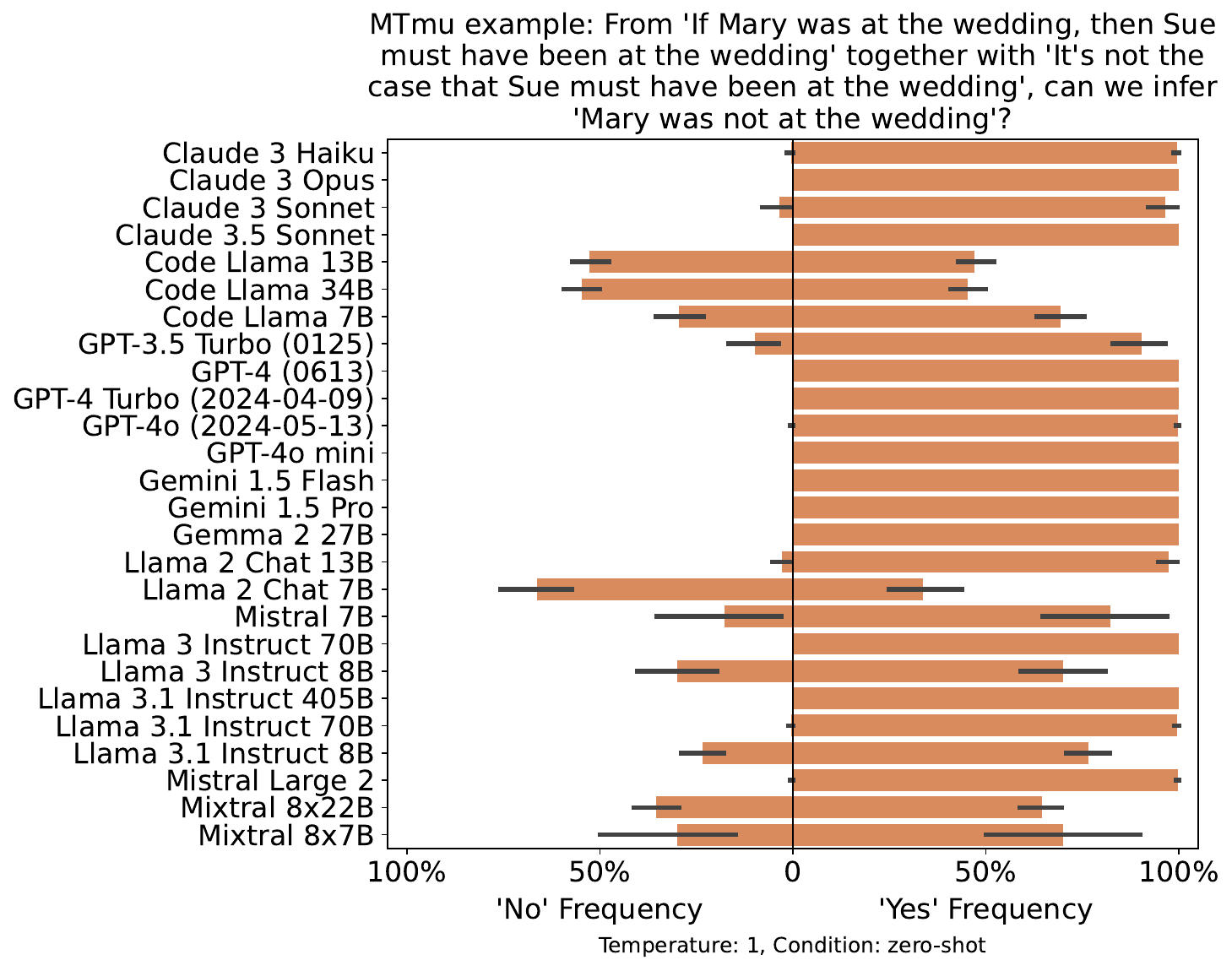}
\end{center}

\newpage

\subsubsection{MT with `might' (MTmi)}

\begin{center}

\includegraphics[scale=.3]{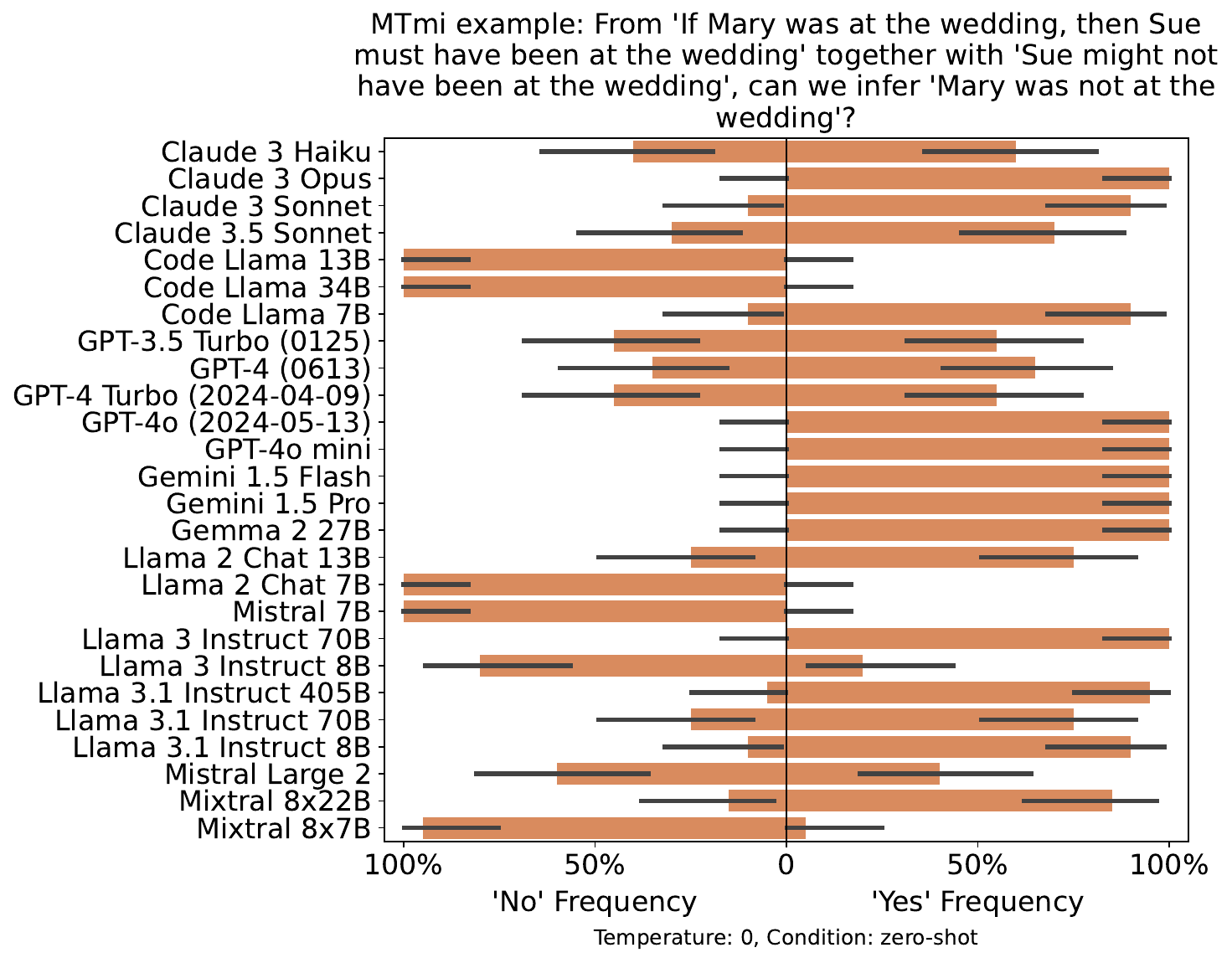}
\includegraphics[scale=.3]{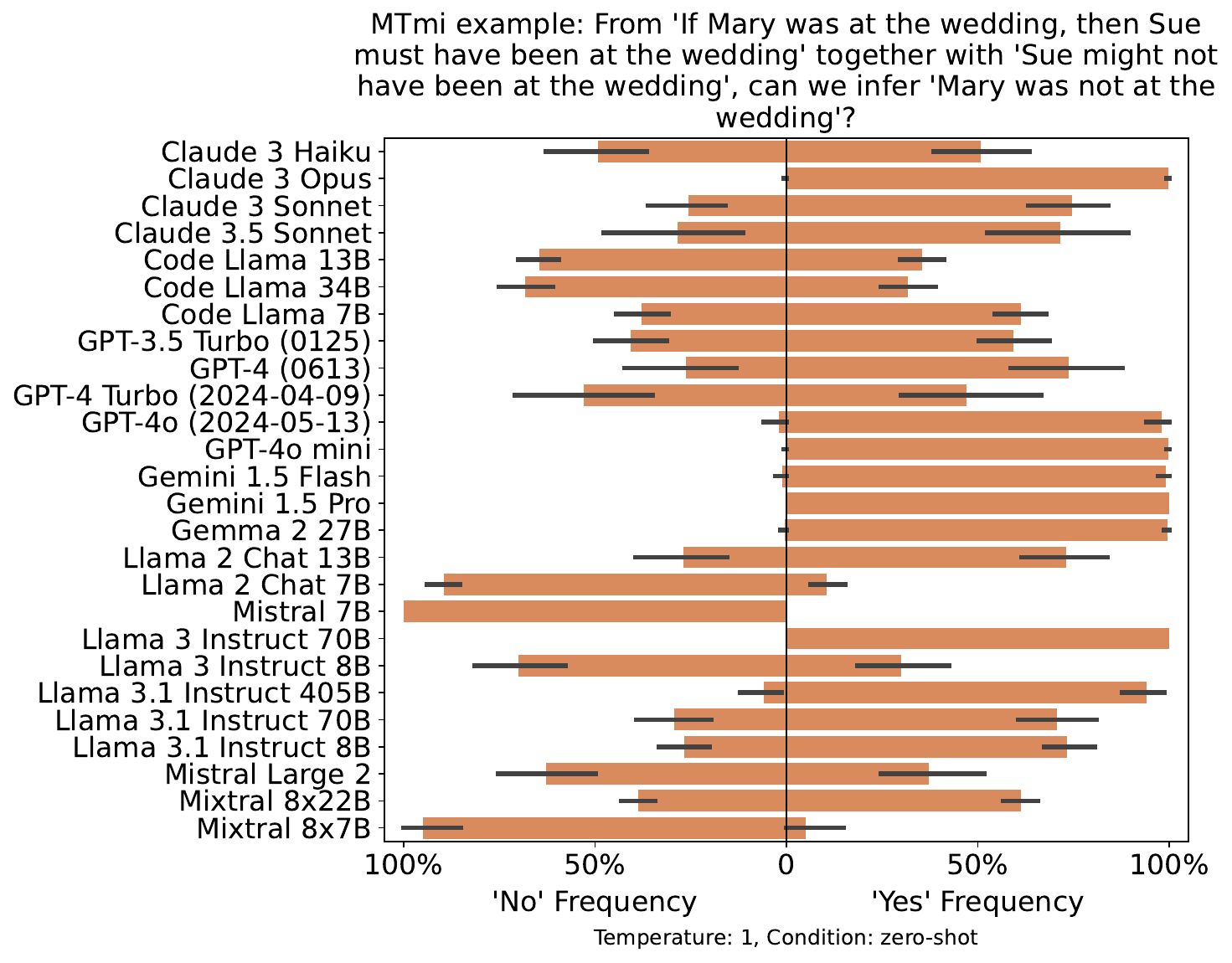}
\end{center}

\newpage

\subsubsection{Complex Modus Ponens (CMP)}

\begin{center}
\includegraphics[scale=.3]{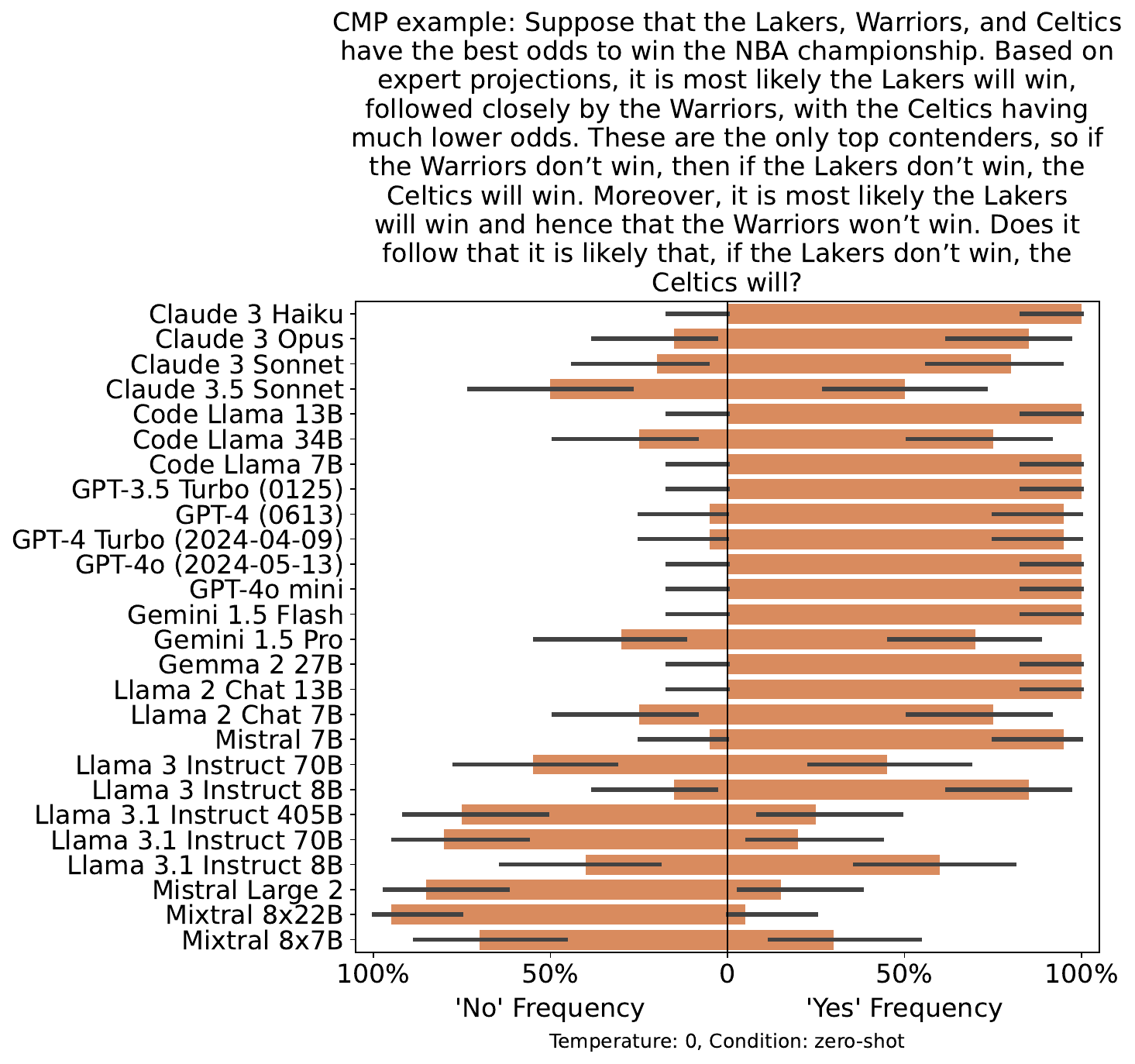}
\includegraphics[scale=.3]{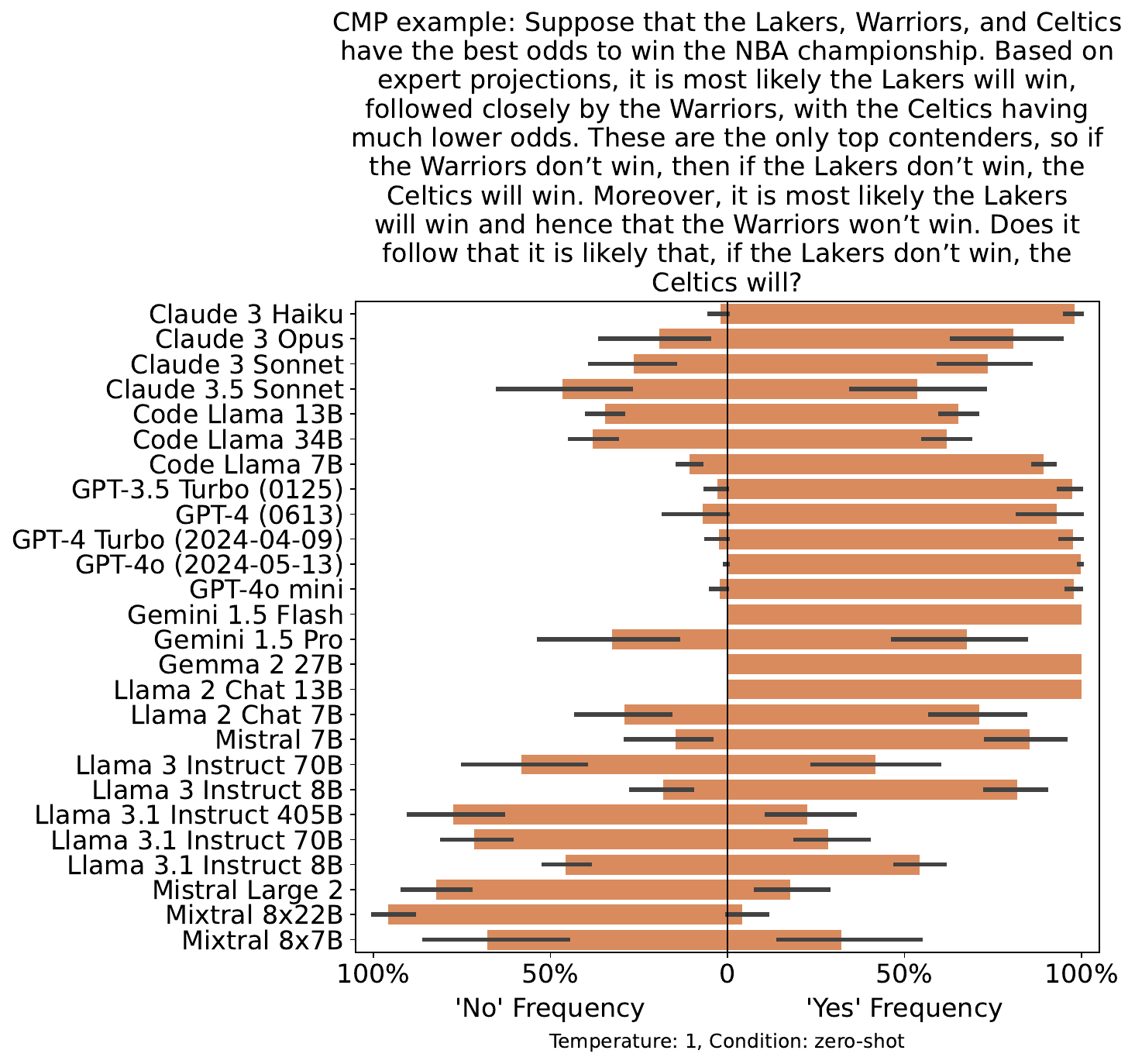}
\end{center}

\newpage

$\,$

\newpage

\subsubsection{`Must' distribution over `or' (MuDistOr)}\label{MuDistOr}

\begin{center}
\includegraphics[scale=.3]{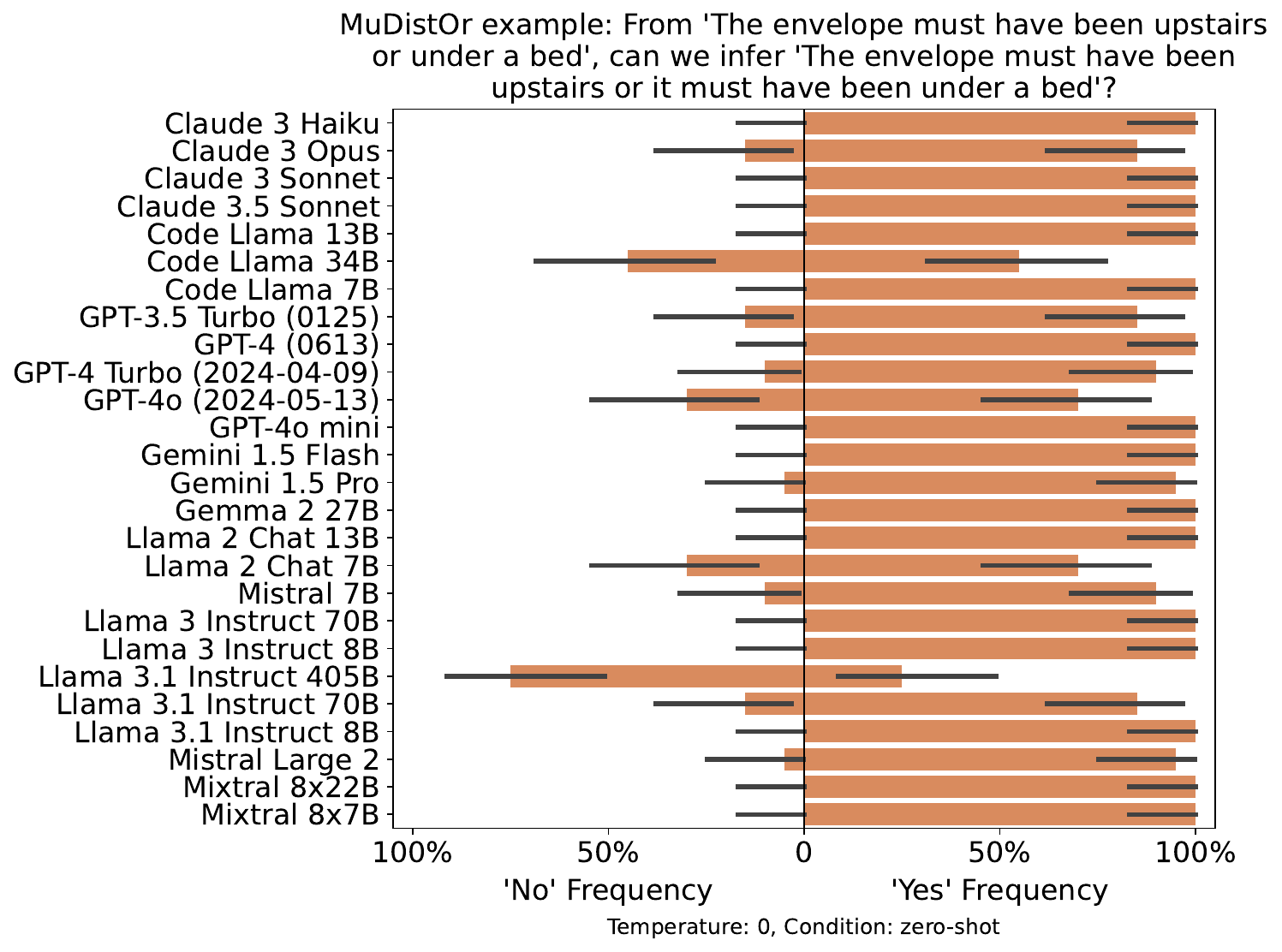}
\includegraphics[scale=.3]{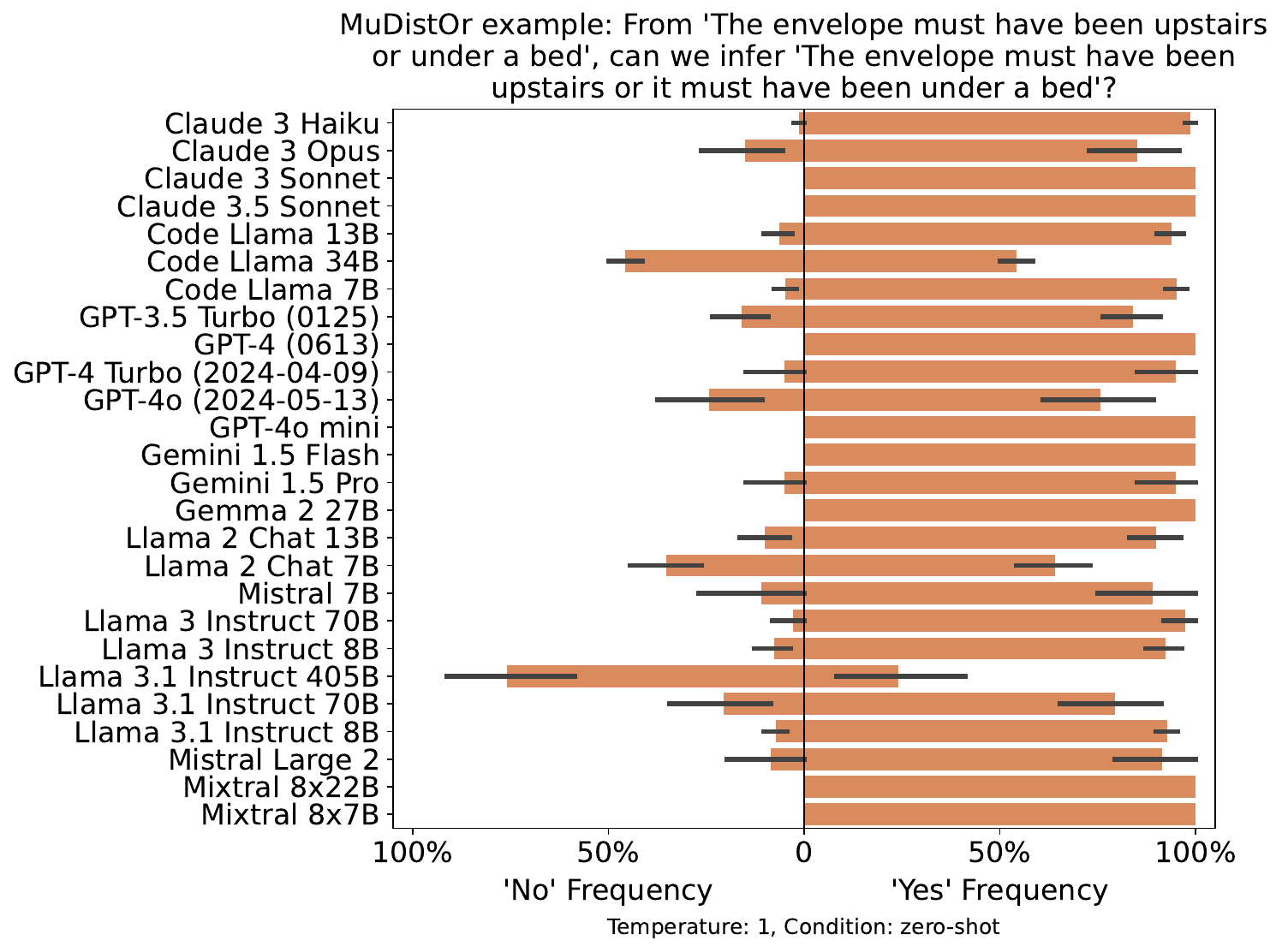}
\end{center}

\subsubsection{`Might' agglomeration (MiAg)}

\begin{center}
\includegraphics[scale=.3]{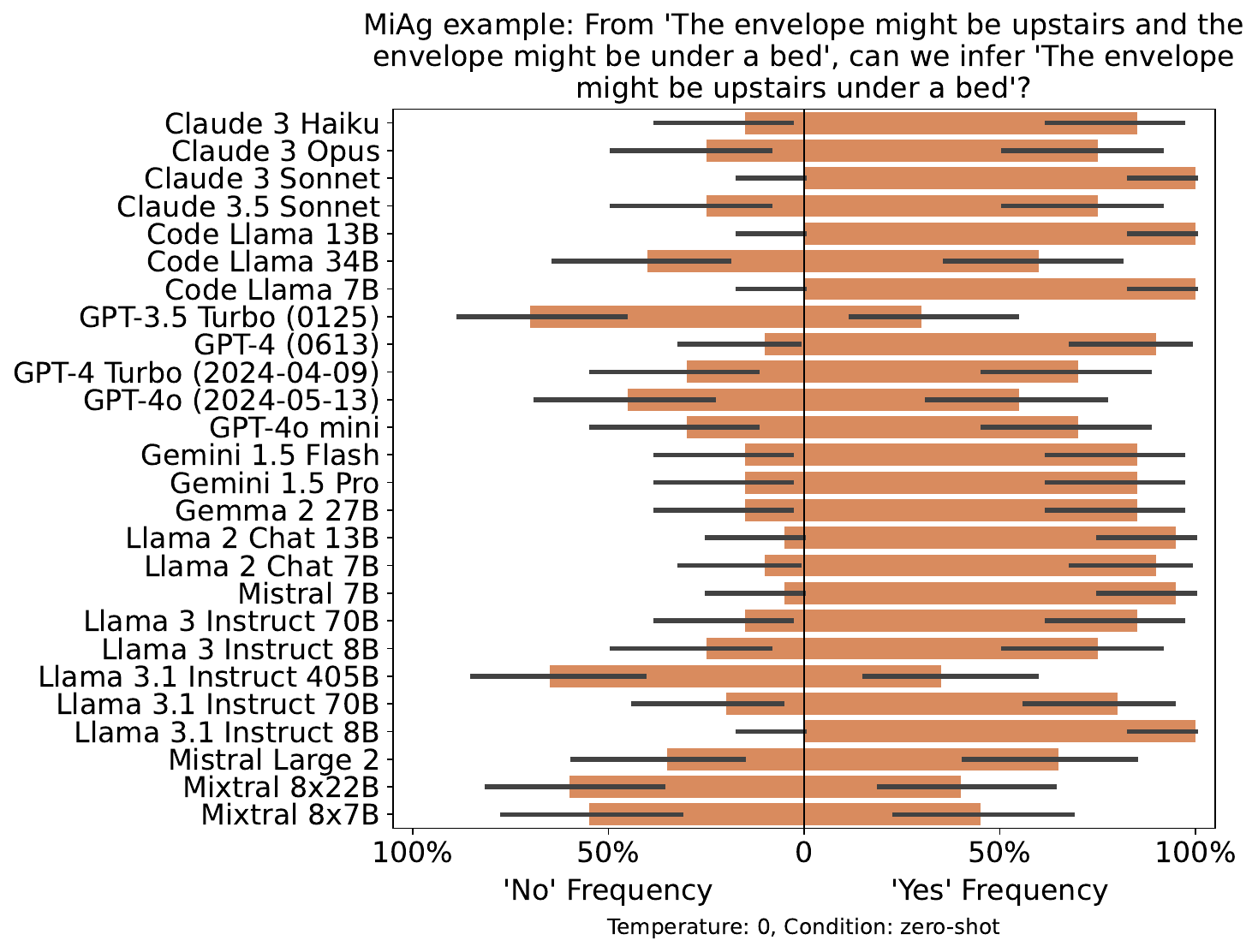}
\includegraphics[scale=.3]{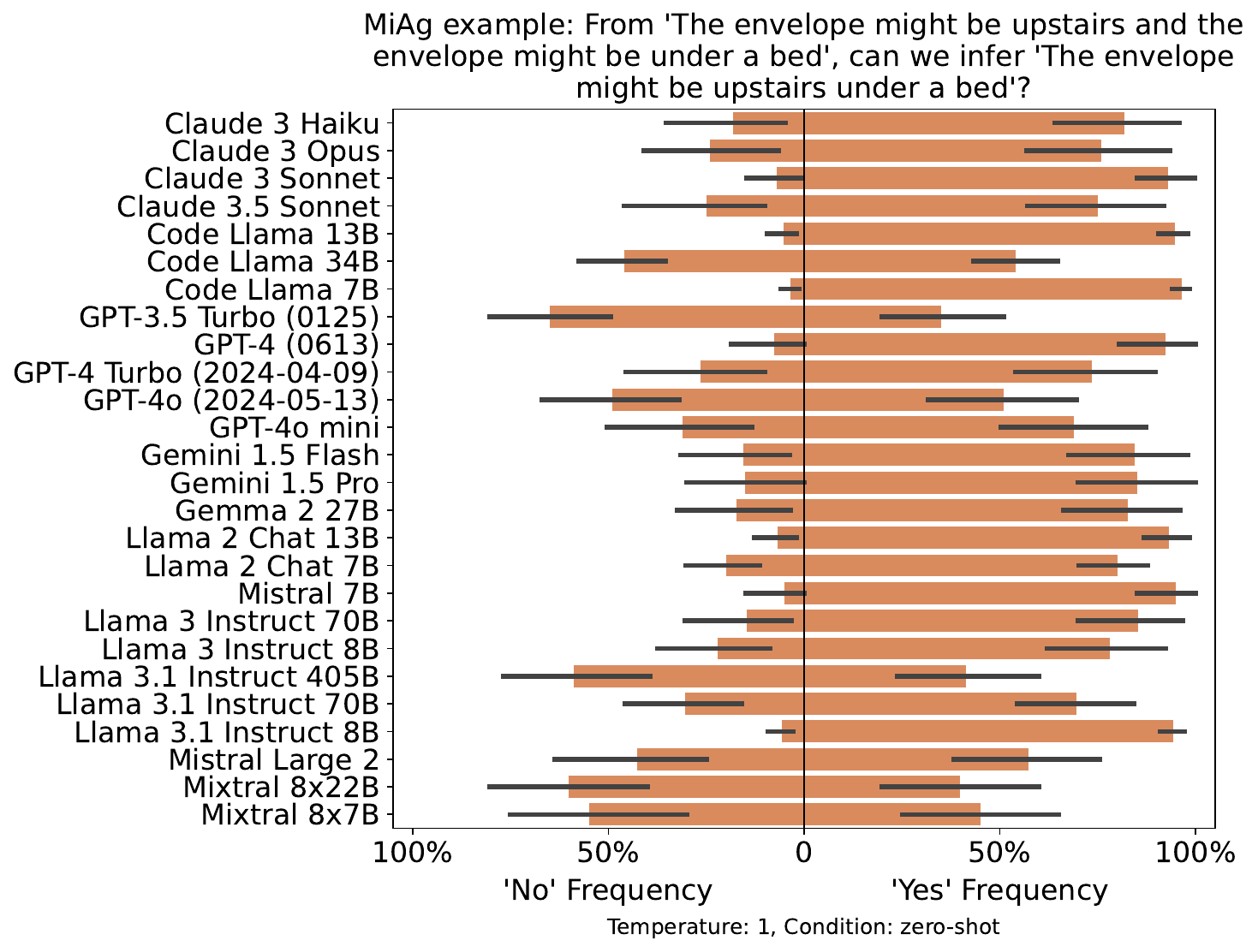}
\end{center}

\newpage

\subsubsection{Narrow-scope free choice (NSFC)}

\begin{center}
\includegraphics[scale=.3]{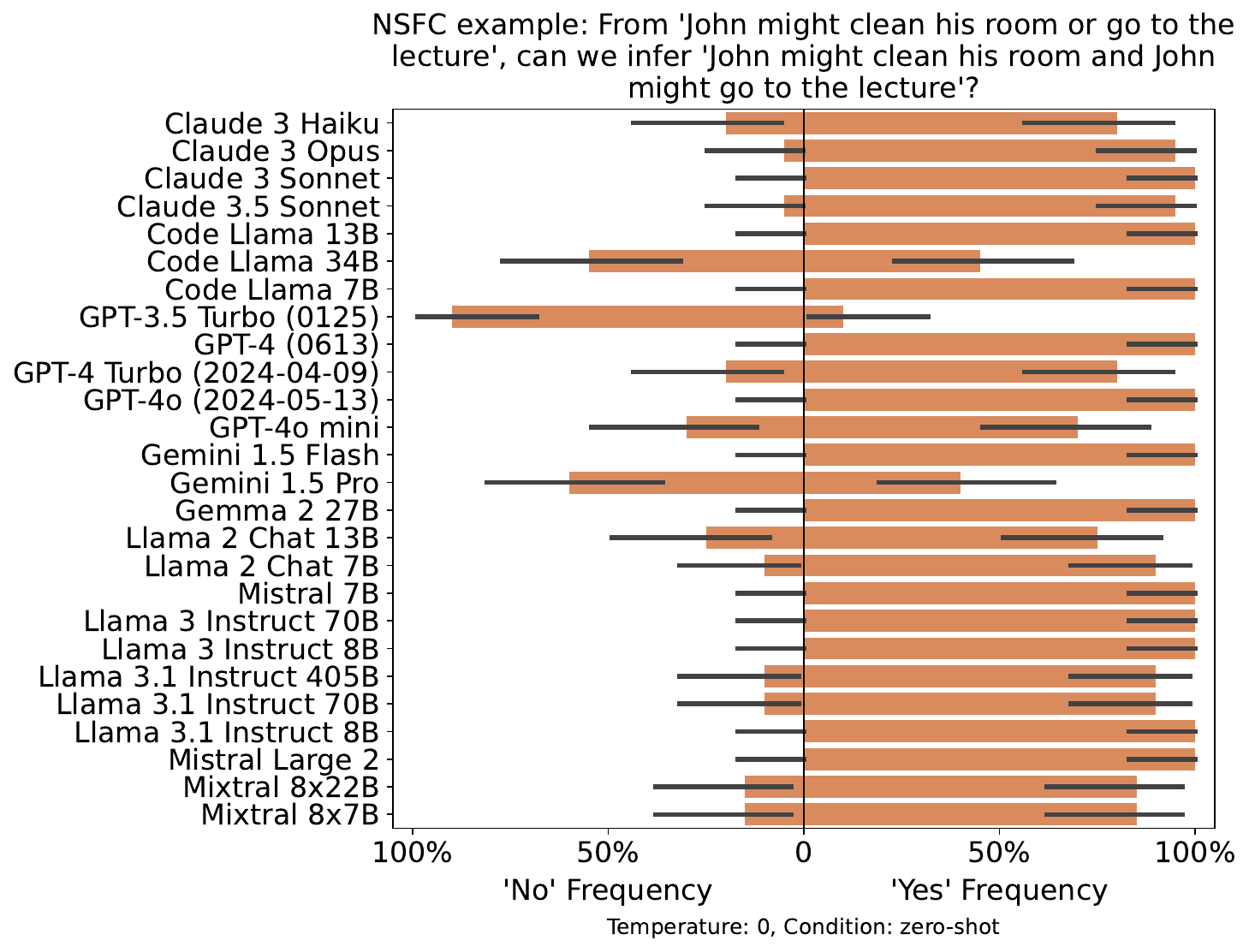}
\includegraphics[scale=.3]{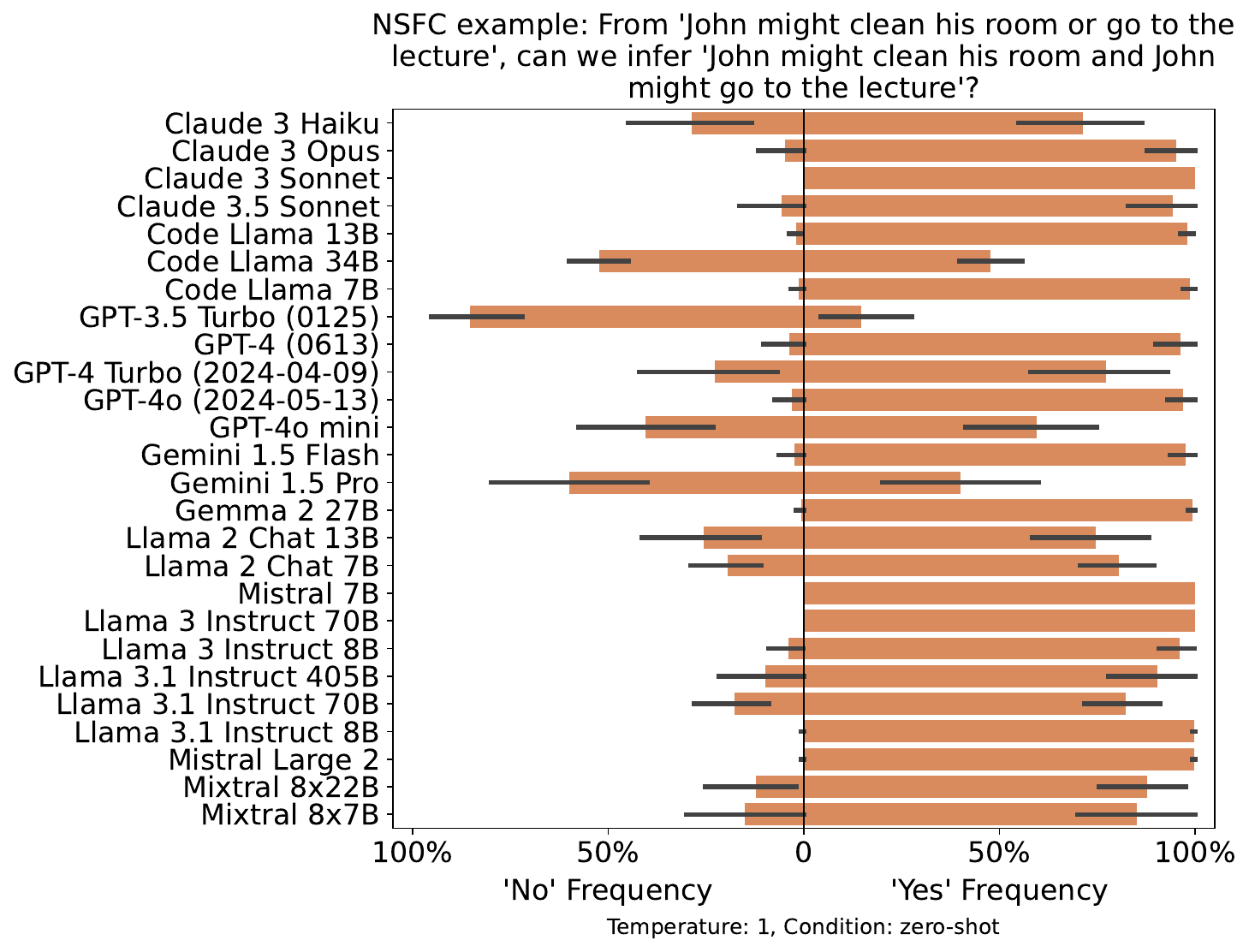}
\end{center}

\subsubsection{Wide-scope free choice (WSFC)}\label{WSFC}

\begin{center}
\includegraphics[scale=.3]{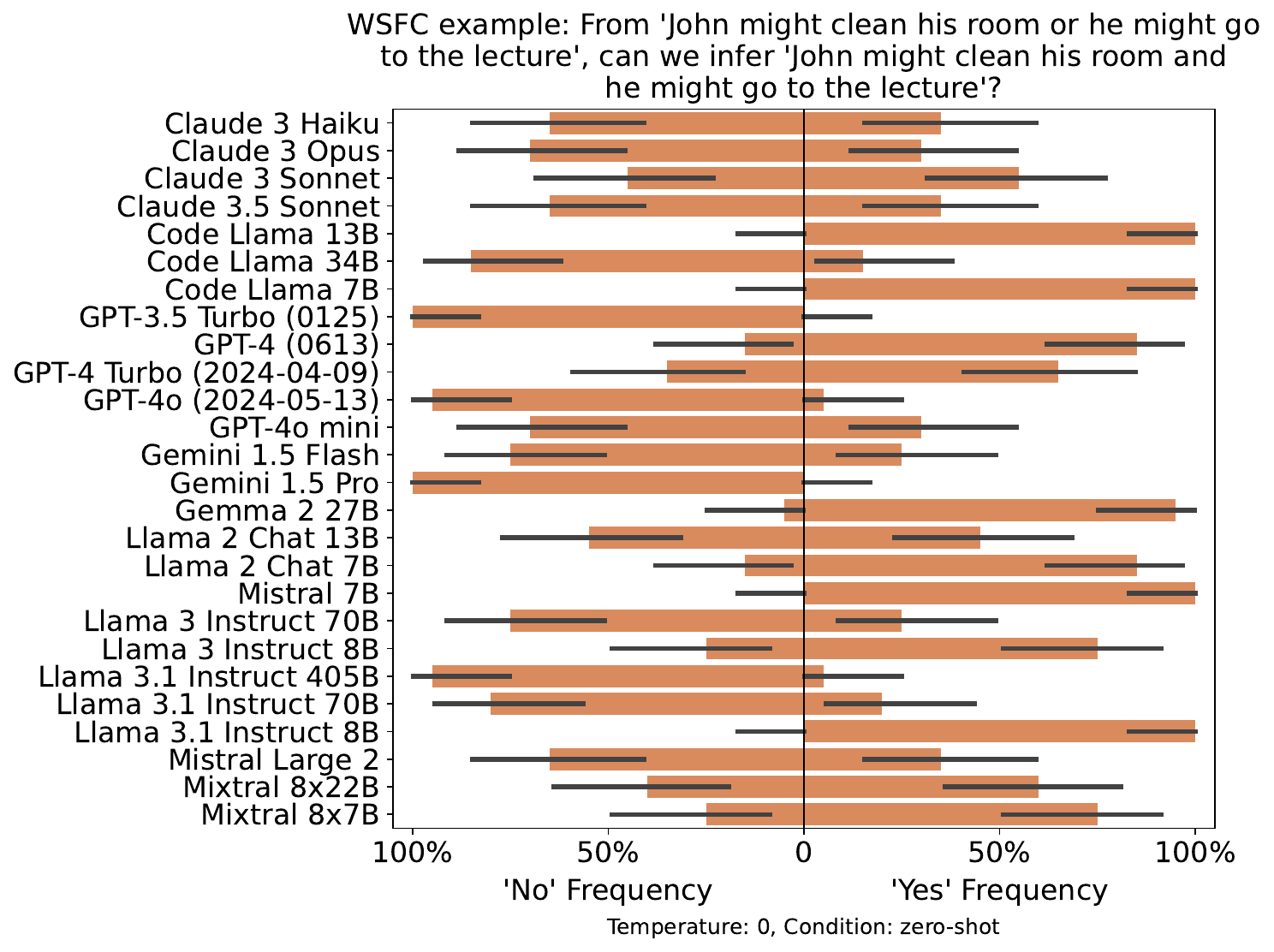}
\includegraphics[scale=.3]{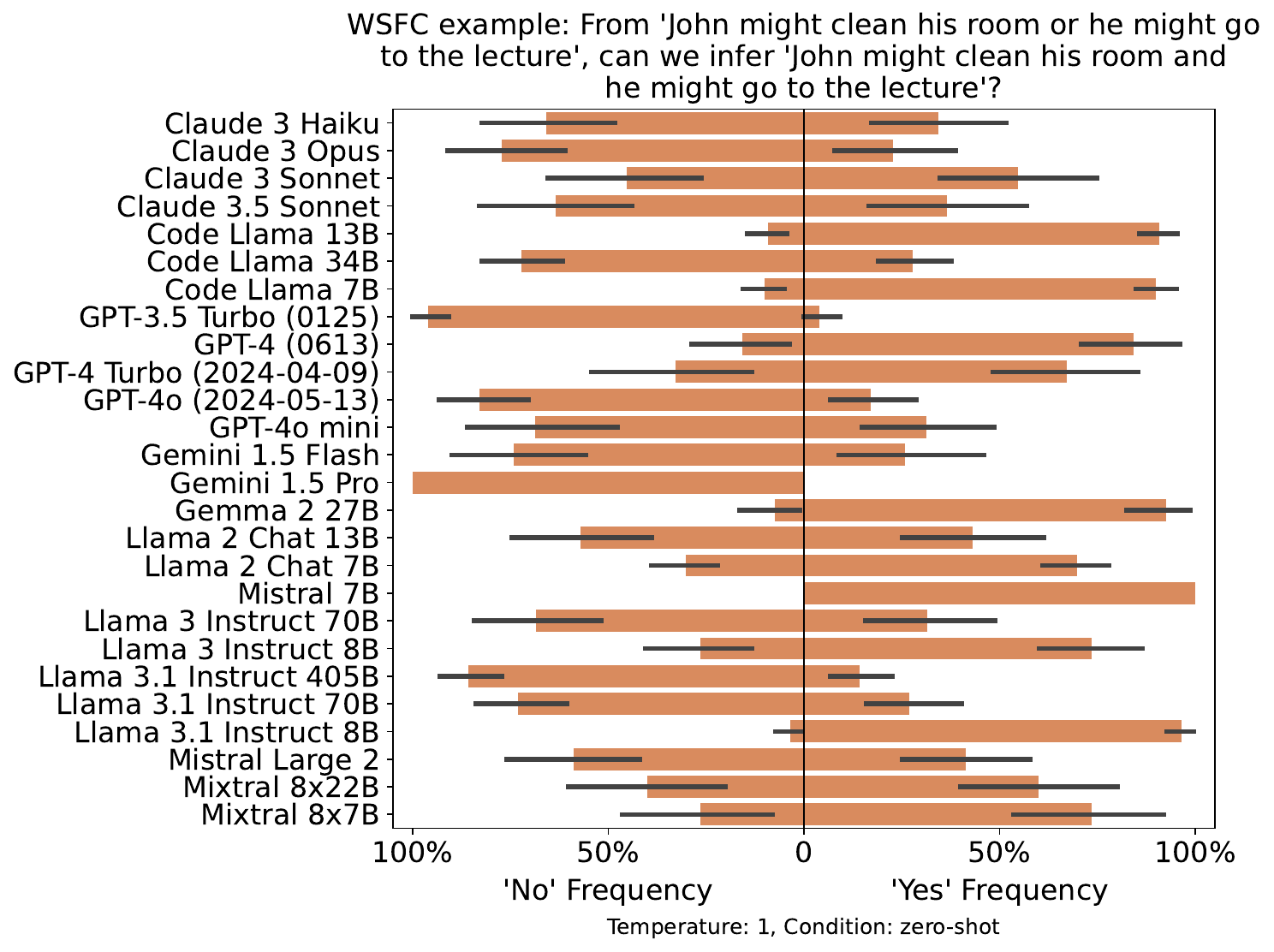}
\end{center}

\newpage

\subsection{Performance summaries}

\begin{figure}[H]
\includegraphics[scale=.3]{figures/summary_graph_multicolor_temp_0_zero-shot.pdf}
\includegraphics[scale=.3]{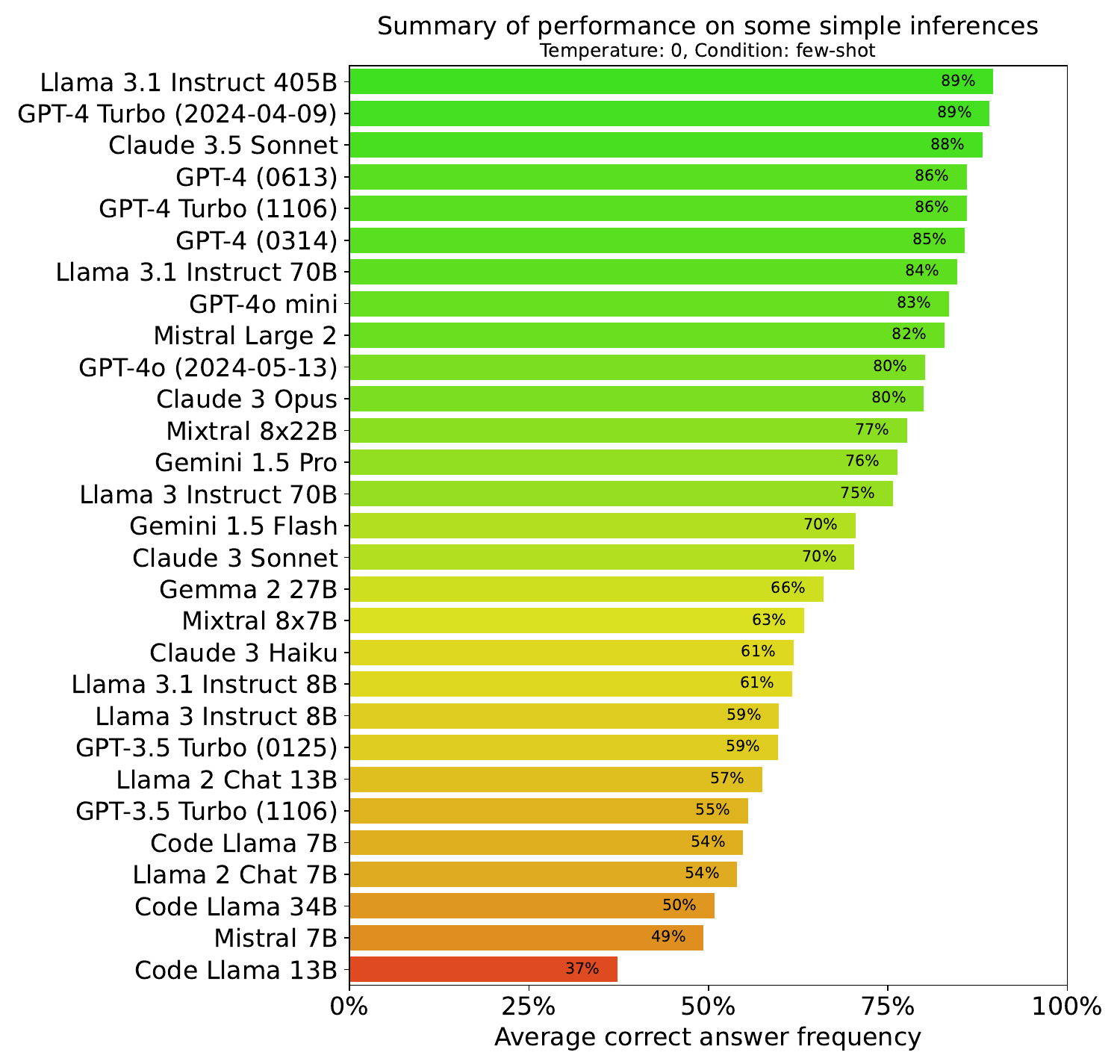}
\includegraphics[scale=.3]{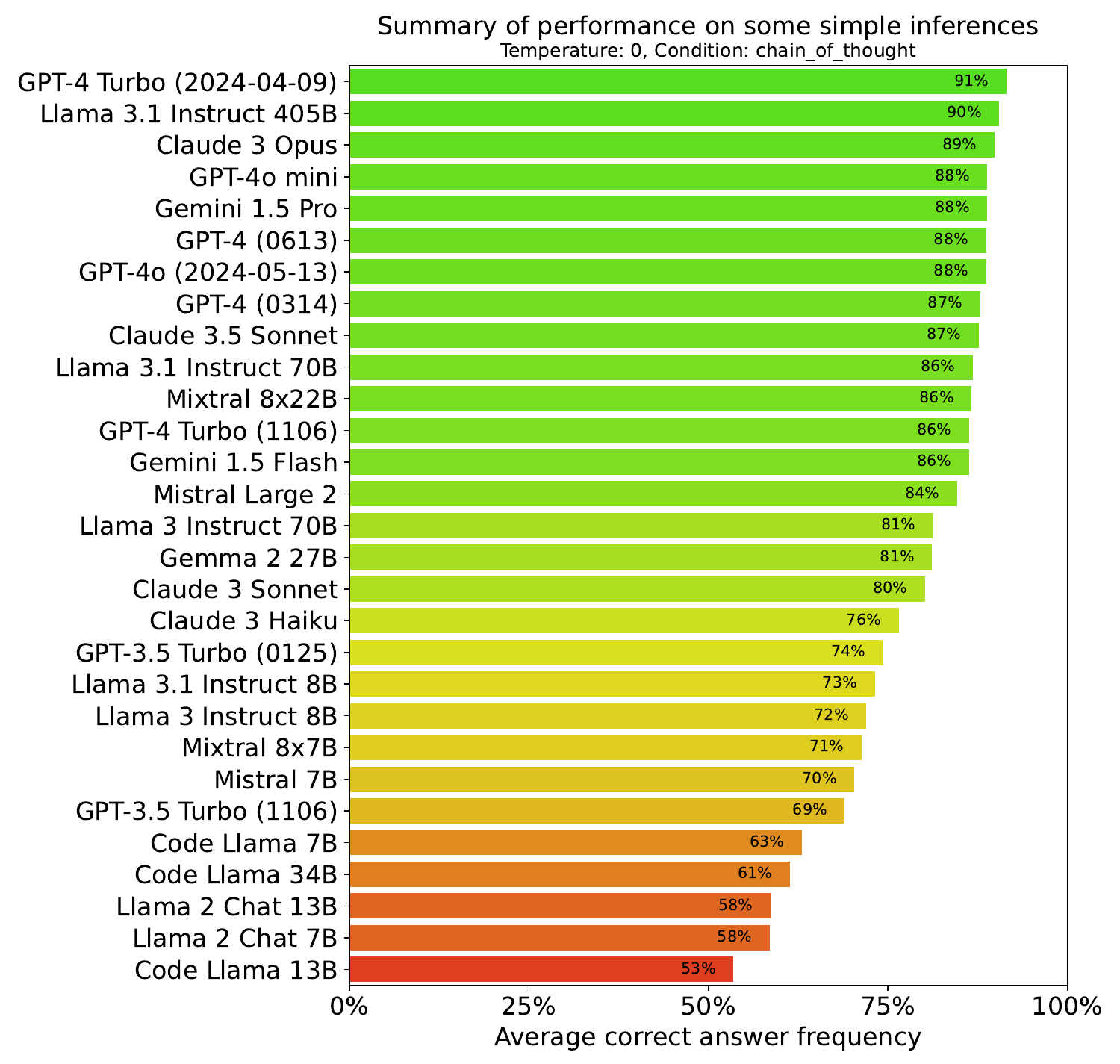}
\caption{Summary of performance on the uncontroversial logical inference patterns under different conditions and temperature 0.}
\end{figure}

\begin{figure}
\includegraphics[scale=.3]{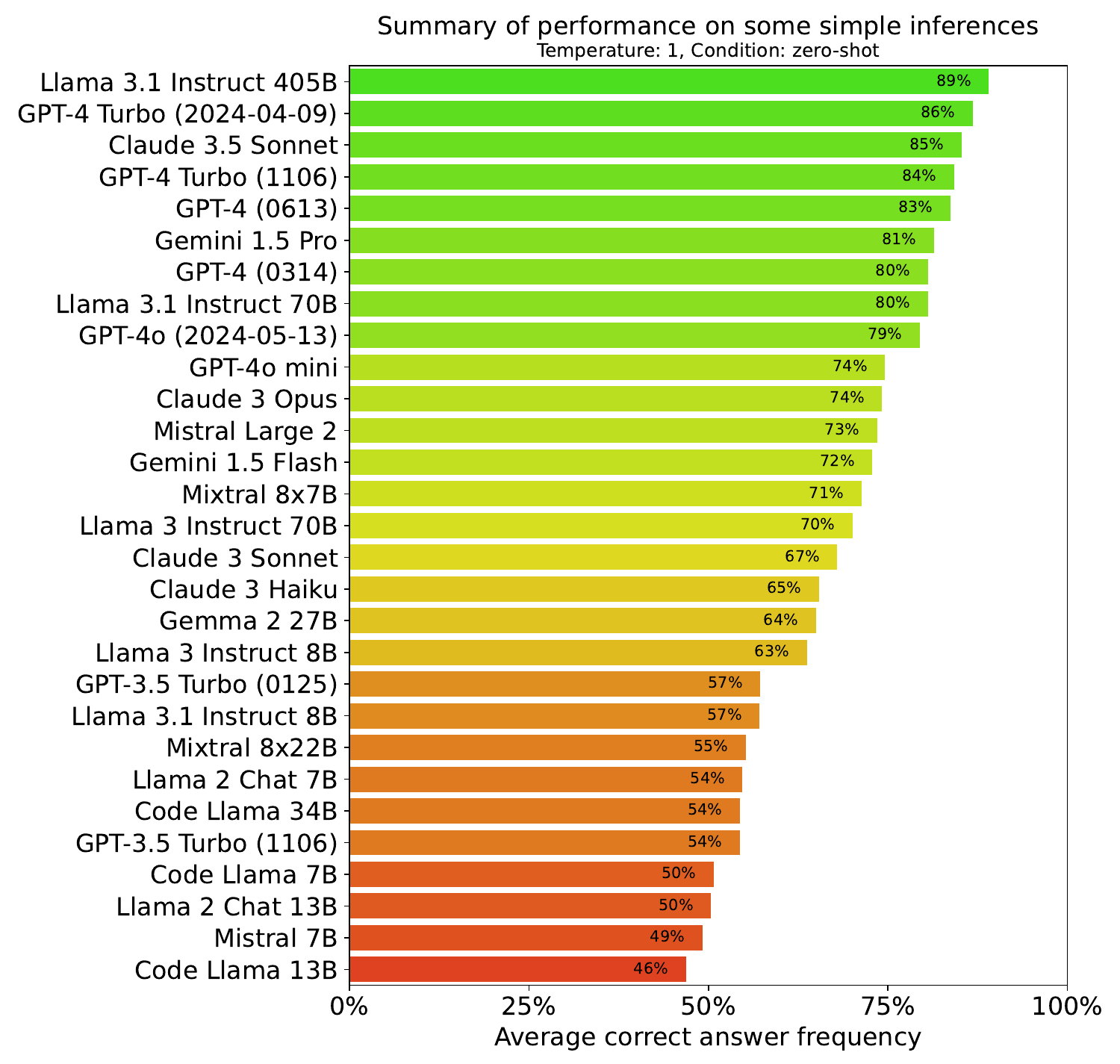}
\includegraphics[scale=.3]{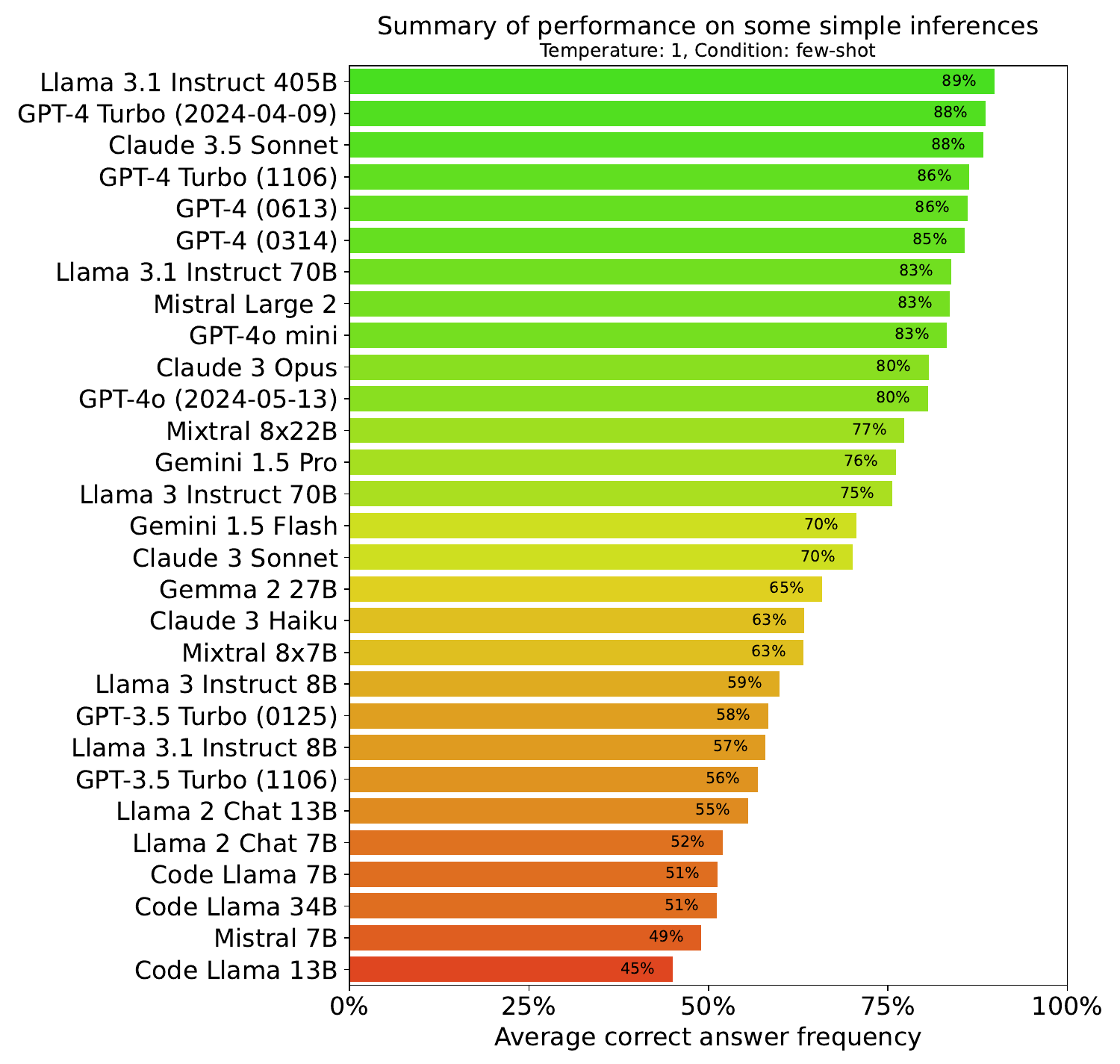}
\caption{Summary of performance on the uncontroversial logical inference patterns under zero-shot and few-shot conditions with temperature 1.}
\end{figure}

\end{document}